%% file: paper_camera_ready.tex
\documentclass[10pt,twocolumn,letterpaper]{article}

\usepackage{cvpr}              % To produce the CAMERA-READY version
\usepackage{graphicx}
\usepackage{amsmath}
\usepackage{amsthm}           % for proof
\usepackage{booktabs}
\usepackage{epsfig}
\usepackage{tabularx}
\usepackage[table,dvipsnames]{xcolor}       % for colored stuff
\usepackage{verbatim}            % For multiline comments
\usepackage{relsize}
\usepackage{multirow}
\usepackage{scrextend}           % Put footmark in the bottom of the page
\usepackage{array}               % Thicker vertical rules in tables, lines in only certain columns
\usepackage{makecell}            % For thicker horizontal rules in tables
\usepackage{subcaption}          % For subfigures
\usepackage[nointegrals]{wasysym}% For right filled arrow
\usepackage{lipsum}              % For random filler text
\usepackage{esvect}              % For vv command arrow above symbols
\usepackage{soul}                % For striked text command \st and hightlight
\usepackage{float}
\usepackage{caption}
\usepackage{bm}
\usepackage{enumitem}            % for noindent itemize
\usepackage{empheq}
\usepackage{gensymb}             % for degree symbol
\usepackage[thicklines]{cancel}  % for cancel
\usepackage{arydshln}            % for horizontal dash line
\usepackage{pifont}              % for ticks and cross http://ctan.org/pkg/pifont
\usepackage{hhline}              % for hhline
\usepackage{tocloft}             % Adjusting TOC

\setlist[itemize, 1]{label =\raisebox{-0.4\height}{\scalebox{1.7}{\textbullet}}}
\setlist[itemize]{noitemsep, topsep=0cm, leftmargin=3mm}
\allowdisplaybreaks              % Break aligned equations
\graphicspath{{images/}}

\advance\cftsecnumwidth 0.25cm\relax
\advance\cftsubsecnumwidth 0.08cm\relax
\advance\cftsubsecnumwidth 0.15cm\relax

\newtheorem{theorem}{Theorem}

\newtheorem{lemma}{Lemma}

\makeatletter
\@namedef{ver@everyshi.sty}{}
\makeatother
\usepackage{tikz} % for some figures
\usetikzlibrary{shapes.geometric} % for trapezium
\usetikzlibrary{calc}  % for centerarc
\usetikzlibrary{arrows}
\usetikzlibrary{arrows.meta}
\usetikzlibrary{shapes.arrows}
\usetikzlibrary{fadings, shadows}
\usetikzlibrary{decorations.text}
\def\centerarc[#1](#2)(#3:#4:#5)% Syntax: [draw options] (center) (initial angle:final angle:radius)
    { \draw[#1] ($(#2)+({#5*cos(#3)},{#5*sin(#3)})$) arc (#3:#4:#5); }

\input{notations}

\newcommand{\methodName}{SeaBird\xspace}
\newcommand{\methodNameFull}{Segmentation in Bird's View\xspace}
\newcommand{\methodNameFullColored}{\textcolor{theme_color}{Se}gment\textcolor{theme_color}{a}tion in \textcolor{theme_color}{Bird}'s View\xspace}
\newcommand{\paperTitle}{\textcolor{theme_color}{\methodName}: \methodNameFullColored with Dice Loss Improves\\Monocular 3D Detection of Large Objects\xspace}

\usepackage[pagebackref,breaklinks,colorlinks,citecolor=cvprblue,urlcolor=theme_color]{hyperref}
\usepackage[capitalize]{cleveref}
\crefname{section}{Sec.}{Secs.}
\Crefname{section}{Section}{Sections}
\Crefname{table}{Table}{Tables}
\crefname{table}{Tab.}{Tabs.}

\def\paperID{7680} % *** Enter the Paper ID here
\def\confName{CVPR}
\def\confYear{2024}

\title{\paperTitle}

\author{Abhinav Kumar$^{1}$\qquad Yuliang Guo$^{2}$\qquad Xinyu Huang$^{2}$\qquad Liu Ren$^{2}$\qquad Xiaoming Liu$^{1}$\\
~~~~~$^{1}$Michigan State University\qquad~~$^{2}$Bosch Research North America, Bosch Center for AI~\\
{\small\tt~~$^{1}$[kumarab6,liuxm]@msu.edu ~~~ \tt\small$^{2}$[yuliang.guo2,xinyu.huang,liu.ren]@us.bosch.com}\\
{\small\url{https://github.com/abhi1kumar/SeaBird}}
}

\begin{document}
%============================================================================
% Title
%============================================================================
\twocolumn[{%
\renewcommand\twocolumn[1][]{#1}%
\maketitle
\vspace{-1mm}
\noindent\begin{minipage}{\linewidth}
    \centering
    \begin{minipage}[t]{.28\linewidth}
        \includegraphics[width=\linewidth]{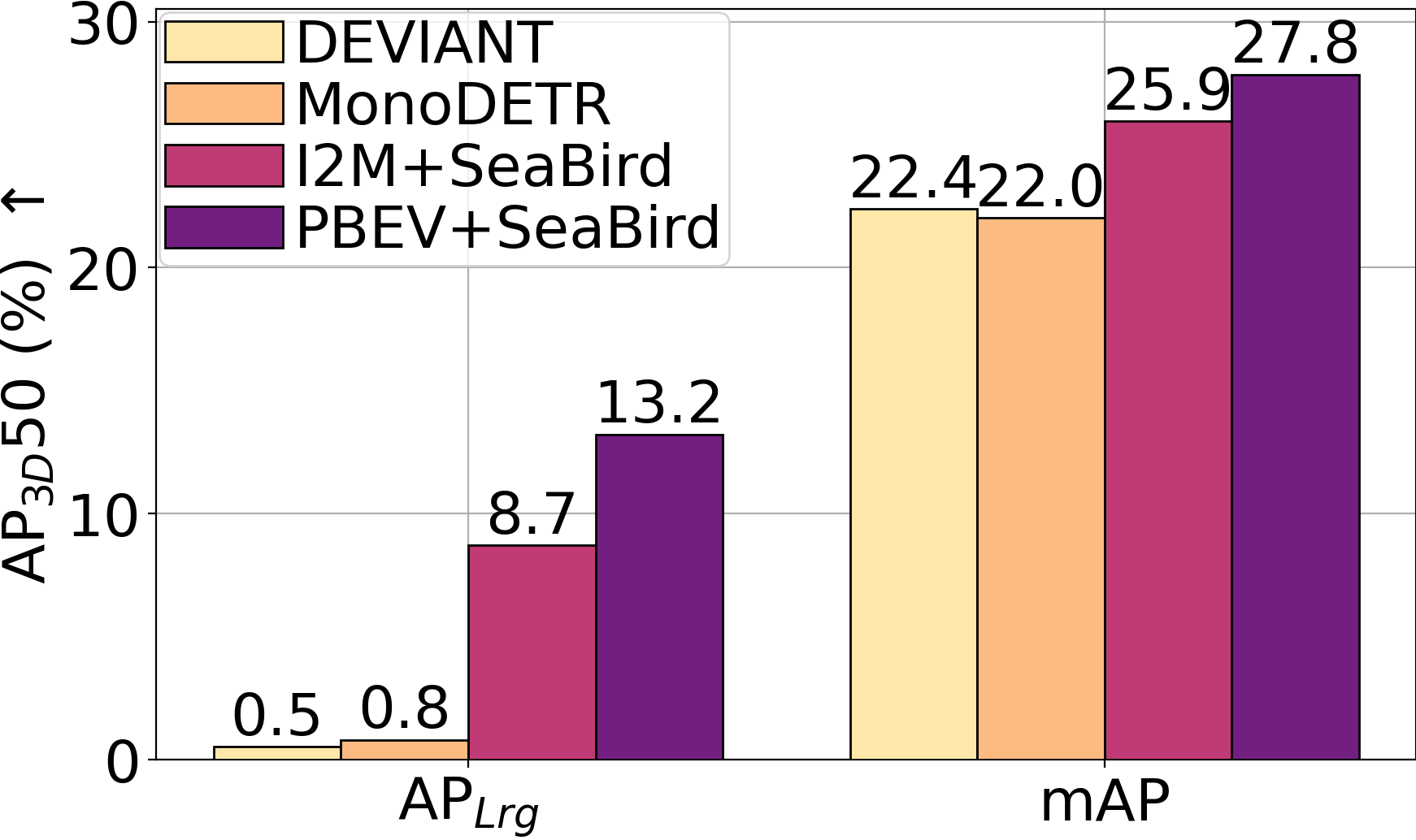}\\
        \vspace{-0.6cm}
        \captionof*{figure}{(a) Improve \kittiThreeSixty \val \sota.}
    \end{minipage}%
    \begin{minipage}[t]{.41\linewidth}
        \includegraphics[width=\linewidth]{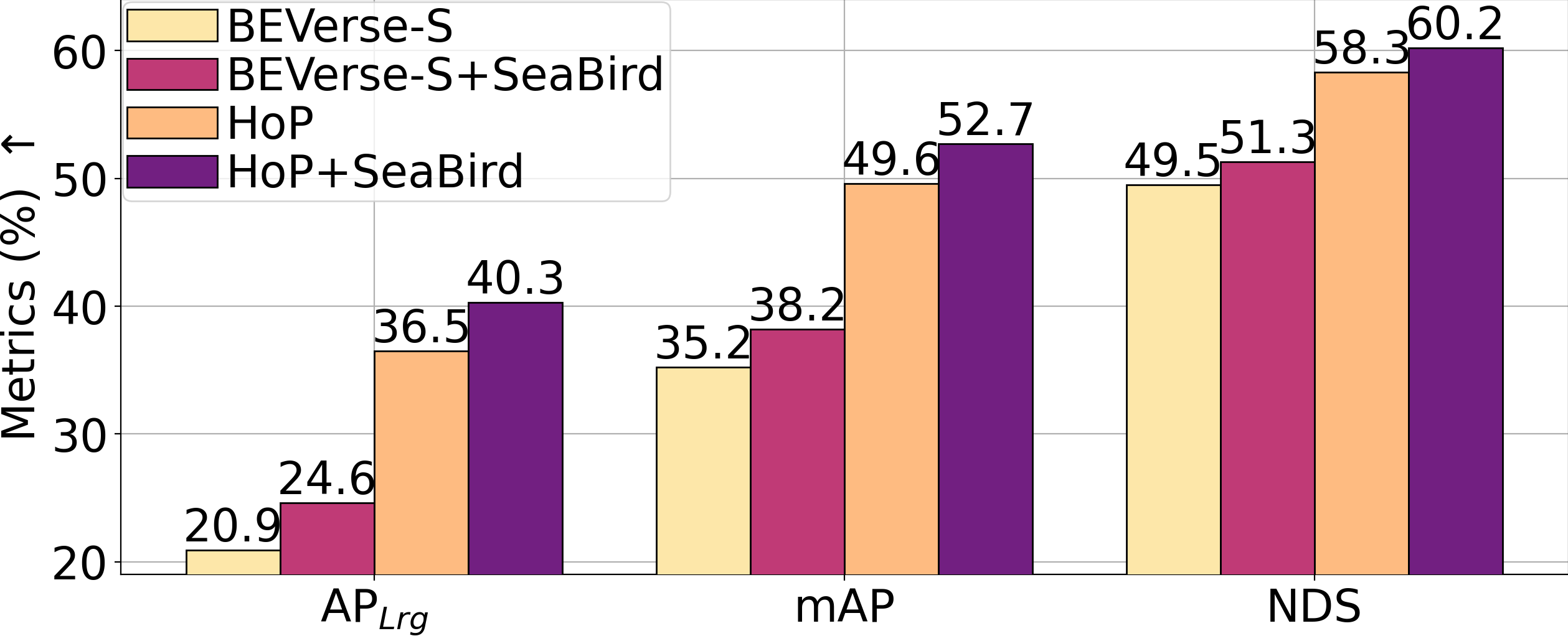}\\
        \vspace{-0.6cm}
        \captionof*{figure}{(b) Improve \nuscenes \val \sota.}
    \end{minipage}%\hfill
    \begin{minipage}[t]{.28\linewidth}
        \includegraphics[width=\linewidth]{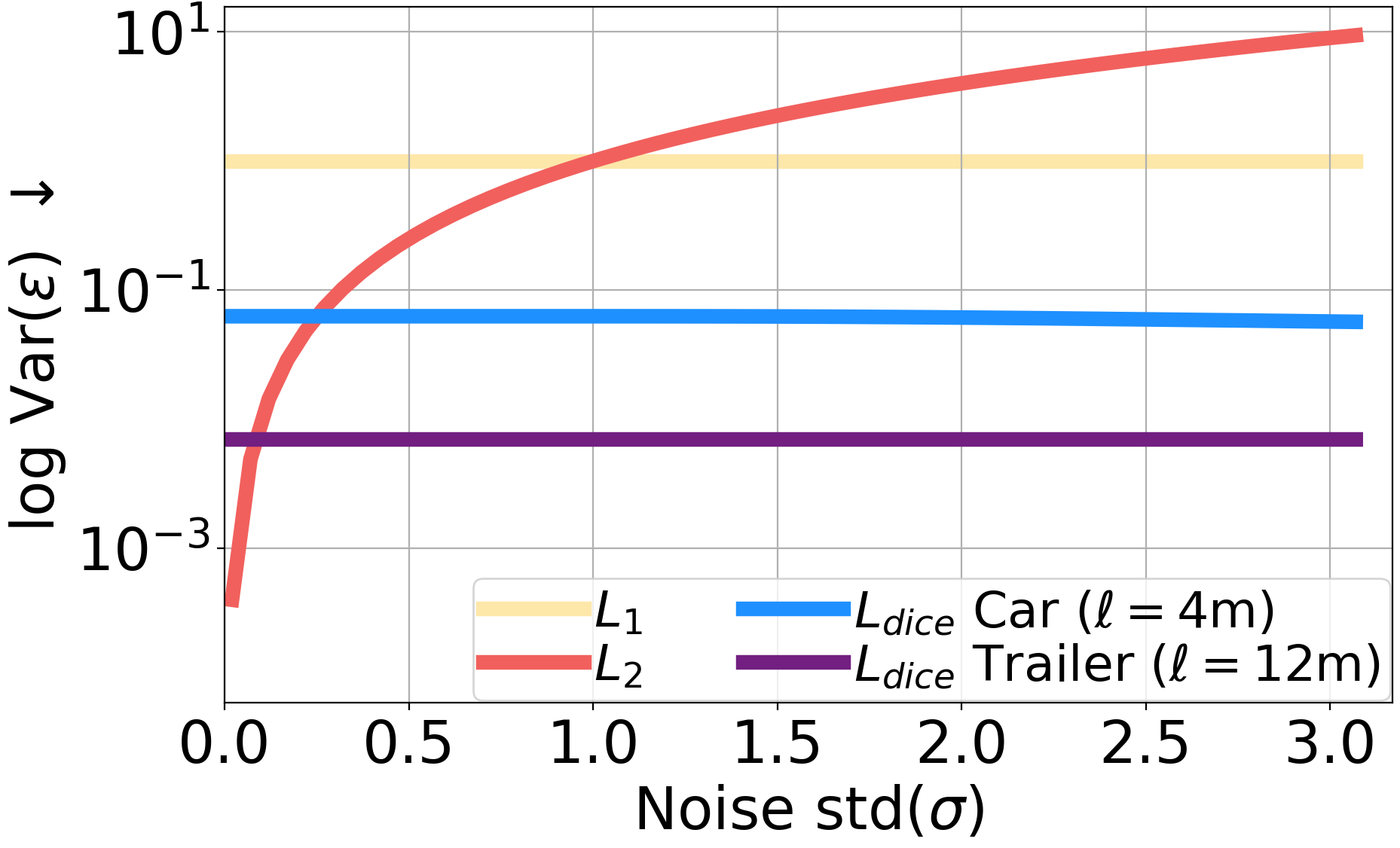}\\
        \vspace{-0.6cm}
        \captionof*{figure}{(c) Theory Advancement.}
    \end{minipage}
    \vspace{-0.2cm}
    \captionof{figure}
    {\textbf{Teaser (a)} \sota frontal detectors struggle with large objects (low \MAPLarge) even on a nearly balanced \kittiThreeSixty dataset (Skewness in \cref{fig:skew}).
        Our proposed \methodName achieves significant \monoThreeD improvements, particularly for large objects. 
        \textbf{(b)} \methodName also improves two \sota \bev detectors, \beVerseSmall \cite{zhang2022beverse} and \hop \cite{zong2023hop} on the \nuscenes dataset, particularly for large objects.
        \textbf{(c)} Plot of convergence variance $\var(\funcNoise)$ of \dice and regression losses with the noise $\normalSig$ in depth prediction. 
        The $y$-axis denotes the deviation from the optimal weight, so the lower the better. \methodName leverages \textbf{\dice loss}, which we prove is more noise-robust than regression losses for large objects.
    }
    \label{fig:teaser}
\end{minipage}
\vspace{0.2cm}
}]

%============================================================================
%============================================================================
%============================================================================
\begin{abstract}
\vspace{-0.33cm}
    Monocular \threeD detectors achieve remarkable performance on cars and smaller objects. 
    However, their performance drops on larger objects, leading to fatal accidents. 
    Some attribute the failures to training data scarcity or their receptive field requirements of large objects.
    In this paper, we highlight this understudied problem of generalization to large objects.
    We find that modern frontal detectors struggle to generalize to large objects even on nearly balanced datasets.
    We argue that the cause of failure is the sensitivity of depth regression losses to noise of larger objects.
    To bridge this gap, we comprehensively investigate regression and dice losses, examining their robustness under varying error levels and object sizes.
    We mathematically prove that the dice loss leads to superior noise-robustness and model convergence for large objects compared to regression losses for a simplified case.
    Leveraging our theoretical insights, we propose \methodName (\methodNameFull) as the first step towards generalizing to large objects.
    \methodName effectively integrates BEV segmentation on foreground objects for 3D detection, with the segmentation head trained with the dice loss.
    \methodName achieves SoTA results on the \kittiThreeSixty leaderboard and improves existing detectors on the \nuscenes leaderboard, particularly for large objects. 
\end{abstract}

    \begin{figure*}[!t]
        \centering
        \includegraphics[width=0.85\linewidth]{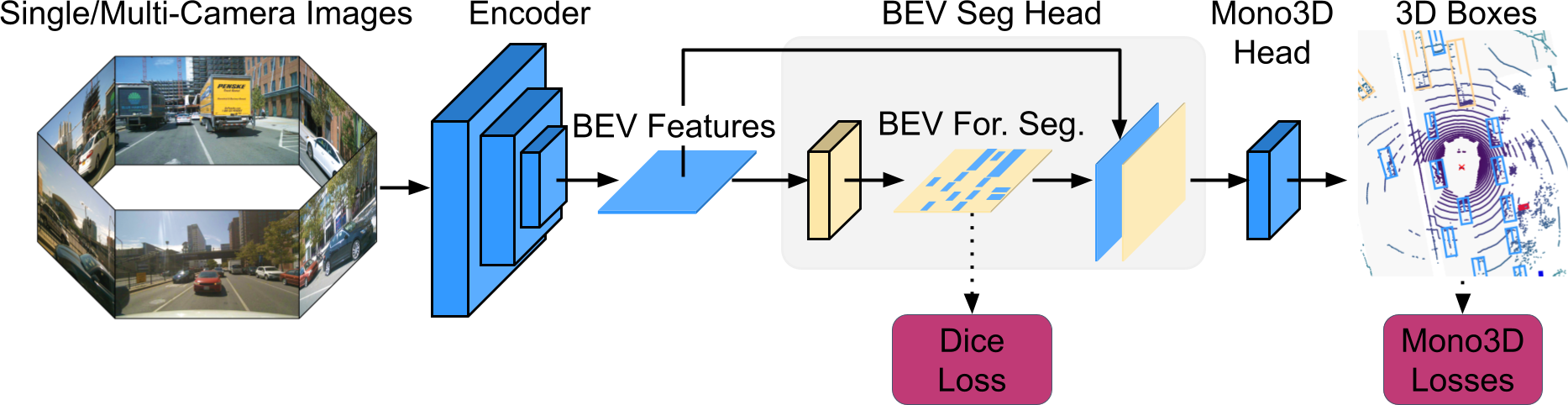}
        \caption{
            \textbf{\methodName Pipeline.} 
            \methodName uses the predicted \bev foreground segmentation (For. Seg.) map to predict accurate \threeD boxes for large objects.
            \methodName training protocol involves \bev segmentation pre-training with the noise-robust \dice loss and \monoThreeD fine-tuning.
        }
        \label{fig:pipeline}
        \vspace{-0.5cm}
    \end{figure*}

\addtocontents{toc}{\protect\setcounter{tocdepth}{-2}}
%============================================================================
%============================================================================
%============================================================================
\section{Introduction}\label{sec:intro}

    Monocular \threeD object detection (\monoThreeD) task aims to estimate both the \threeD position and dimensions of objects in a scene from a single image. 
    Its applications span autonomous driving \cite{park2021pseudo,kumar2022deviant,li2022bevformer}, robotics \cite{saxena2008robotic}, and augmented reality \cite{alhaija2018augmented,Xiang2018RSS,park2019pix,merrill2022symmetry}, where accurate \threeD understanding of the environment is crucial. 
    Our study focuses explicitly on \threeD object detectors applied to autonomous vehicles (AVs), considering the challenges and motivations deviate drastically across different applications.

    AVs demand object detectors that generalize to diverse intrinsics \cite{brazil2023omni3d}, camera-rigs \cite{jia2023monouni,tzofi2023towards}, rotations \cite{moon2023rotation}, weather and geographical conditions \cite{dong2023benchmarking} and also are robust to adversarial examples \cite{zhu2023understanding}.
    Since each of these poses a significant challenge, recent works focus exclusively on the generalization of object detectors to all these out-of-distribution shifts. 
    However, our focus is on the generalization of another type, which, thus far, has been understudied in the literature – {\it \monoThreeD generalization to large objects}. 
    
    Large objects like trailers, buses and trucks are harder to detect \cite{wu2023talk} in \monoThreeD, sometimes resulting in fatal accidents \cite{caldwell2022tesla,fernandez2023tesla}. 
    Some attribute these failures to training data scarcity \cite{zhu2019class} or the receptive field requirements \cite{wu2023talk} of large objects, but, to the best of our knowledge, no existing literature provides a comprehensive analytical explanation for this phenomenon.
    The goal of this paper is, thus, to bring understanding and a first analytical approach to this real-world problem in the AV space – \monoThreeD generalization to large objects. 

    We conjecture that the generalization issue stems not only from limited training data or larger receptive field but also from the noise sensitivity of depth regression losses in \monoThreeD. 
    To substantiate our argument, we analyze the \monoThreeD performance of state-of-the-art (\sota) frontal detectors on the \kittiThreeSixty dataset \cite{liao2022kitti360}, which includes almost equal number ($1\!:\!2$) of large objects and cars.
    We observe that \sota detectors struggle with large objects on this  dataset (\cref{fig:teaser}\textcolor{red}{a}).
    Next, we carefully investigate the SGD convergence of losses used in \monoThreeD task and mathematically prove that the \dice loss, widely used in \bev segmentation, exhibits superior noise-robustness than the regression losses, particularly for large objects (\cref{fig:teaser}\textcolor{red}{c}).
    Thus, the \dice loss facilitates better model convergence than regression losses, improving \monoThreeD of large objects.

    Incorporating \dice loss in detection introduces unique challenges. 
    Firstly, the \dice loss does not apply to sparse detection centers and only incorporates depth information when used in the \bev space. 
    Secondly, naive joint training of \monoThreeD and \bev segmentation tasks with image inputs does not always benefit \monoThreeD task \cite{li2022bevformer,ma2022vision} due to negative transfer \cite{crawshaw2020multi}, and the underlying reasons remain unclear. 
    Fortunately, many \monoThreeD segmentors and detectors are in the \bev space, where the \bev segmentor can seamlessly apply \dice loss and the \bev detector can readily benefit from the segmentor in the same space. 
    To mitigate negative transfer, we find it effective to train the \bev segmentation head on the foreground detection categories. 
    
    Building upon our theoretical findings about the \dice loss, we propose a simple and effective pipeline called \methodNameFull (\methodName) for enhancing \monoThreeD of large objects. 
    \methodName employs a sequential approach for the \bev segmentation and \monoThreeD heads (\cref{fig:pipeline}).  
    \methodName first utilizes a \bev segmentation head to predict the segmentation of only foreground objects, supervised by the \dice loss. 
    The \dice loss offers superior noise-robustness for large objects, ensuring stable convergence, while focusing on foreground objects in segmentation mitigates negative transfer. 
    Subsequently, \methodName concatenates the resulting \bev segmentation map with the original \bev features as an additional feature channel and feeds this concatenated feature to a \monoThreeD head supervised by \monoThreeD losses\footnote{Only \monoThreeD head predicts additional \threeD attributes, namely object's height and elevation.}. 
    Building upon this, we adopt a two-stage training pipeline: the first stage exclusively focuses on training the \bev segmentation head with \dice loss, which fully exploits its noise-robustness and superior convergence in localizing large objects. The second stage involves both the detection loss and dice loss to finetune the \monoThreeD head.
    
    In our experiments, we first comprehensively evaluate \methodName and conduct ablations on the balanced single-camera \kittiThreeSixty dataset \cite{liao2022kitti360}. 
    \methodName outperforms the \sota baselines by a substantial margin. 
    Subsequently, we integrate \methodName as a plug-in-and-play module into two \sota detectors on the multi-camera \nuscenes dataset \cite{caesar2020nuscenes}.
    \methodName again significantly improves the original detectors, particularly on large objects. 
    Additionally, \methodName consistently enhances \monoThreeD performance across backbones with those two \sota detectors (\cref{fig:teaser}\textcolor{red}{b}), demonstrating its utility in both edge and cloud deployments.

    In summary, we make the following contributions:
    \begin{itemize}
        \item We highlight the understudied problem of generalization to large objects in \monoThreeD, showing that even on nearly balanced datasets, \sota frontal models struggle to generalize due to the noise sensitivity of regression losses.
        \item We mathematically prove that the \dice loss leads to superior noise-robustness and model convergence for large objects compared to regression losses for a simplified case and provide empirical support for more general settings.
        \item We propose \methodName, which 
        treats \bev segmentation head on foreground objects and \monoThreeD head sequentially and trains in a two-stage protocol to fully harness the noise-robustness of the \dice loss.
        \item We empirically validate our theoretical findings and show significant improvements, particularly for large objects, on both \kittiThreeSixty and \nuscenes leaderboards.
    \end{itemize}

%============================================================================
%============================================================================
%============================================================================
\section{Related Work}

    %============================================================================
    \noIndentHeading{Mono3D.}
        \monoThreeD popularity stems from its high accessibility from consumer vehicles compared to \lidar/Radar-based detectors \cite{shi2019pointrcnn,yin2021center,long2023radiant} and computational efficiency compared to stereo-based detectors \cite{Chen2020DSGN}.
        Earlier approaches \cite{payet2011contours, chen2016monocular} leverage hand-crafted features, while the recent ones use deep networks. 
        Advancements include introducing new architectures \cite{shi2023multivariate,huang2022monodtr,xu2023mononerd}, equivariance \cite{kumar2022deviant, chen2023viewpoint}, losses \cite{brazil2019m3d,chen2020monopair}, uncertainty \cite{lu2021geometry,kumar2020luvli} and incorporating auxiliary tasks such as depth \cite{zhang2021objects,min2023neurocs}, 
        NMS \cite{shi2020distance,kumar2021groomed,liu2023monocular}, corrected extrinsics \cite{zhou2021monoef}, CAD models \cite{chabot2017deep, liu2021autoshape, lee2023baam} or \lidar \cite{reading2021categorical} in training.
        A particular line of work called \pseudoLidar \cite{wang2019pseudo, ma2019accurate} shows generalization by first estimating the depth, followed by a point cloud-based \threeD detector.
        
        Another line of work encodes image into latent \bev features \cite{ma2023towards} and attaches multiple heads for downstream tasks \cite{zhang2022beverse}. 
        Some focus on pre-training \cite{xie2022m2bev} and rotation-equivariant convolutions \cite{feng2022aedet}. 
        Others introduce new coordinate systems \cite{jiang2023polarformer}, queries \cite{luo2022detr4d,li2023fast}, or positional encoding \cite{shu2023dppe} in a transformer-based detection framework \cite{carion2020detr}.
        Some use pixel-wise depth \cite{huang2021bevdet}, object-wise depth \cite{chu2023oabev,choi2023depth,liu2021voxel}, or depth-aware queries \cite{zhang2023dabev}, while many utilize temporal fusion \cite{wang2022sts,wang2023stream,liu2023petrv2,brazil2020kinematic} to boost performance.
        A few use longer frame history \cite{park2022time,zong2023hop}, distillation \cite{klingner2023x3kd,wang2023distillbev} or stereo \cite{wang2022sts,li2023bevstereo}.
        We refer to \cite{ma20233d,ma2022vision} for the survey.
        \methodName also builds upon the \bev-based framework since it flexibly accepts single or multiple images as input and uses \dice loss. 
        Different from the majority of other detectors, %which develop advanced \bev feature encoding, 
        \methodName improves \monoThreeD of large objects using the power of \dice loss. 
        \methodName is also the first work to mathematically prove and justify this loss choice for large objects.

    %============================================================================
    \noIndentHeading{\bev Segmentation.}
        \bev segmentation typically utilizes \bev features transformed from \twoD image features. Various methods encode single or multiple images into \bev features using MLPs \cite{pan2020cross} or transformers \cite{roddick2020predicting,saha2022translating}. 
        Some employ learned depth distribution \cite{philion2020lift,hu2021fiery}, while others use attention \cite{saha2022translating,zhou2022cross} or attention fields \cite{chitta2021neat}. 
        \imageToMapsLong \cite{saha2022translating} utilizes polar ray, while \panopticBEVLong \cite{gosala2022bev} uses transformers. 
        FIERY \cite{hu2021fiery} introduces uncertainty modelling and temporal fusion, while Simple-\bev \cite{harley2022simple} uses \radar aggregation.
        Since \bev segmentation lacks object height and elevation, one also needs a \monoThreeD head to predict \threeD boxes.

    %============================================================================
    \noIndentHeading{Joint \monoThreeD and \bev Segmentation.}
        Joint \threeD detection and \bev segmentation using \lidar data \cite{shi2019pointrcnn,fan2022fully} as input benefits both tasks \cite{yang2023lidar,wang2023segmentation}. 
        However, joint learning on image data often hinders detection performance \cite{li2022bevformer,zhang2022beverse,xie2022m2bev,ma2022vision}, while the \bev segmentation improvement is inconsistent across categories \cite{ma2022vision}.
        Unlike these works which treat the two heads in parallel and decrease \monoThreeD performance \cite{ma2022vision}, \methodName treats the heads sequentially and increases \monoThreeD performance, particularly for large objects.

%============================================================================
%============================================================================
%============================================================================
\section{\methodName}\label{sec:proposed}

    \methodName is driven by a deep understanding of the distinctions between monocular regression and \bev segmentation losses. 
    Thus, in this section, we delve into the problem and discuss existing results. 
    We then present our theoretical findings and, subsequently, introduce our pipeline.
    
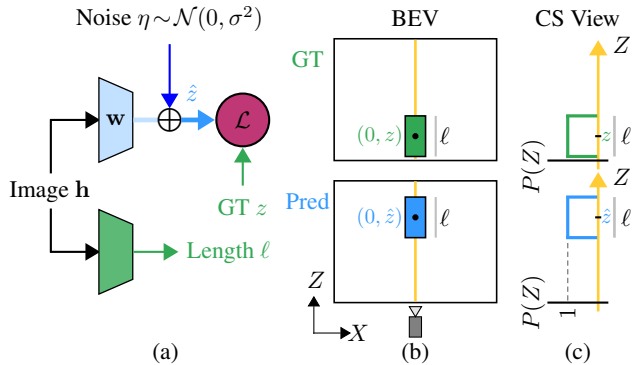
\begin{figure}[!t]
        \centering
        \input{images/problem_setup}
        \vspace{-0.6cm}
        \caption{
            \textbf{(a) Problem setup}. The single-layer neural network takes an image $\image$ (or its features) and predicts depth $\depthPred$ and the object length $\length$.
            The noise $\noise$ is the additive error in depth prediction and is a normal random variable.
            The GT depth $\depthGT$ supervises the predicted depth $\depthPred$ with a loss $\loss$ in training.
            We assume the network predicts the GT length $\length$.
            Frontal detectors directly regress the depth with $\lOne$, $\lTwo$, or $\smoothLOne$ loss, while \methodName projects to \bev plane and supervises through \dice loss $\lDice$.
            \textbf{(b) Shifting of \textcolor{my_blue}{predictions}} in \bev along the \textcolor{rayShade}{ray} due to the noise $\noise$.
            \textbf{(c) Cross Section (CS) view} along the \textcolor{rayShade}{ray} with classification scores $P(Z)$.
        }
        \label{fig:problem_setup}
        \vspace{-0.5cm}
    \end{figure}

    We introduce the problem and refer to \cref{lemma:1} from the literature \cite{shalev2007pegasos, lacoste2012simpler}, which \emph{evaluates} loss quality by measuring the deviation of trained weight (after SGD updates) from the optimal weight. 
    \cref{fig:problem_setup}\textcolor{red}{a} illustrates the problem setup.
    Figs. \ref{fig:problem_setup}\textcolor{red}{b} and \ref{fig:problem_setup}\textcolor{red}{c} visualize the \bev and cross-section view, respectively. 
    Since this deviation depends on the gradient variance of losses, we next derive the gradient variance of the \dice loss in \cref{lemma:2}.
    By comparing the distance between trained weight and optimal weight, we assess the effectiveness of \dice loss versus MAE $(\lOne)$ and MSE $(\lTwo)$ losses in \cref{lemma:3}, and choose the representation and loss combination. 
    Combining these findings, we establish \cref{th:1} that the model trained with \dice loss achieves better \ap than the model trained with regression losses. 
    Finally, we present our pipeline, \methodName, which integrates \bev segmentation supervised by \dice loss for \monoThreeD.

    %============================================================================
    %============================================================================
    \subsection{Background and Problem Statement}
        \monoThreeD networks \cite{lu2021geometry, kumar2022deviant} commonly employ regression losses, such as $\lOne$ or $\lTwo$ loss, to compare the predicted depth with ground truth (GT) depth~\cite{kumar2022deviant, zhang2022beverse}. 
        In contrast, \bev segmentation utilizes \dice loss \cite{saha2022translating} or cross-entropy loss \cite{hu2021fiery} at each \bev location, comparing it with GT. 
        Despite these distinct loss functions, we evaluate their effectiveness under an idealized model, where we measure the model \emph{quality} by the expected deviation of trained weight (after SGD updates) from the optimal weight \cite{shalev2007pegasos}.\\
        \vspace{-0.6cm}
        \begin{lemma}\label{lemma:1}
            \textbf{Convergence analysis }\cite{shalev2007pegasos}.
            Consider a linear regression model with trainable weight $\layerWeight$ for depth prediction $\depthPred$ from an image $\image$.
            % Let $\layerWeightOptimal$ denote the optimal weight for predicting depth. 
            Assume the noise $\noise$ is an additive error in depth prediction and is a normal random variable $\normal(0, \normalVar)$.
            Also, assume SGD optimizes the model parameters with loss function $\loss$ during training with square summable steps $\step_j$, \thatIs $\stepSumTrue\!=\!\lim\limits_{t \rightarrow \infty} \sum\limits_{j=1}^\instant \step_j^2$ exists and $\noise$ is independent of the image. 
            Then, the expected deviation of the trained weight $\layerWeightConv$ from the optimal weight $\layerWeightOptimal$ obeys
            \begin{align}
                \label{eqn:conv:weight:dist}
                \expect\left(\norm{\layerWeightConv\!-\!\layerWeightOptimal}_2^2\right) &= \stepConstant \var(\funcNoise) + \uselessConstant,
            \end{align}
            where $\funcNoise\!=\!\frac{\partial \loss(\noise)}{\partial \noise}$ is the gradient of the loss $\loss$ wrt noise, $\stepConstant\!=\!\stepSumTrue\expect(\image^T\image)$ and $\uselessConstant$ are constants independent of the loss.
        \end{lemma}

        We refer to \cref{sec:proof_converged} for the proof.
        \cref{eqn:conv:weight:dist} demonstrates that training losses $\loss$ exhibit varying gradient variances $\var(\funcNoise)$. 
        Hence, comparing this term for different losses allows us to evaluate their quality.

    %============================================================================
    %============================================================================
    \subsection{Loss Analysis: \Dice vs. Regression}

        Given that \cite{shalev2007pegasos} provides the gradient variance $\var(\funcNoise)$, for $\lOne$ and $\lTwo$ losses, we derive the corresponding gradient variance for \dice and \iou losses in this paper to facilitate comparison. 
        First, we express the \dice loss, $\lDice$, as a function of noise $\noise$ as per its definition from \cite{saha2022translating} for \cref{fig:problem_setup}\textcolor{red}{c} as:
        \begin{align}
            \lDice(\noise) = 1\!-\!2\frac{\text{Pred}~\text{GT}}{\text{Pred} + \text{GT}}
            &= \begin{cases}
                1\!-\!2\frac{\length-|\noise|}{2\length} \text{ , }|\noise|\le \length \\
                1 \qquad\qquad \text{, }|\noise|\ge \length
                \end{cases} \nonumber \\
            \implies \lDice(\noise) &= \begin{cases}
                \frac{|\noise|}{\length} \text{ , }|\noise|\le \length\\
                1\quad\text{, }|\noise|\ge \length 
               \end{cases} ,
            \label{eq:dice}
        \end{align}
        where $\length$ denotes the object length. 
        \cref{eq:dice} shows that the \dice loss $\lDice$ depends on the object size $\length$. 
        % We visualize this dependency in the \bev and cross-section view in Figs. \ref{fig:problem_setup}\textcolor{red}{b} and \ref{fig:problem_setup}\textcolor{red}{c}, respectively. 
        With the given \dice loss $\lDice$, we proceed to derive the following lemma:

    \begin{table}[!t]
        \caption{\textbf{Convergence variance} of training loss functions. 
        Gradient variance of $\lDice$ is more noise-robust for large objects, resulting in better detectors.
        We do not analyze cross-entropy loss theoretically since its Var$(\funcNoise)$ is infinite, but empirically in \cref{tab:ablation}.}
        \label{tab:optimality_bounds}
        \centering
        \vspace{-0.2cm}
        \scalebox{\scaleFraction}{
        \setlength\tabcolsep{0.1cm}
        \begin{tabular}{l|c|c}
            \myTopRule
            Loss $\loss$ & Gradient $\funcNoise$ & Var$(\funcNoise)$ (\downarrowRHDSmall)\\ %$\frac{\expect((\layerWeightConv\!-\!\layerWeightOptimal)^2)\!-\!\uselessConstant}{\stepConstant}$ \\
            \hline
            $\lOne$ \cite{shalev2007pegasos} (App. \ref{sec:supp_var_lOne}) & $\sign(\noise)$ & $1$ \\
            $\lTwo$ \cite{shalev2007pegasos} (App. \ref{sec:supp_var_lTwo}) & $\noise$ & $\normalVar$ \\
            \Dice (\cref{lemma:2})& $\begin{cases}
                \frac{\sign(\noise)}{\length} \quad\text{  , }|\noise|\le \length \\
                0 \quad\quad~~~~\text{ , }|\noise|\ge \length 
            \end{cases}$& $\dfrac{1}{\length^2}\normalErf\left(\dfrac{\length}{\sqrt{2}\normalSig}\right)$\\
            % \iou (App.~\ref{sec:supp_var_iou})& $\begin{cases}
            % \frac{2\length\sign(\noise)}{(\length+|\noise|)^2}~\text{ , }|\noise|\le \length \\
            % 0 \qquad~~~~~\text{ , }|\noise|\ge \length
            % \end{cases}$& $\approx \frac{4}{\length^2}\left(1\!-\!\frac{\normalSig}{\bigNumber}\right)$\\
            \myTopRule
        \end{tabular}
        }
        \vspace{-0.4cm}
    \end{table}
    
    \begin{lemma}\label{lemma:2}
        \textbf{Gradient variance of \dice loss.}
        Let $\noise= \normal(0, \normalVar)$ be an additive normal random variable and $\length$ be the object length.
        Let $\normalErf$ be the error function.
        Then, the gradient variance of the \dice loss $\var_{dice}(\funcNoise)$ wrt noise $\noise$ is
        \begin{align}
            \var_{dice}(\funcNoise) &= \frac{1}{\length^2}\normalErf\left(\frac{\length}{\sqrt{2}\normalSig}\right).
            \label{eq:grad_var_dice}
        \end{align}
    \end{lemma}
        We refer to \cref{sec:supp_var_dice} for the proof.  
        \cref{eq:grad_var_dice} shows that gradient variance of the \dice loss $\var_{dice}(\funcNoise)$ also varies inversely to the object size $\length$ and the noise deviation $\normalSig$ (See \cref{sec:supp_dice_properties}). 
        These two properties of \dice loss are particularly beneficial for large objects. 

        \cref{tab:optimality_bounds} summarizes these losses, their gradients, and gradient  variances. 
        With $\var_{\dice}(\funcNoise)$ derived for the \dice loss, we now compare the deviation of trained weight with the deviations from $\lOne$ or $\lTwo$ losses, leading to our next lemma.

    \begin{figure}[!t]
        \centering
        \includegraphics[width=0.7\linewidth]{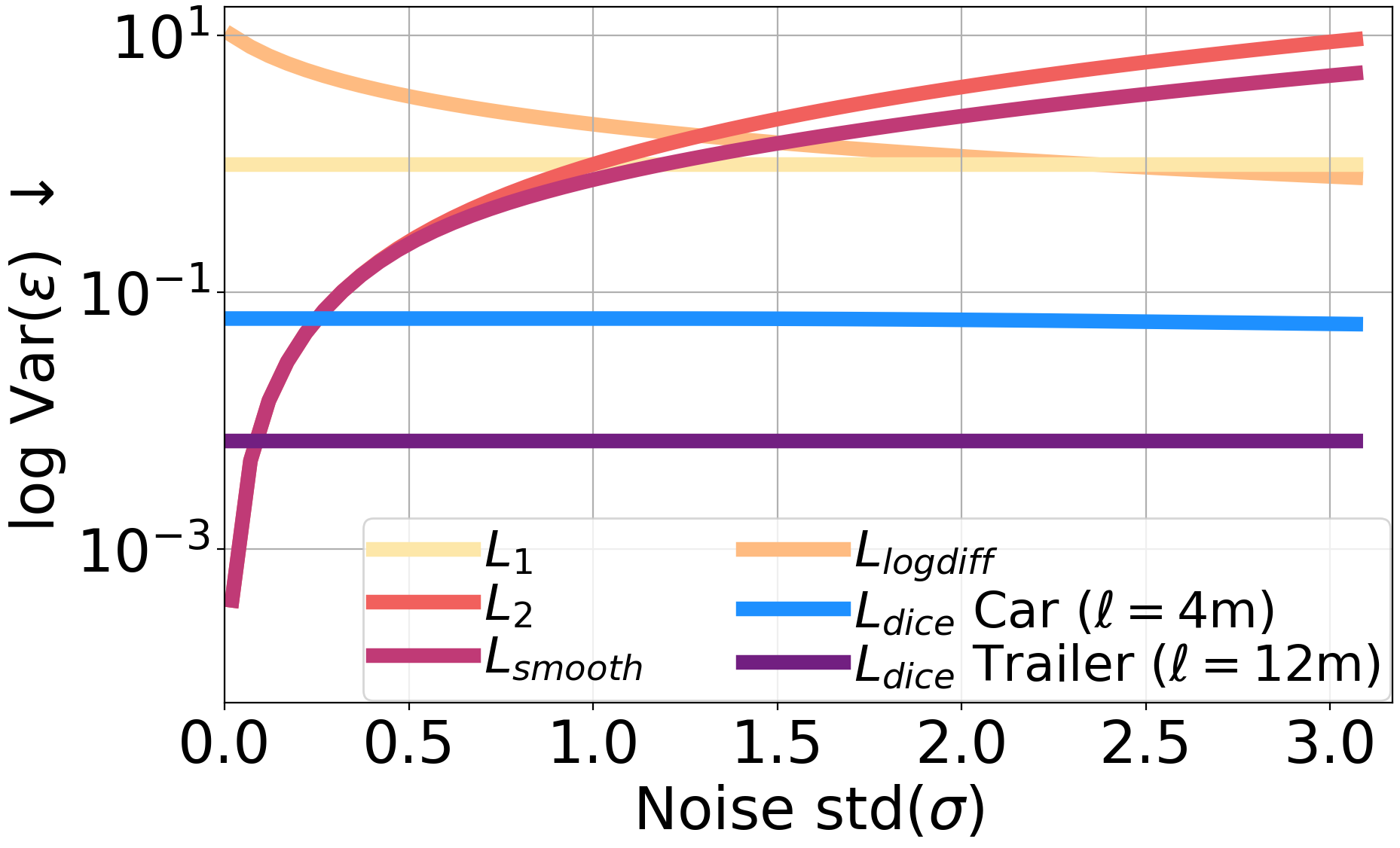}
        \vspace{-0.25cm}
        \caption{\textbf{Plot of convergence variance} Var$(\funcNoise)$ of loss functions with the noise $\normalSig$. \Dice loss has minimum convergence variance with large noise, resulting in better detectors for large objects.}
        \label{fig:conv_analysis}
        \vspace{-0.38cm}
    \end{figure}
    \begin{lemma}\label{lemma:3}
        \textbf{\Dice model is closer to optimal weight than regression loss models.}
        Based on \cref{lemma:1} and assuming the object length $\length$ is a constant,  
        if $\normalSigTh$ is the solution of the equation $\normalVar\!=\!\frac{1}{\length^2}\normalErf\left(\frac{\length}{\sqrt{2}\normalSig}\right)$ and the noise deviation $\normalSig\!\ge\! \normalSigCr\!=\!\max \left(\normalSigTh, \frac{\sqrt{2}}{\length}\normalErfInv(\length^2)\right)$, then the converged weight $\layerWeightConvDice$ with the \dice loss $\lDice$ is better than the converged weight $\layerWeightConvReg$ with the $\lOne$ or $\lTwo$ loss, \thatIs
        \begin{align}
            \expect\left(\norm{\layerWeightConvDice-\layerWeightOptimal}_2\right) &\le \expect\left(\norm{\layerWeightConvReg-\layerWeightOptimal}_2\right).
        \end{align}
    \end{lemma}
        We refer to \cref{sec:supp_proof_lemma_3} for the proof.
        Beyond noise deviation threshold $\normalSigCr\!=\!\max \left(\normalSigTh, \frac{\sqrt{2}}{\length}\normalErfInv(\length^2)\right)$, the convergence gap between \dice and regression losses widens as the object size $\length$ increases. 
        \cref{fig:conv_analysis} depicts the superior convergence of \dice loss compared to regression losses under increasing noise deviation $\normalSig$ pictorially. 
        Taking the car category with $\length\!=\!4m$ and the trailer category with $\length\!=\!12m$ as examples, the noise threshold $\normalSigCr$, beyond which \dice loss exhibits better convergence, are $\normalSigCr\!=\!0.3m$ and $\normalSigCr\!=\!0.1m$ respectively.
        Combining these lemmas, we finally derive:\\
        \vspace{-0.6cm}
        \begin{theorem}\label{th:1}
            \textbf{\Dice model has better \apThreeD.}
            Assume the object length $\length$ is a constant and depth is the only source of error for detection.
            Based on \cref{lemma:1}, if $\normalSigTh$ is the solution of the equation $\normalVar\!=\!\frac{1}{\length^2}\normalErf\left(\frac{\length}{\sqrt{2}\normalSig}\right)$ and the noise deviation $\normalSig\!\ge\!\normalSigCr\!=\!\max \left(\normalSigTh, \frac{\sqrt{2}}{\length}\normalErfInv(\length^2)\right)$, then the Average Precision (\apThreeD) of the \dice model is better than \apThreeD from $\lOne$ or $\lTwo$ model.
        \end{theorem}
        We refer to \cref{sec:supp_proof_theorem_1} and \cref{tab:assumption_comp} for the proof and assumption comparisons respectively.

    %============================================================================
    %============================================================================
    \subsection{Discussions}

        %============================================================================
        \noIndentHeading{Comparing classification and regression losses.}
            We now explain how we compare classification (\dice) and regression losses.
            Our analysis assumes one-class classification in \bev segmentation with perfect predicted foreground scores $P(\varZ)=1$ (\cref{fig:problem_setup}\textcolor{red}{c}).
            Hence, \dice analysis focuses on object localization along the \bev ray  (\cref{fig:problem_setup}\textcolor{red}{b}) instead of classification probabilities thus allowing comparison of \dice and regression losses.
            \cref{lemma:1} links these losses by comparing the deviation of learned and optimal weights. 
    
        %============================================================================
        \noIndentHeading{Regression losses work better than \dice loss for regression tasks?}
            Our key message is NOT always! 
            We mathematically and empirically show that regression losses work better only when the \textbf{noise $\normalSig$ is less} in \cref{fig:conv_analysis}.

    %============================================================================       
    %============================================================================
    \subsection{\methodName Pipeline}\label{sec:pipeline}
        
        %============================================================================
        \noIndentHeading{Architecture.} 
            Based on theoretical insights of \cref{th:1}, we propose \methodName, a novel pipeline, in \cref{fig:pipeline}. 
            To effectively involve the \dice loss which originally designed for segmentation task to assist \monoThreeD, \methodName treats \bev segmentation of foreground objects and \monoThreeD head sequentially. 
            Although \bev segmentation map provides depth information (hardest \cite{kumar2022deviant,ma2021delving} \monoThreeD parameter), it lacks elevation and height information for \monoThreeD task.
            To address this, \methodName concatenates \bev features with predicted \bev segmentation (\cref{fig:pipeline}), and feeds them into the detection head to predict \threeD boxes in a $7$-DoF representation: \bev \twoD position, elevation, \threeD dimension, and yaw. 
            Unlike most works \cite{zhang2022beverse, li2022bevformer} that treat segmentation and detection branches in parallel, the sequential design
            directly utilizes refined \bev localization information to enhance \monoThreeD. 
            Ablations in \cref{sec:ablation} validate this design choice.
            We defer the details of baselines to \cref{sec:experiments}.
            Notably, our foreground \bev segmentation supervision with \dice loss does not require dense \bev segmentation maps, as we efficiently prepare them from GT \threeD boxes.

        %============================================================================
        \noIndentHeading{Training Protocol.} 
            \methodName trains the \bev segmentation head first, employing the \dice loss between the predicted and the GT \bev semantic segmentation maps, which fully utilizes the \dice loss's noise-robustness and superior convergence in localizing large objects. 
            In the second stage, we jointly fine-tune the \bev segmentation head and the \monoThreeD head. 
            We validate the effectiveness of training protocol via the ablation in \cref{sec:ablation}.

%============================================================================
%============================================================================
%============================================================================
\section{Experiments}\label{sec:experiments}

\begin{figure}[!t]
            \centering
            \includegraphics[width=\linewidth]{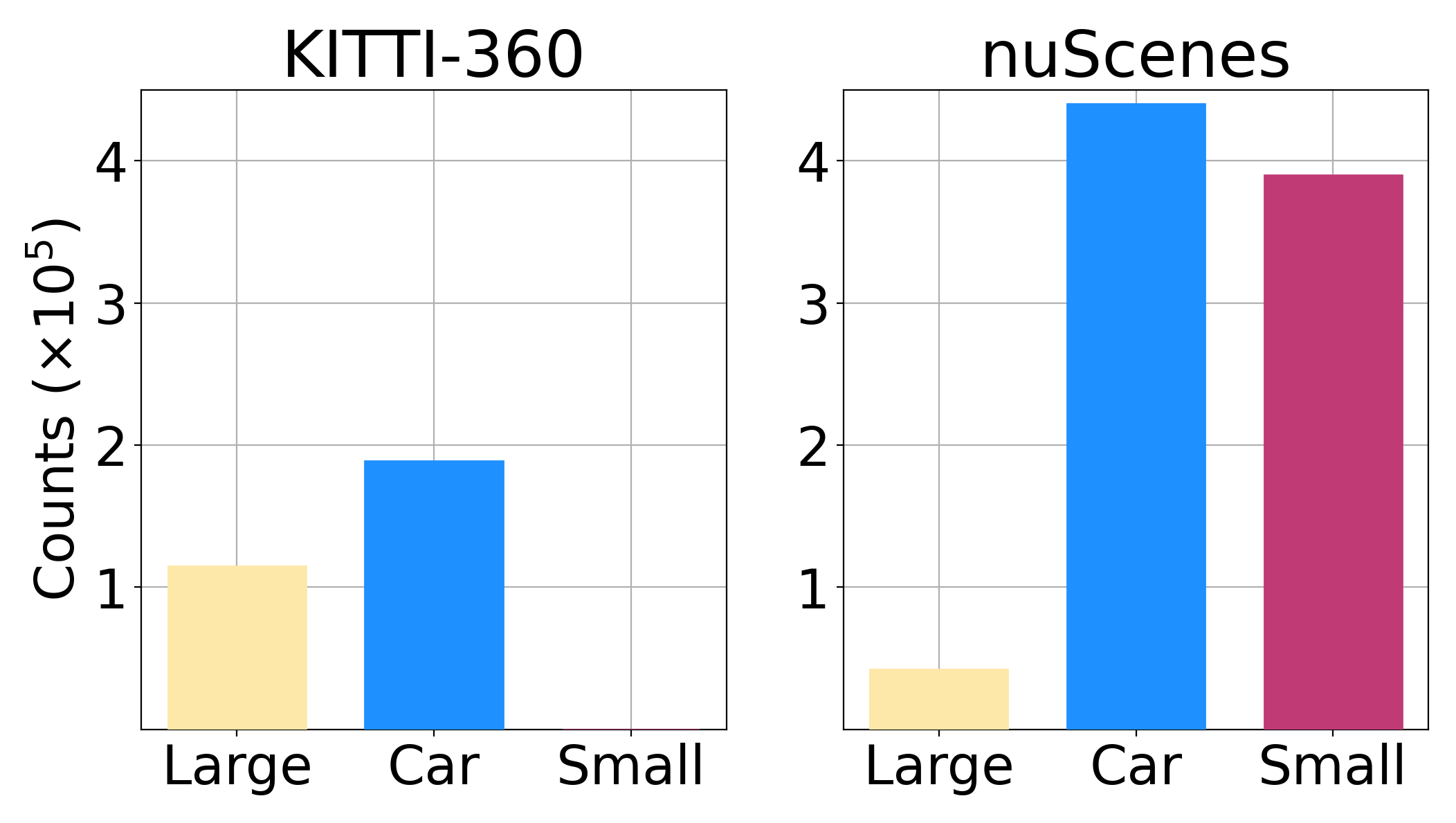}
            \caption{
                \textbf{Skewness in datasets.} 
                The ratio of \textcolor{rayShade}{large} objects to other objects is approximately $1\!:\!2$ in \kittiThreeSixty \cite{liao2022kitti360}, while the skewness is about $1\!:\!21$ in \nuscenes \cite{caesar2020nuscenes}.
            }
            \label{fig:skew}
            \vspace{-0.5cm}
        \end{figure}

    %============================================================================
    \noIndentHeading{Datasets.}
        Our experiments utilize two datasets with large objects: \kittiThreeSixty \cite{liao2022kitti360} and \nuscenes \cite{caesar2020nuscenes} encompassing both single-camera and multi-camera configurations.
        We opt for \kittiThreeSixty instead of \kitti \cite{geiger2012we} for four reasons: 
        1) \kittiThreeSixty includes large objects, while \kitti does not; 
        2) \kittiThreeSixty exhibits a balanced distribution of large objects and cars; 
        3) an extended version, \kittiThreeSixtyPanoptic \cite{gosala2022bev}, includes \bev segmentation GT for ablation studies, while \kitti \threeD detection and the \semanticKITTI dataset \cite{behley2019semantickitti} do not overlap in sequences; 
        4) \kittiThreeSixty contains about $10\times$ more images than \kitti.
        We compare these datasets in \cref{tab:dataset_comparison} and show their skewness in \cref{fig:skew}.

        \begin{table}[!t]
            \caption{\textbf{Datasets comparison.} We use \kittiThreeSixty and \nuscenes datasets for our experiments. See \cref{fig:skew} for the skewness.}
            \vspace{-2mm}
            \label{tab:dataset_comparison}
            \centering
            \scalebox{0.8}{
            \setlength\tabcolsep{0.1cm}
            \begin{tabular}{l m c c c c}
                & \kitti\!\cite{geiger2012we} & \waymo\!\cite{sun2020scalability} & \kittiThreeSixty\!\cite{liao2022kitti360} & \nuscenes\!\cite{caesar2020nuscenes}\\
                \myTopRule
                Large objects & \xmark & \xmark & \cmark & \cmark \\
                Balanced &	\xmark & \xmark & \cmark & \xmark\\
                \bev Seg. GT & \xmark & \cmark & \cmark & \cmark\\ 	
                \#images (k) & $4$ & $52$~\cite{kumar2022deviant} & $49$ & $168$ \\
            \end{tabular}
            }
            \vspace{-0.5cm}
        \end{table}

    %============================================================================
    \noIndentHeading{Data Splits.}
        We use the following splits of the two datasets:
        \begin{itemize}
            \item \textit{\kittiThreeSixty Test split}: This benchmark \cite{liao2022kitti360} contains $300$ training and $42$ testing windows. 
            These windows contain $61{,}056$ training and $910$ testing images.

            \item \textit{\kittiThreeSixty \val split}: It partitions the official train into $239$ train and $61$ validation windows \cite{liao2022kitti360}. 
            This split contains $48{,}648$ training and $1{,}294$ validation images.

            \item \textit{\nuscenes Test split}: It has $34{,}149$ training and $6{,}006$ testing samples \cite{caesar2020nuscenes} from the six cameras. This split contains $204{,}894$ training and $36{,}036$ testing images.
            
            \item \textit{\nuscenes \val split}: It has $28{,}130$ training and $6{,}019$ validation samples \cite{caesar2020nuscenes} from the six cameras. This split contains $168{,}780$ training and $36{,}114$ validation images.
        \end{itemize}

    %============================================================================
    \noIndentHeading{Evaluation Metrics.}
        We use the following metrics:
        \begin{itemize}
            \item \textit{Detection}: \kittiThreeSixty uses the mean \apThreeDFifty percentage across categories to benchmark models \cite{liao2022kitti360}.
                \nuscenes \cite{caesar2020nuscenes} uses the \nuscenes Detection Score (\NDS) as the metric. \NDS is the weighted average of mean \ap (\MAP) and five TP metrics.
                We also report \MAP over large categories (truck, bus, trailers and construction vehicles), cars, and small categories (pedestrians, motorcyle, bicycle, cone and barrier) as \MAPLarge, \MAPCar and \MAPSmall respectively.
            \item \textit{Semantic Segmentation}: We report mean \iou over foreground and all categories at $200\!\times\!200$ resolution \cite{saha2022translating,zhang2022beverse}. 
        \end{itemize}

        \begin{table}[!t]
            \caption{\textbf{\kittiThreeSixty Test detection results.} 
            \methodName pipelines outperform all monocular baselines, and also outperform old \lidar baselines.
            Click for the \href{https://www.cvlibs.net/datasets/kitti-360/leaderboard_scene_understanding.php?task=box3d}{\kittiThreeSixty leaderboard} as well as our 
            \href{https://www.cvlibs.net/datasets/kitti-360/eval_bbox_detect_detail.php?benchmark=bbox3d&result=2c29dba83ec92b4efa4b9bf67d9dcae2bef57828}{\panopticBEVWithMethod}
            and 
            \href{https://www.cvlibs.net/datasets/kitti-360/eval_bbox_detect_detail.php?benchmark=bbox3d&result=7f8612f009cc35fbebe749a345b5e49158f1efa0}{\imageToMapsWithMethod} entries.
            [Key: \firstKey{Best}, \secondKey{Second Best}, L= \lidar, C= Camera, \retrained= Retrained].
            }
            \vspace{-1mm}
            \label{tab:det_results_kitti_360_test}
            \centering
            \scalebox{\scaleFraction}{
            \setlength\tabcolsep{0.04cm}
            \begin{tabular}{cc m l | c m c | c}
                \multicolumn{2}{cm}{Modality} & \multirow{2}{*}{Method} & \multirow{2}{*}{Venue} & \apThreeDFifty (\uparrowRHDSmall) & \apThreeDTwentyFive (\uparrowRHDSmall)\\ 
                L & C & & & \MAP~\bracketPercentage & \MAP~\bracketPercentage\\
                \myTopRule
                \checkmark & & \voteNet\cite{qi2019deep}~~~~~& ICCV19 & $3.40$ & $30.61$\\
                \checkmark & & \lidarBoxNet\cite{qi2019deep} & ICCV19 & $4.08$ & $23.59$\\
                \hline
                & \checkmark &GrooMeD~\retrained\cite{kumar2021groomed} & CVPR21 & $0.17$ & $16.12$\\
                & \checkmark &\monodle\retrained\cite{ma2021delving} & CVPR21 & $0.85$ & $28.99$\\
                & \checkmark &\gupNet\retrained\cite{lu2021geometry} & ICCV21 & $0.87$ & $27.25$\\
                & \checkmark &\deviant\retrained\cite{kumar2022deviant} & ECCV22 & $0.88$ & $26.96$\\
                & \checkmark &\cubeRCNN\retrained \cite{brazil2023omni3d} & CVPR23 & $0.80$ & $15.57$\\
                & \checkmark &\monodetr\retrained\cite{zhang2023monodetr} & ICCV23 & $0.79$ & $27.13$\\
                \rowcolor{my_gray}& \checkmark & \textbf{\imageToMapsWithMethod} & CVPR24 & \second{3.14} & \second{35.04}\\
                \rowcolor{my_gray}& \checkmark & \textbf{\panopticBEVWithMethod} & CVPR24 & \first{4.64} & \first{37.12} \\ 
            \end{tabular}
            }
            \vspace{-0.4cm}
        \end{table}

    %============================================================================
    \noIndentHeading{\kittiThreeSixty Baselines and \methodName Implementation.}
        Our evaluation on the \kittiThreeSixty focuses on the detectors taking single-camera image as input. 
        We evaluate \methodName pipelines against six \sota frontal detectors: \groomedNMS \cite{kumar2021groomed}, \monodle\cite{ma2021delving}, \gupNet \cite{lu2021geometry}, \deviant \cite{kumar2022deviant}, \cubeRCNN \cite{brazil2023omni3d} and \monodetr \cite{zhang2023monodetr}.
        The choice of these models encompasses anchor \cite{kumar2021groomed, brazil2023omni3d} and anchor-free methods \cite{ma2021delving,kumar2022deviant},
        CNN~\cite{ma2021delving,lu2021geometry}, group CNN~\cite{kumar2022deviant} and transformer-based \cite{zhang2023monodetr} architectures.
        Further, \monodle normalizes loss with GT box dimensions.
        
        Due to \methodName's \bev-based approach, we do not integrate it with these frontal view detectors. 
        Instead, we extend two \sota image-to-\bev segmentation methods, \imageToMapsLong (\imageToMaps) \cite{saha2022translating} and \panopticBEVLong (\panopticBEV) \cite{gosala2022bev} with \methodName. 
        Since both \bev segmentors already include their own implementations of the image encoder, the image-to-\bev transform, and the segmentation head, implementing the \methodName pipeline  only involves adding a detection head, which we chose to be \orBoxNet \cite{yi2021oriented}.
        \methodName extensions employ \dice loss for \bev segmentation, $\smoothLOne$ losses \cite{girshick2015fast} in the \bev space to supervise the \bev \twoD position, elevation, and \threeD dimension, and cross entropy loss to supervise orientation. 

    %============================================================================
    \noIndentHeading{\nuscenes Baselines and \methodName Implementation.}
        We integrate \methodName into two prototypical \bev-based detectors, \beVerse \cite{zhang2022beverse} and \hop \cite{zong2023hop} to prove the effectiveness of \methodName. 
        Our choice of these models encompasses both transformer and convolutional backbones, multi-head and single-head architectures, shorter and longer frame history, and non-query and query-based detectors. 
        This comprehensively allows us to assess \methodName's impact on large object detection.
        \beVerse employs a multi-head architecture with a transformer backbone and shorter frame history.
        \hop is single-head query-based \sota model utilizing \bevDetFourD \cite{huang2022bevdet4d} with CNN backbone, and longer frame history. 
    
        \beVerse \cite{zhang2022beverse} includes its own implementation of detection head and \bev segmentation head in parallel. 
        We reorganize the two heads to follow our sequential design and adhere to our training protocol for network training. 
        Since \hop \cite{zong2023hop} lacks a \bev segmentation head, we incorporate the one from \beVerse into this \hop extension with \methodName.

    %============================================================================
    %============================================================================
    \subsection{\kittiThreeSixty \monoThreeD}\label{sec:detection_results_kitti_360_val}
        
        %============================================================================
        \noIndentHeading{\kittiThreeSixty Test.}\label{sec:detection_results_kitti_360_test}
            \cref{tab:det_results_kitti_360_test} presents \kittiThreeSixty leaderboard results, demonstrating the superior performance of both 
            \methodName pipelines compared to all monocular baselines across all metrics. 
            Moreover, \panopticBEVWithMethod also outperforms both legacy \lidar baselines on all metrics, while \imageToMapsWithMethod surpasses them on the \apThreeDTwentyFive metric.
            
        %============================================================================
        \noIndentHeading{\kittiThreeSixty \val.}
            \cref{tab:det_seg_results_kitti_360_val} presents the results on \kittiThreeSixty \val split, reporting the \textbf{median} model over three different seeds with the model being the final checkpoint as \cite{kumar2022deviant}. 
            \methodName pipelines outperform all monocular baselines on all but one metric, similar to \cref{tab:det_results_kitti_360_test} results. 
            Due to the \dice loss in \methodName, the biggest improvement shows up on larger objects.
            \cref{tab:det_seg_results_kitti_360_val} also includes the upper-bound oracle, where we train the \orBoxNet with the GT \bev segmentation maps.

        \begin{figure}[!t]
            \centering
            \begin{subfigure}{.485\linewidth}
                \centering
                \includegraphics[width=\linewidth]{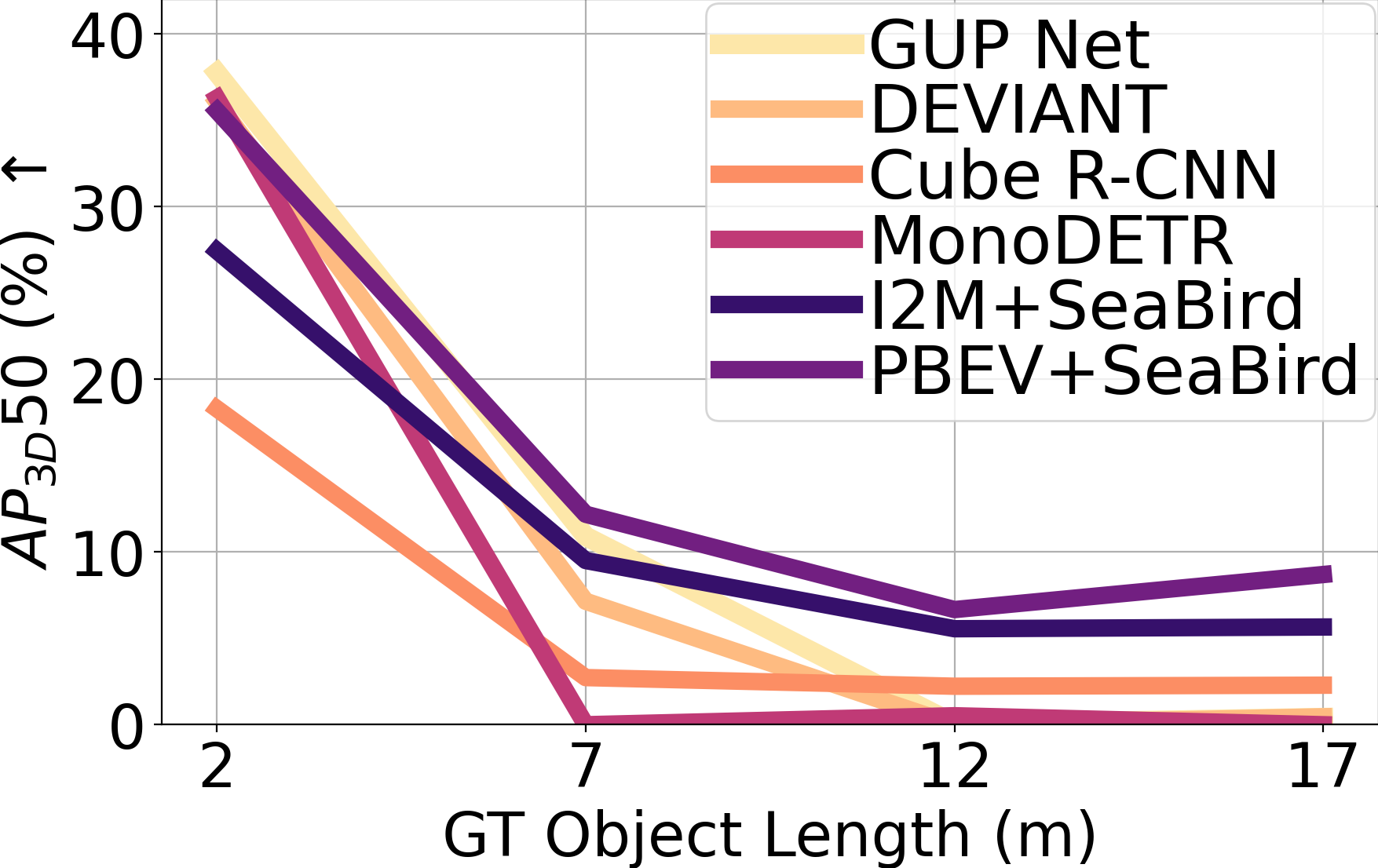}
                \caption{\apThreeDFifty comparison.}
            \end{subfigure}%
            \hfill
            \begin{subfigure}{.485\linewidth}
                \centering
                \includegraphics[width=\linewidth]{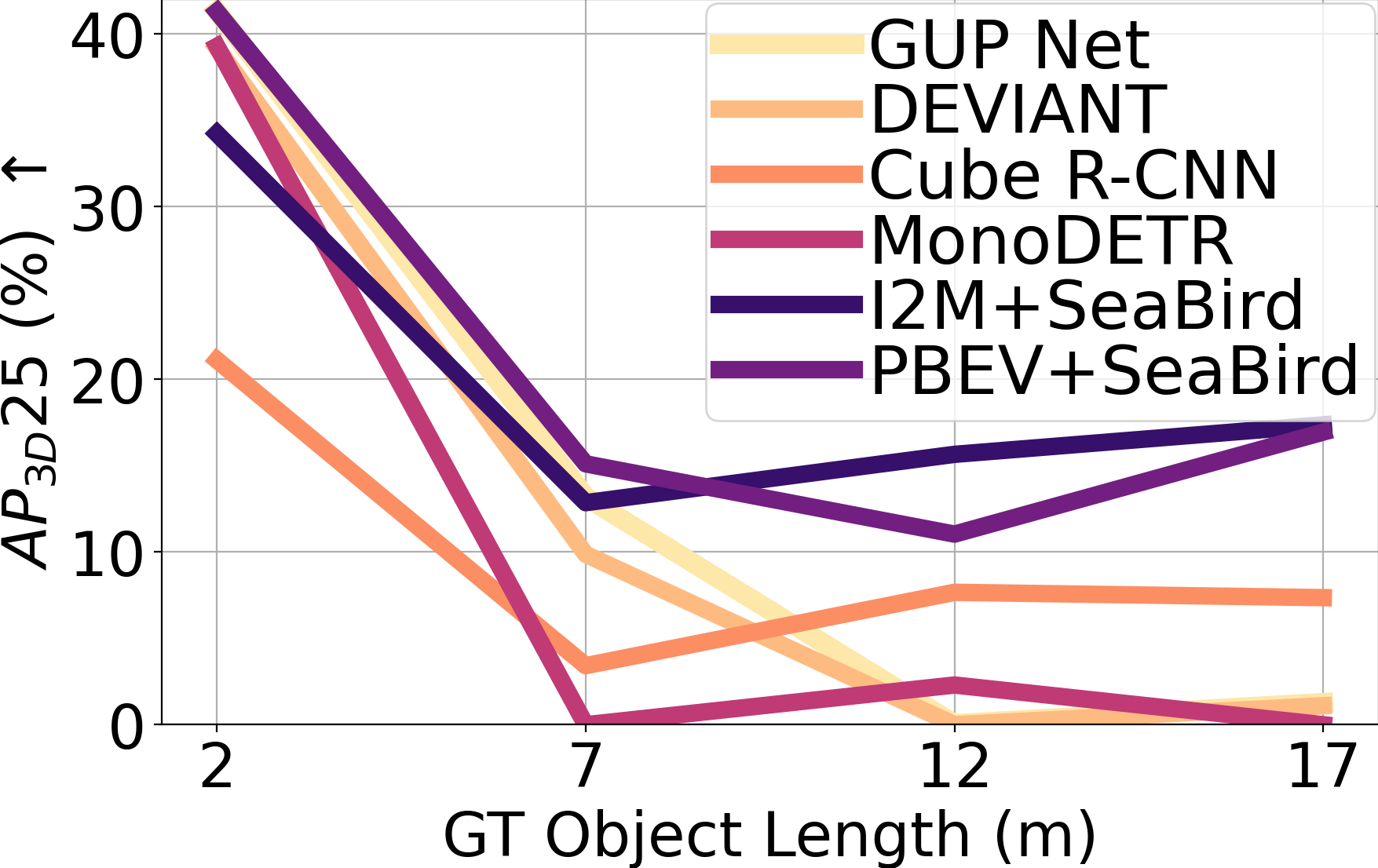}
                \caption{\apThreeDTwentyFive comparison.}
            \end{subfigure}
            \vspace{-0.2cm}
            \caption{\textbf{Lengthwise \ap Analysis} of four \sota detectors of \cref{tab:det_seg_results_kitti_360_val} and two \methodName pipelines on \kittiThreeSixty \val split. 
            \methodName pipelines outperform all baselines on large objects with over $10$m in length.}
            \label{fig:lengthwise_analysis}
            \vspace{-0.5cm}
        \end{figure}
        \begin{table*}[!t]
            \caption{\textbf{\kittiThreeSixty \val detection and segmentation results.}
            \methodName pipelines outperform all frontal monocular baselines, particularly for large objects.
            \Dice loss in \methodName also improves the \bev only (w/o \dice) version of \methodName pipelines.
            \imageToMaps and \panopticBEV are \bev segmentors. 
            So, we do not report their \monoThreeD performance.
            [Key: \firstKey{Best}, \secondKey{Second Best}, \retrained= Retrained]
            }
            \label{tab:det_seg_results_kitti_360_val}
            \vspace{-0.2cm}
            \centering
            \scalebox{\scaleFraction}{
            \setlength\tabcolsep{0.15cm}
            \begin{tabular}{c m l | c c m bcc | bcc m ccc}
                \addlinespace[0.01cm]
                \multirow{2}{*}{View} & \multirow{2}{*}{Method} & \bev Seg & \multirow{2}{*}{Venue} &\multicolumn{3}{c|}{\apThreeDFifty~\bracketPercentage (\uparrowRHDSmall)} & \multicolumn{3}{cm}{\apThreeDTwentyFive~\bracketPercentage (\uparrowRHDSmall)} & \multicolumn{3}{c}{\bev Seg \iou~\bracketPercentage (\uparrowRHDSmall)}\\ 
                && Loss & & \MAPLarge & \MAPCar & \MAP & \MAPLarge & \MAPCar & \MAP & Large & Car & \meanFor \\
                \myTopRule
                \multirow{6}{*}{Frontal} & \groomedNMS\retrained\cite{kumar2021groomed} & \multirow{6}{*}{\mathDash} & CVPR21 & $0.00$	& $33.04$ & $16.52$ & $0.00$ & $38.21$ & $19.11$ & \mathDash	& \mathDash	& \mathDash\\
                & \monodle\retrained\cite{ma2021delving} &  & CVPR21 & $0.94$ & $44.81$ & $22.88$ & $4.64$ & $50.52$ & $27.58$ & \mathDash	& \mathDash	& \mathDash\\
                & \gupNet\retrained \cite{lu2021geometry} & & ICCV21 &
                ${0.54}$ & \second{45.11} & ${22.83}$ & $0.98$ & ${50.52}$ & ${25.75}$ & \mathDash	& \mathDash	& \mathDash	\\
                & \deviant\retrained \cite{kumar2022deviant} & & ECCV22 &
                $0.53$	& $44.25$ & $22.39$ & ${1.01}$ & $48.57$ & $24.79$ & \mathDash	& \mathDash	& \mathDash\\
                & \cubeRCNN\retrained \cite{brazil2023omni3d} & & CVPR23 & $0.75$ & $22.52$ & $11.63$ & $5.55$ & $27.12$ & $16.34$ & \mathDash	& \mathDash	& \mathDash\\
                & \monodetr\retrained\cite{zhang2023monodetr} & & ICCV23 & $0.81$ & $43.24$ & $22.02$ & $4.50$ & $48.69$ & $26.60$ & \mathDash	& \mathDash	& \mathDash\\
                \hline
                \multirow{7}{*}{\bev} & \imageToMaps\retrained \cite{saha2022translating} & \Dice & ICRA22 & \mathDash & \mathDash & \mathDash & \mathDash & \mathDash & \mathDash & $20.46$ & $38.04$	& $29.25$ \\
                & \imageToMapsWithMethod & \xmark & CVPR24 & $4.86$ & $45.09$ & $24.98$ & $26.33$ & $52.31$ & $39.32$ & $0.00$ & $7.07$ & $3.54$\\
                & \cellcolor{my_gray}\textbf{\imageToMapsWithMethod} & \cellcolor{my_gray}\Dice & \cellcolor{my_gray}CVPR24 &
                \cellcolor{my_gray}\second{8.71}	& \cellcolor{my_gray}$43.19$ & \cellcolor{my_gray}$25.95$	& \cellcolor{my_gray}\second{35.76}	& \cellcolor{my_gray}$52.22$ & \cellcolor{my_gray}\second{43.99} & \cellcolor{my_gray}${23.23}$ & \cellcolor{my_gray}${39.61}$	& \cellcolor{my_gray}${31.42}$	\\
                \hhline{|~|------------|}
                & \panopticBEV\retrained \cite{gosala2022bev} & CE & RAL22 & \mathDash & \mathDash & \mathDash & \mathDash & \mathDash & \mathDash & \second{23.83} & \first{48.54}	& \first{36.18} \\
                & \panopticBEVWithMethod & \xmark & CVPR24 & $7.64$ & \first{45.37} & \second{26.51} & $29.72$ & \first{53.86} & $41.79$ & $2.07$ & $1.47$	& $1.57$\\ 
                & \cellcolor{my_gray}\textbf{\panopticBEVWithMethod} & \cellcolor{my_gray}Dice & \cellcolor{my_gray}CVPR24 &
                \cellcolor{my_gray}\first{13.22}	& \cellcolor{my_gray}$42.46$ & \cellcolor{my_gray}\first{27.84}	& \cellcolor{my_gray}\first{37.15}	& \cellcolor{my_gray}\second{52.53}   & \cellcolor{my_gray}\first{44.84} & \cellcolor{my_gray}\first{24.30}	& \cellcolor{my_gray}\second{48.04}	& \cellcolor{my_gray}\second{36.17}	\\
                \hhline{|~|------------|}
                & Oracle (GT \bev) & & \mathDash &
                $26.77$ & $51.79$ & $39.28$ & $49.74$ & $56.62$ & $53.18$ & $100.00$ & $100.00$	& $100.00$	\\
            \end{tabular}
            }
        \end{table*}

        %============================================================================
        \noIndentHeading{Lengthwise \ap Analysis.}
            \cref{th:1} states that training a model with \dice loss should lead to lower errors and, consequently, a better detector for large objects.
            To validate this claim, we analyze the detection performance with \apThreeDFifty and \apThreeDTwentyFive metrics against the object's lengths.
            For this analysis, we divide objects into four bins based on their GT object length (max of sizes): $[0, 5), [5,10), [10, 15), [15+ m$.
            \cref{fig:lengthwise_analysis} shows that \methodName pipelines excel for large objects, where the baselines' performance drops significantly.

        %============================================================================
        \noIndentHeading{\bev Semantic Segmentation.} \label{sec:segmentation_results_kitti_360_val}
            \cref{tab:det_seg_results_kitti_360_val} also presents the \bev semantic segmentation results on the \kittiThreeSixty \val split.
            \methodName pipelines outperforms the baseline \imageToMaps \cite{saha2022translating}, and achieve similar performance to \panopticBEV \cite{gosala2022bev} in \bev segmentation. 
            We retrain all \bev segmentation models only on foreground detection categories for a fair comparison.

        \begin{table*}[!t]
            \caption{\textbf{Ablation studies} on \kittiThreeSixty \val.
            [Key: \bestKey{Best}, \secondKey{Second Best}]
            }
            \label{tab:ablation}
            \vspace{-0.3cm}
            \centering
            \scalebox{\scaleFraction}{
            \setlength\tabcolsep{0.15cm}
            \begin{tabular}{l | l m bcc | bcc m cccc}
                \addlinespace[0.01cm]
                \multirow{2}{*}{Changed} & \multirow{2}{*}{From $\rightarrowRHD$ To} & \multicolumn{3}{c|}{\apThreeDFifty~\bracketPercentage (\uparrowRHDSmall)} & \multicolumn{3}{cm}{\apThreeDTwentyFive~\bracketPercentage (\uparrowRHDSmall)} & \multicolumn{4}{c}{\bev Seg \iou~\bracketPercentage (\uparrowRHDSmall)}\\ 
                && \MAPLarge & \MAPCar & \MAP & \MAPLarge & \MAPCar & \MAP & Large & Car & \meanFor & \meanEleven\\
                \myTopRule
                \multirow{4}{*}{Segmentation Loss} & \Dice$\rightarrowRHD$No Loss &$4.86$ & \best{45.09} & $24.98$ & $26.33$ & $52.31$ & $39.32$ & $0.00$ & $7.07$ & $3.54$ & \mathDash \\
                & \Dice$\rightarrowRHD$Smooth~$\lOne$& $7.63$ & $36.69$ & $22.16$ & $31.01$ & $47.51$ & $39.26$ & $17.16	$ & $34.67$ & $25.92$ & \mathDash\\
                & \Dice$\rightarrowRHD$MSE & $7.04$ & $35.59$ & $21.32$ & $30.90$ & $44.71$ & $37.81$ & $17.46$ & $34.85$ & $26.16$ & \mathDash\\
                & \Dice$\rightarrowRHD$CE & $7.06$ & $35.60$ & $21.33$ & \second{33.22} & $47.60$ & $40.41$ & $21.83$ & $38.11$ & $29.97$ & \mathDash\\
                \hline
                Segmentation Head & Yes$\rightarrowRHD$No & $7.52$ & $39.24$ & $23.38$ & $31.83$ & $47.88$ & $39.86$ & \mathDash & \mathDash & \mathDash & \mathDash \\
                Detection Head & Yes$\rightarrowRHD$No & \mathDash & \mathDash & \mathDash & \mathDash & \mathDash & \mathDash & $20.46$ & $38.04$	& $29.25$ & \mathDash \\
                \hline
                \multirow{2}{*}{Semantic Category} & For.$\rightarrowRHD$All & $1.61$ & \second{44.12} & $22.87$ & $15.36$ & $51.76$ & $33.56$ & $19.26$ & $34.46$ & $26.86$ & \first{24.34}\\
                & For.$\rightarrowRHD$Car & $4.17$ & $43.01$ & $23.59$ & $22.68$ & $51.58$ & $37.13$ & \mathDash & $40.28$ & $20.14$ & \mathDash	\\
                \hline
                Multi-head Arch.  & Sequential$\rightarrowRHD$Parallel & \best{9.12} & $40.27$ & $24.69$ & $32.45$ & $51.55$ & $42.00$ & $22.19$ & \second{40.37} & $31.28$ & \mathDash\\
                \hline 
                \bev Shortcut & Yes$\rightarrowRHD$No & $6.53$ & $38.12$ & $22.33$ & $32.05$ & \second{52.62} & \second{42.34} & \second{23.00} & \best{40.39} & \best{31.70}		& \mathDash\\
                \hline
                \multirow{2}{*}{Training Protocol} & S+J$\rightarrowRHD$J \cite{zhang2022beverse}& $7.42$ & $	42.73$ & \second{25.08} & $31.94$ & $49.88$ & $40.91$ & $22.91$ & $39.66$ & $31.29$	& \mathDash\\
                & S+J$\rightarrowRHD$D+J \cite{yang2023lidar}& $6.07$ & $43.43$ & $24.75$ & $29.24$ & \best{52.96} & $41.10$ & $20.71$ & $35.68$ & $28.20$ & \mathDash\\
                \hline
                \textbf{\imageToMapsWithMethod} & \mathDash & 
                \second{8.71}	& $43.19$ & \best{25.95}	& \best{35.76}	& $52.22$ & \best{43.99} & \best{23.23}	& $39.61$ & \second{31.42}	& \mathDash	\\
            \end{tabular}
            }
            \vspace{-0.4cm}
        \end{table*}

    %============================================================================
    %============================================================================
    \subsection{Ablation Studies on \kittiThreeSixty \val}\label{sec:ablation}

        \cref{tab:ablation} ablates \imageToMaps \cite{saha2022translating} +\methodName on the \kittiThreeSixty \val split, following the experimental settings of \cref{sec:detection_results_kitti_360_val}.

        %============================================================================
        \noIndentHeading{\Dice Loss.}
            \cref{tab:ablation} shows that both \dice loss and \bev representation are crucial to \monoThreeD of large objects.
            Replacing \dice loss with MSE or $\smoothLOne$ loss, or only \bev representation (w/o \dice) reduces \monoThreeD performance.

        %============================================================================
        \noIndentHeading{\monoThreeD and \bev Segmentation.}
            \cref{tab:ablation} shows that removing the segmentation head hinders \monoThreeD performance. 
            Conversely, removing detection head also diminishes the \bev segmentation performance for the segmentation model. 
            This confirms the mututal benefit of sequential \bev segmentation on foreground objects and \monoThreeD.

        %============================================================================
        \noIndentHeading{Semantic Category in \bev Segmentation.}
            We next analyze whether background categories play any role in \monoThreeD. 
            \cref{tab:ablation} shows that changing the foreground (For.) categories to foreground + background (All) does not help \monoThreeD. 
            This aligns with the observations of \cite{zhang2022beverse,xie2022m2bev,ma2022vision} that report lower performance on joint \monoThreeD and \bev segmentation with all categories.
            We believe this decrease happens because the network gets distracted while getting the background right.
            We also predict one foreground category (Car) instead of all in \bev segmentation.
            \cref{tab:ablation} shows that predicting all foreground categories in \bev segmentation is crucial for overall good \monoThreeD.% in \methodName.

        %============================================================================
        \noIndentHeading{Multi-head Architecture.}
            \methodName employs a sequential architecture (Arch.) of segmentation and detection heads instead of parallel architecture.
            \cref{tab:ablation} shows that the sequential architecture outperforms the parallel one. 
            We attribute this \monoThreeD boost to the explicit object localization provided by segmentation in the \bev plane.

        %============================================================================
        \noIndentHeading{\bev Shortcut.}
            \cref{sec:pipeline} mentions that \methodName's \monoThreeD head utilizes both the \bev segmentation map and \bev features.
            \cref{tab:ablation} demonstrates that providing \bev features to the detection head is crucial for good \monoThreeD.
            This is because the \bev map lacks elevation information, and incorporating \bev features helps estimate elevation.

        %============================================================================
        \noIndentHeading{Training Protocol.}
            \methodName trains segmentor first and then jointly trains detector and segmentor (S+J).
            We compare with direct joint training (J) of \cite{zhang2022beverse} and training detection followed by joint training (D+J) of \cite{yang2023lidar}.
            \cref{tab:ablation} shows that \methodName training protocol works best.

        \begin{table*}[!t]
            \caption{\textbf{\nuscenes Test detection results.}
            \methodName pipelines achieve the best \MAPLarge among methods without Class Balanced Guided Sampling (CBGS) \cite{zhu2019class} and future frames.
            Results are from the nuScenes leaderboard or corresponding papers on \vovNet or R101 backbones.
            [Key: 
            \firstKey{Best}, \secondKey{Second Best}, 
            S= Small, \reimplemented= Reimplementation, \cbgs= CBGS, \future= Future Frames.]
            }
            \vspace{-2mm}
            \label{tab:nuscenes_test}
            \centering
            \scalebox{\scaleFraction}{
            \setlength\tabcolsep{0.15cm}
            \begin{tabular}{l l | c c m b c c c c}
                Resolution & Method & BBone &  Venue & \MAPLarge\!(\uparrowRHDSmall) & \MAPCar\!(\uparrowRHDSmall) & \MAPSmall\!(\uparrowRHDSmall) & \MAP\!(\uparrowRHDSmall)  & \NDS\!(\uparrowRHDSmall) \\
                \myTopRule
                \multirow{6}{*}{$512\!\times\!1408$}
                & \bevDepth\cite{li2023bevdepth} in \cite{kim2023predict} & R101 & AAAI23 & \mathDash 
                & \mathDash & \mathDash & $39.6$ & $48.3$ \\%& $0.593$ & \mathDash & $0.533$ & \mathDash & \mathDash\\
                & \bevStereo\cite{li2023bevstereo} in \cite{kim2023predict} & R101 & AAAI23 & \mathDash & \mathDash & \mathDash & $40.4$ & $50.2$ \\%& $0.587$ & \mathDash & $0.518$ & \mathDash & \mathDash\\
                & P2D \cite{kim2023predict} & R101 & ICCV23 & \mathDash & \mathDash & \mathDash & $43.6$ & $53.0$ \\%& $0.550$ & \mathDash & $0.517$ & \mathDash & \mathDash\\
                & \beVerseSmall\cite{zhang2022beverse}& Swin-S  & \arxiv & $24.4$ & $60.4$ & $47.0$ & $39.3$ & $53.1$ \\%& $0.541$ & $0.247$	& $0.394$ & $0.345$	& $0.129$\\
                & \hop\reimplemented\cite{zong2023hop}  & R101 & ICCV23 & \second{36.0}    & \second{65.0}    & \second{53.9} & \second{47.9} & \first{57.5} \\
                & \cellcolor{my_gray}\textbf{\hopWithMethod} & \cellcolor{my_gray}R101 & \cellcolor{my_gray}CVPR24 & \cellcolor{my_gray}\first{36.6}    & \cellcolor{my_gray}\first{65.8}    & \cellcolor{my_gray}\first{54.7} & \cellcolor{my_gray}\first{48.6} & \cellcolor{my_gray}\second{57.0}\\
                \myTopRule
                \multirow{13}{*}{$640\!\times\!1600$}
                %& BEVDet \cite{huang2021bevdet} & \vovNet & \arxiv & $29.6$ & $61.7$ & $48.2$ & $42.1$ & $48.2$ \\%& $0.529$ & $0.236$ & $0.396$ & $0.979$ & $0.152$\\
                & SpatialDETR \cite{doll2022spatialdetr} & \vovNet & ECCV22 & $30.2$ & $61.0$ & $48.5$ & $42.5$ & $48.7$ \\%& $0.613$ & $0.253$ & $0.402$ & $0.857$ & $0.131$\\30.2	61.0	48.5
                & \threeDPPE\cite{shu2023dppe} & \vovNet & ICCV23 & \mathDash & \mathDash & \mathDash & $46.0$ & $51.4$ \\%& $0.569$ & $0.255$ & $0.394$ & $0.796$ & $0.138$\\
                & \xThreeKDAll \cite{klingner2023x3kd} & R101 & CVPR23 & \mathDash & \mathDash & \mathDash & $45.6$    & $56.1$\\
                & \petrVTwo\cite{liu2023petrv2} & \vovNet & ICCV23 & $36.4$    & $66.7$    & $55.6$    & $49.0$    & $58.2$    \\%& $0.561$    & $0.243$    & $0.361$    & $0.343$    & $0.12$\\
                & \veDet\cite{chen2023viewpoint} & \vovNet & CVPR23 & \second{37.1}    & $68.5$    & \first{57.7}    & $50.5$    & $58.5$\\ 
                & \frustumFormer\cite{wang2023frustumformer} & \vovNet & CVPR23 & \mathDash & \mathDash & \mathDash & \first{51.6}    & $58.9$\\    
                & MV2D \cite{wang2023object}  & \vovNet & ICCV23 & \mathDash & \mathDash & \mathDash & \second{51.1}    & \second{59.6}    \\%& $0.525$    & $0.243$    & $0.357$    & $0.357$    & $0.120$\\
                & \hop\reimplemented\cite{zong2023hop} & \vovNet & ICCV23 & \second{37.1} & \second{68.7} & $55.6$ & $49.4$ & $58.9$ \\
                & \cellcolor{my_gray}\textbf{\hopWithMethod} & \cellcolor{my_gray}\vovNet & \cellcolor{my_gray}CVPR24 & \cellcolor{my_gray}\first{38.4}    & \cellcolor{my_gray}\first{70.2} & \cellcolor{my_gray}\second{57.4}    & \cellcolor{my_gray}\second{51.1}    & \cellcolor{my_gray}\first{59.7}    \\
                \hhline{|~|--------|}
                %& \bevDetFourD\cbgs\cite{huang2022bevdet4d} & Swin-B & \arxiv & $30.7$ & $65.0$ & $52.6$ & $45.1$ & $56.9$ \\%& $0.511$ & $0.241$ & $0.386$ & $0.301$ & $0.121$\\
                & \saBEV\cbgs\cite{zhang2023sabev} & \vovNet & ICCV23 & $40.5$ & $68.9$ & $60.5$ & $53.3$ & $62.4$ \\%& $0.430$ & $0.241$ & $0.338$ & $0.282$ & $0.139$\\
                & \fbBEV\cbgs\cite{li2023fbbev} & \vovNet & ICCV23 & $39.3$    & $71.7$    & $61.6$   & $53.7$ & $62.4$\\
                & \cape\cbgs\cite{xiong2023cape} & \vovNet & CVPR23 & $41.3$    & $71.4$    & $63.3$    & $55.3$    & $62.8$\\
                & \sparseBEV\future\cite{liu2023sparsebev} & \vovNet & ICCV23 & $45.6$    & $76.3$    & $68.8$    & $60.3$    & $67.5$    \\%& $0.425$    & $0.239$    & $0.311$    & $0.172$    & $0.116$\\
                \myTopRule
                \multirow{5}{*}{$900\!\times\!1600$} 
                & \parametricBEV\!\cite{yang2023parametric} & R101 & ICCV23 & \mathDash & \mathDash & \mathDash & $46.8$ & $49.5$ \\
                % & \petr \cite{liu2022petr} & \vovNet & ECCV22 & $32.5$ & $63.1$ & $49.6$ & $44.1$ & $50.4$ \\%& $0.593$ & $0.249$ & $0.383$ & $0.808$ & $0.132$\\
                & \uvtr \cite{li2022unifying} & R101 & NeurIPS22 & $35.1$ & $67.3$ & $52.9$ & $47.2$ & $55.1$ \\
                & \bevFormer\cite{li2022bevformer} & \vovNet & ECCV22 & $34.4$ & $67.7$ & $55.2$ &
                $48.9$ & $56.9$ \\%& $0.582$ & $0.256$ & $0.375$ & $0.378$ & $0.126$\\
                & \polarFormer\cite{jiang2023polarformer} & \vovNet & AAAI23 & $36.8$ & $68.4$ & $55.5$ & $49.3$ & $57.2$ \\%& $0.556$ & $0.256$ & $0.364$ & $0.440$ & $0.127$\\
                & \stxd\cite{jang2023stxd} & \vovNet & NeurIPS23 & \mathDash & \mathDash & \mathDash & $49.7$ & $58.3$\\ 
                
            \end{tabular}
            }
        \end{table*}

        \begin{table*}[!t]
            \caption{\textbf{\nuscenes \val detection results.}
            \methodName pipelines outperform the two baselines \beVerse and \hop, particularly for large objects.
            We train all models without CBGS. 
            See \cref{tab:nuscenes_val_more} for a detailed comparison.
            [Key: 
            S= Small, T= Tiny, \released= Released, \reimplemented= Reimplementation]
            }
            \label{tab:nuscenes_val}
            \vspace{-0.2cm}
            \centering
            \scalebox{\scaleFraction}{
            \setlength\tabcolsep{0.15cm}
            \begin{tabular}{l  l | c c m a l l l l}
                Resolution & Method & BBone & Venue & \multicolumn{1}{c}{\CYMyFix\MAPLarge (\uparrowRHDSmall)} & \multicolumn{1}{c}{\MAPCar (\uparrowRHDSmall)} & \multicolumn{1}{c}{\MAPSmall (\uparrowRHDSmall)} & \multicolumn{1}{c}{\MAP (\uparrowRHDSmall)}  & \multicolumn{1}{c}{\NDS (\uparrowRHDSmall)} \\
                \myTopRule
                \multirow{4}{*}{$256\!\times\!704$}  
                & \beVerseTiny\released\cite{zhang2022beverse}& \multirow{2}{*}{Swin-T} & \arxiv & $18.5$        & $53.4$ & $38.8$           & $32.1$        & $46.6$           \\%& $43.6$        & $35.2$ \\
                & \textbf{+\methodName}         & & CVPR24 & \good{19.5}{1.0} & \good{54.2}{0.8} & \good{41.1}{2.3}    & \good{33.8}{1.5} & \good{48.1}{1.7} \\%& \good{45.1}{1.5} & \good{36.1}{0.9} \\  
                \hhline{|~|--------|}
                & \hop\released\cite{zong2023hop}  & \multirow{2}{*}{R50} & ICCV23 & $27.4$    & $57.2$    & $46.4$ & $39.9$ & $50.9$ \\
                & \textbf{+\methodName}      & & CVPR24 & \good{28.2}{0.8}    & \good{58.6}{1.4}    & \good{47.8}{1.4} & \good{41.1}{1.2} & \good{51.5}{0.6} \\
                \myTopRule
                \multirow{4}{*}{$512\!\times\!1408$} 
                &\beVerseSmall\released\cite{zhang2022beverse}& \multirow{2}{*}{Swin-S} &  \arxiv & $20.9$           & $56.2$           & $42.2$        & $35.2$        & $49.5$ \\
                & \textbf{+\methodName}                          & & CVPR24 & \good{24.6}{3.7}    & \good{58.7}{2.5}    & \good{45.0}{2.8}  & \good{38.2}{3.0} & \good{51.3}{1.8} \\
                \hhline{|~|--------|}
                & \hop\reimplemented\cite{zong2023hop}             & \multirow{2}{*}{R101} & ICCV23 & 
                $31.4$    & $63.7$    & $52.5$    & $45.2$    & $55.0$ \\ 
                & \textbf{+\methodName}                          & & CVPR24 & \good{32.9}{1.5}    & \good{65.0}{1.3}    & \good{53.1}{0.6} & \good{46.2}{1.0} & \bad{54.7}{0.3} \\
                \myTopRule
                \multirow{2}{*}{$640\!\times\!1600$}
                & \hop\reimplemented\cite{zong2023hop}             & \multirow{2}{*}{\vovNet} 
                & ICCV23 & $36.5$    & $69.1$    & $56.1$    & $49.6$    & $58.3$ \\
                & \textbf{+\methodName}                          & & CVPR24 & \good{40.3}{3.8}    & \good{71.7}{2.6}    & \good{58.8}{2.7} & \good{52.7}{3.1} & \good{60.2}{1.9} \\
            \end{tabular}
            }
            \vspace{-0.2cm}
        \end{table*}

    %============================================================================
    %============================================================================
    \subsection{\nuscenes \monoThreeD}
    
        We next benchmark \methodName on \nuscenes \cite{caesar2020nuscenes}, which encompasses more diverse object categories such as trailers, buses, cars and traffic cones, compared to \kittiThreeSixty \cite{liao2022kitti360}.

    %============================================================================
    \noIndentHeading{\nuscenes Test.}\label{sec:detection_nuscenes_test}
        \cref{tab:nuscenes_test} presents the results of incorportaing \methodName to the \hop models with the \vovNet and R101 backbones.
        \methodName with both \vovNet and R101 backbones outperform several \sota methods on the \nuscenes leaderboard, as well as the baseline \hop, on nearly every metric.
        Interestingly, \methodName pipelines also outperform several baselines which use higher resolution $(900\!\times\!1600)$ inputs. % \methodName with
        Most importantly, \methodName pipelines achieve the highest \MAPLarge performance, providing empirical support for the claims of \cref{th:1}.

    %============================================================================
    \noIndentHeading{\nuscenes \val.}\label{sec:detection_nuscenes_val}
        \cref{tab:nuscenes_val} showcases the results of integrating \methodName with \beVerse \cite{zhang2022beverse} and \hop \cite{zong2023hop} at multiple resolutions, as described in \cite{zhang2022beverse,zong2023hop}.
        \cref{tab:nuscenes_val} demonstrates that integrating \methodName consistently improves these detectors on almost every metric at multiple resolutions.
        The improvements on \MAPLarge empirically support the claims of \cref{th:1} and validate the effectiveness of \dice loss and \bev segmentation in localizing large objects.

%============================================================================
%============================================================================
%============================================================================
\section{Conclusions}\label{sec:conclusions}
    This paper highlights the understudied problem of \monoThreeD generalization to large objects.
    Our findings reveal that modern frontal detectors struggle to generalize to large objects even when trained on balanced datasets.
    To bridge this gap, we investigate the regression and dice losses, 
    %associated with \monoThreeD and \bev segmentation tasks
    examining their robustness under varying error levels and object sizes.
    We mathematically prove that the dice loss 
    outperforms regression losses in noise-robustness and model convergence for large objects for a simplified case.
    Leveraging our theoretical insights, we propose \methodName (\methodNameFull) as the first step towards generalizing to large objects.
    \methodName effectively integrates \bev segmentation with the dice loss for \monoThreeD.
    \methodName achieves \sota results on the \kittiThreeSixty leaderboard and consistently improves existing detectors on the \nuscenes leaderboard, particularly for large objects.
    We hope that this initial step towards generalization will contribute to safer AVs.

%============================================================================
%============================================================================
%============================================================================
{
    \small
    \bibliographystyle{ieeenat_fullname}
    \bibliography{references}
}

\addtocontents{toc}{\protect\setcounter{tocdepth}{3}}
%============================================================================
%============================================================================
%============================================================================
\clearpage
\maketitlesupplementary

\begingroup
\let\clearpage\relax
\hypersetup{linkcolor=blue}
\tableofcontents
\endgroup

% Start appendices sections with A1, A2 instead of 1,2
\renewcommand{\thesection}{A\arabic{section}}
\setcounter{section}{0}
% \setcounter{page}{13}

%============================================================================
%============================================================================
%============================================================================
\section{Additional Explanations and Proofs}\label{sec:additional}

    We now add some explanations and proofs which we could not put in the main paper because of the space constraints.

    %============================================================================
    %============================================================================
    \subsection{Proof of Converged Value}\label{sec:proof_converged}
        We first bound the converged value from the optimal value. 
        These results are well-known in the literature \cite{shalev2007pegasos, lacoste2012simpler}. 
        We reproduce the result from using our notations for completeness.
        \begin{align}
            &\expect\left(\norm{\layerWeightConv\!-\!\layerWeightOptimal}_2^2\right) \nonumber\\
            &= \expect\left(\norm{\layerWeightConv\!-\!\layerWeightMean + \layerWeightMean\!-\!\layerWeightOptimal}^2_2\right) \nonumber \\
            &= \expect\left(\left(\layerWeightConv\!-\!\layerWeightMean + \layerWeightMean\!-\!\layerWeightOptimal\right)^T\left(\layerWeightConv\!-\!\layerWeightMean + \layerWeightMean\!-\!\layerWeightOptimal\right)\right) \nonumber \\
            &= \expect((\layerWeightConv\!-\!\layerWeightMean)^T(\layerWeightConv\!-\!\layerWeightMean)) + \expect((\layerWeightMean\!-\!\layerWeightOptimal)^T(\layerWeightMean\!-\!\layerWeightOptimal)) \nonumber \\
            &~~~~+ 2\expect((\layerWeightConv\!-\!\layerWeightMean)^T(\layerWeightMean\!-\!\layerWeightOptimal)) \nonumber \\
            &= \var(\layerWeightConv) + \expect((\layerWeightMean\!-\!\layerWeightOptimal)^T(\layerWeightMean\!-\!\layerWeightOptimal)) 
            \label{eq:weight_bound_with_mean}
        \end{align}
        where $\layerWeightMean=\expect(\layerWeightConv)$ is the mean of the layer weight and 
         $\var(\mathbf{\weight})$ denotes the variance of $\sum_\instantTwo w_\instantTwo^2$.
    
        \noIndentHeading{SGD.}
        We begin the proof by writing the value of $\layerWeightTime$ at every step.
        The model uses SGD, and so, the weight $\layerWeightTime$ after $\instant$ gradient updates is
        \begin{align}
            \layerWeightTime &= \layerWeightZero - \step_1 {}^\loss\gradient_1 - \step_2 {}^\loss\gradient_2 - \cdots - s_\instant \gradTime,
            \label{eq:sgd_step}
        \end{align}
        where $\gradTime$ denotes the gradient of $\layerWeight$ at every step $\instant$.
        Assume the loss function under consideration $\loss$ is $\loss = f(\layerWeightTimeVanilla\image - \depthGT) = f(\noise)$. Then, we have,
        \begin{align}
            \gradTime &= \dfrac{\partial \loss}{\partial \layerWeightTimeVanilla} \nonumber \\
            &= \dfrac{\partial \loss(\layerWeightTimeVanilla\image - \depthGT)}{\partial \layerWeightTimeVanilla} \nonumber \\
            &= \dfrac{\partial \loss(\layerWeightTimeVanilla\image - \depthGT)}{\partial (\layerWeightTimeVanilla\image - \depthGT)} \dfrac{\partial (\layerWeightTimeVanilla\image - \depthGT)}{\partial \layerWeightTimeVanilla} \nonumber \\
            &= \dfrac{\partial \loss(\noise)}{\partial \noise} \image \nonumber \\
            &= \image \dfrac{\partial \loss(\noise)}{\partial \noise} \nonumber \\
            \implies \gradTime &= \image \funcNoise,
        \end{align}
        with $\funcNoise = \dfrac{\partial \loss(\noise)}{\partial \noise}$ is the gradient of the loss function wrt noise. 
        
        \noIndentHeading{Expectation and Variance of Gradient $\gradTime$}
        Since the image $\image$ and noise $\noise$ are statistically independent, the image and the noise gradient $\noise$ are also statistically independent. 
        So, the expected gradients
        \begin{align}
            \expect(\gradTime) &= \expect(\image) \expect(\funcNoise) = 0.
            \label{eq:mean_gradTime}
        \end{align}
    
        Note that if the loss function is an even function (symmetric about zero), its gradient $\funcNoise$ is an odd function (anti-symmetric about $0$), and so its mean $\expect(\funcNoise) = 0$. 
        
        Next, we write the gradient variance $\var(\gradTime)$ as
        \begin{align}
            \var(\gradTime) = \var(\image \funcNoise) 
            &= \expect(\image^T\image) \expect(\funcNoise^2) - \expect^2(\image)\expect^2(\funcNoise) \nonumber \\
            &= \expect(\image^T\image) \left[ \var(\funcNoise) + \expect^2(\funcNoise) \right] \nonumber\\
            &~~~~ - \expect^2(\image)\expect^2(\funcNoise) \nonumber \\
            \implies \var(\gradTime) &= \expect(\image^T\image)\var(\funcNoise)\quad\text{as } \expect(\funcNoise)=0
            \label{eq:var_gradTime}
        \end{align}
    
        \noIndentHeading{Expectation and Variance of Converged Weight $\layerWeightTime$}
        We first calculate the expected  converged weight as
        \begin{align}
            \expect(\layerWeightTime) &= 
            \expect(\layerWeightZero) + \left(\sum_{\instantTwo=1}^\instant \step_\instantTwo  \expect\left(\gradTimeTwo\right)  \right) ,\text{using \cref{eq:sgd_step}} \nonumber \\
            &= \mathbf{0} \quad\text{using \cref{eq:mean_gradTime}} \nonumber \\
            \implies \expect(\layerWeightConv) &= \lim_{t\rightarrow\infty} \expect(\layerWeightTime) \nonumber \\ 
            \implies \expect(\layerWeightConv) &= \layerWeightMean = \mathbf{0}
            \label{eq:mean_conv_weight}
        \end{align}
    
        We finally calculate the variance of the converged weight. Because the SGD step size is independent of the gradient, we write using \cref{eq:sgd_step}, 
        \begin{align}
            \var(\layerWeightTime) &= \var(\layerWeightZero) + \step_1^2\var\left( \gradient_1\right) + \step_2^2\var\left(\gradient_2\right) \nonumber\\
            &~~~~+ \cdots + \step_\instant^2\var\left(\gradTime\right)
        \end{align}
    
        Assuming the gradients $\gradTime$ are drawn from an identical distribution, we have
        \begin{align}
            \var(\layerWeightTime) &= \var(\layerWeightZero) + \left(\sum_{\instantTwo=1}^\instant \step_\instantTwo^2\right)  \var\left(\gradTime\right) \nonumber \\
            \implies \var(\layerWeightConv) &= \lim_{t\rightarrow\infty} \var(\layerWeightTime) \nonumber \\
           %\implies \var(\layerWeightConv)  
           &= \var(\layerWeightZero) + \left(\lim_{t\rightarrow\infty} \sum_{\instantTwo=1}^\instant \step_\instantTwo^2\right)  \var\left(\gradTime\right) \nonumber \\ 
            \implies \var(\layerWeightConv) &= \var(\layerWeightZero) + \stepSumTrue \var\left(\gradTime\right)
            \label{eq:weight_bound_2}
        \end{align}
        An example of square summable step-sizes of SGD is $\step_\instantTwo= \frac{1}{\instantTwo}$, and then the constant $\stepSumTrue = \sum\limits_{\instantTwo=1}  \step_\instantTwo^2 = \frac{\pi^2}{6}$.
        This assumption is also satisfied by modern neural networks since their training steps are always finite. 
    
        Substituting \cref{eq:var_gradTime} in \cref{eq:weight_bound_2}, we have
        \begin{align}
            \var(\layerWeightConv) &= \var(\layerWeightZero) + \stepSumTrue \expect(\image^T\image)\var(\funcNoise)
            \label{eq:var_conv_weight}
        \end{align}
        Substituting mean and variances from \cref{eq:mean_conv_weight,eq:var_conv_weight} in \cref{eq:weight_bound_with_mean}, we have
        \begin{align}
            \expect\left(\norm{\layerWeightConv\!-\!\layerWeightOptimal}_2^2\right) 
            &= \var(\layerWeightZero) + \stepSumTrue \expect(\image^T\image)  \var(\funcNoise) \nonumber \\
            &\qquad + \expect(||\layerWeightOptimal||^2) \nonumber \\
            &= \stepSumTrue \expect(\image^T\image) \var(\funcNoise) + \var(\layerWeightZero) \nonumber \\
            &\qquad + \expect(||\layerWeightOptimal||^2)  \nonumber \\
            \implies \expect\left(\norm{\layerWeightConv\!-\!\layerWeightOptimal}_2^2\right) &= \stepConstant \var(\funcNoise) + \uselessConstant,
        \end{align}
        where $\funcNoise = \dfrac{\partial \loss(\noise)}{\partial \noise}$ is the gradient of the loss function wrt noise,  and $\stepConstant = \stepSumTrue \expect(\image^T\image)$ and $\uselessConstant$ are terms independent of the loss function $\loss$.

    %============================================================================
    %============================================================================
    \subsection{Comparison of Loss Functions}

        \cref{eqn:conv:weight:dist} shows that different losses $\loss$ lead to different $\var(\funcNoise)$. 
        Hence, comparing this term for different losses asseses the quality of losses.

        %============================================================================
        \subsubsection{Gradient Variance of MAE Loss}\label{sec:supp_var_lOne}
            The result on MAE $(\lOne)$ is well-known in the literature \cite{shalev2007pegasos, lacoste2012simpler}. 
            We reproduce the result from \cite{shalev2007pegasos, lacoste2012simpler} using our notations for completeness.
            
            The $\lOne$ loss is 
            \begin{align}
                \lOne(\noise) &= |\depthPred - \depthGT|_1 = |\layerWeightTime\image - \depthGT|_1 = |\noise|_1\nonumber \\
                \implies \funcNoise &= \dfrac{\partial \lOne(\noise)}{\partial \noise} = \sign(\noise) 
            \end{align}
            Thus, $\funcNoise = \sign(\noise)$ is a Bernoulli random variable with $p(\funcNoise) = 1/2 \text{ for } \funcNoise= \pm 1$. 
            So, mean $\expect(\funcNoise) = 0$ and variance $\var(\funcNoise) = 1$. 

        %============================================================================
        \subsubsection{Gradient Variance of MSE Loss}\label{sec:supp_var_lTwo} 
            The result on MSE $(\lTwo)$ is well-known in the literature \cite{shalev2007pegasos, lacoste2012simpler}. 
            We reproduce the result from \cite{shalev2007pegasos, lacoste2012simpler} using our notations for completeness.          
            %\textbf{Case 1: Regression with $\lTwo$ Loss.}
            The $\lTwo$ loss is 
            \begin{align}
                \lTwo(\noise) &= 0.5|\depthPred - \depthGT|^2 = 0.5|\noise|^2 = 0.5\noise^2\nonumber \\
                \implies \funcNoise &= \dfrac{\partial \lTwo(\noise)}{\partial \noise} = \noise
            \end{align}
            Thus, $\funcNoise = \noise$ is a normal random variable \cite{shalev2007pegasos}.
            So, mean $\expect(\funcNoise) = 0$ and variance $\var(\funcNoise) = \var(\noise) = \normalVar$.

        %============================================================================
        \subsubsection{Gradient Variance of \Dice Loss. (Proof of \cref{lemma:2})}\label{sec:supp_var_dice}

            \begin{proof}\let\qed\relax
                We first write the gradient of \dice loss as a function of noise $(\noise)$ as follows:
                \begin{align}
                    \funcNoise &= \dfrac{\partial \lDice(\noise)}{\partial \noise} = \begin{cases}
                        \dfrac{\sign(\noise)}{\length} \text{ , }|\noise|\le \length \\
                        0 \quad~~~~~~\text{ , }|\noise|\ge \length 
                    \end{cases} 
                \end{align}
                The gradient of the loss $\funcNoise$ is an odd function and so, its mean $\expect(\funcNoise) = 0$. Next, we write its variance $\var(\funcNoise)$ as
                \begin{align}
                    \var(\funcNoise) = \var(\noise) &= \frac{1}{\length^2}\int\limits_{-\length}^\length \dfrac{1}{\sqrt{2\pi}\normalSig} e^{-\frac{\noise^2}{2\normalVar}} d\noise \nonumber \\
                    &= \frac{2}{\length^2}\int\limits_{0}^\length \dfrac{1}{\sqrt{2\pi}\normalSig} e^{-\frac{\noise^2}{2\normalVar}} d\noise \nonumber \\
                    &= \frac{2}{\length^2}\int\limits_{0}^{\length/\normalSig} \dfrac{1}{\sqrt{2\pi}} e^{-\frac{\noise^2}{2}} d\noise \nonumber \\
                    &= \frac{2}{\length^2}\left[ \int\limits_{-\infty}^{\length/\normalSig} \dfrac{1}{\sqrt{2\pi}} e^{-\frac{\noise^2}{2}} d\noise  - \frac{1}{2}\right] \nonumber \\
                    &= \frac{2}{\length^2}\left[ \normalCDF\left(\frac{\length}{\normalSig}\right) - \frac{1}{2}\right] \label{eq:dice_cdf}\\
                    &\quad\text{~~~~~~~~~~~where, } \normalCDF \text{ is the normal CDF}\nonumber 
                \end{align}
                We write the CDF $\normalCDF(x)$ in terms of error function $\normalErf$ as:
                \begin{align}
                    \normalCDF(x) &= \dfrac{1}{2} + \dfrac{1}{2}\normalErf\left(\dfrac{x}{\sqrt{2}}\right)
                \end{align}
                $\text{for } x \ge 0$.
                Next, we put $x = \dfrac{\length}{\normalSig}$ to get
                \begin{align}
                     \normalCDF\left(\frac{\length}{\normalSig}\right) &= \frac{1}{2} + \frac{1}{2}\normalErf\left(\frac{\length}{\sqrt{2}\normalSig}\right)
                \end{align}
                Substituting above in \cref{eq:dice_cdf}, we obtain
                \begin{align}
                \var(\funcNoise) &= \frac{2}{\length^2}\left[ \frac{1}{2} + \frac{1}{2}\normalErf\left(\frac{\length}{\sqrt{2}\normalSig}\right) - \frac{1}{2}\right]\nonumber \\
                \implies \var(\funcNoise) &= \frac{1}{\length^2}\normalErf\left(\frac{\length}{\sqrt{2}\normalSig}\right)
                \end{align}
            \end{proof}

    %============================================================================
    %============================================================================
    \subsection{Proof of \cref{lemma:3}}\label{sec:supp_proof_lemma_3}
    \begin{proof}\let\qed\relax
        It remains sufficient to show that 
        \begin{align}
            \expect\left(\norm{\layerWeightConvDice-\layerWeightOptimal}_2\right) &\le \expect\left(\norm{\layerWeightConvReg-\layerWeightOptimal}_2\right) \nonumber \\
            \implies \expect\left(\norm{\layerWeightConvDice-\layerWeightOptimal}_2^2\right) &\le \expect\left(\norm{\layerWeightConvReg-\layerWeightOptimal}_2^2\right) \label{eq:show}
        \end{align}
        Using \cref{lemma:1}, the above comparison is a comparison between the gradient variance of the loss wrt noise $\var(\funcNoise)$. 
        Hence, we compute the gradient variance of the loss $\loss$, \thatIs, $\var(\funcNoise)$ of regression and \dice losses to derive this lemma.

        \noIndentHeading{Case 1 $\normalSig \le 1$:} Given \cref{tab:optimality_bounds}, if $\normalSig \le 1$, the minimum deviation in converged regression model comes from the $\lTwo$ loss. 
        The difference in the estimates of regression loss and the \dice loss 
        \begin{align}
            \expect\left(\norm{\layerWeightConvReg-\layerWeightOptimal}_2^2\right) &- \expect\left(\norm{\layerWeightConvDice-\layerWeightOptimal}_2^2\right)\nonumber\\
            %&= \expect((\layerWeightConvLTwo\!-\!\layerWeightOptimal)^2) - \expect((\layerWeightConvDice\!-\!\layerWeightOptimal)^2) \nonumber \\
            &\propto \normalVar - \frac{1}{\length^2}\normalErf\left(\frac{\length}{\sqrt{2}\normalSig}\right) %\text{, using \cref{lemma:1,lemma:2}.}
        \end{align}

        Let $\normalSigTh$ be the solution of the equation $\normalVar = \dfrac{1}{\length^2}\normalErf\left(\dfrac{\length}{\sqrt{2}\normalSig}\right)$.
        Note that the above equation has unique solution $\normalSigTh$ since $\normalVar$ is a strictly increasing function wrt $\normalSig$ for $\normalSig > 0$, while $\dfrac{1}{\length^2}\normalErf\left(\dfrac{\length}{\sqrt{2}\normalSig}\right)$ is a strictly decreasing function wrt $\normalSig$ for $\normalSig > 0$.
        If the noise has $\normalSig \ge \normalSigTh$, the RHS of the above equation $\ge 0$, which means \dice loss converges better than the regression loss. 

        \noIndentHeading{Case 2 $\normalSig \ge 1$:} Given \cref{tab:optimality_bounds}, if $\normalSig \ge 1$, the minimum deviation in converged regression model comes from the $\lOne$ loss. 
        The difference in the regression and \dice loss estimates: 
        \begin{align}
            \expect\left(\norm{\layerWeightConvReg-\layerWeightOptimal}_2^2\right) &- \expect\left(\norm{\layerWeightConvDice-\layerWeightOptimal}_2^2\right)\nonumber\\
            %&= \expect((\layerWeightConvLOne\!-\!\layerWeightOptimal)^2) - \expect((\layerWeightConvDice\!-\!\layerWeightOptimal)^2) \nonumber \\
            &\propto 1- \frac{1}{\length^2}\normalErf\left(\frac{\length}{\sqrt{2}\normalSig}\right) %\text{, using \cref{lemma:1,lemma:2}.}
        \end{align}
        If the noise has $\normalSig \ge \dfrac{\sqrt{2}}{\length}\normalErfInv(\length^2)$, the RHS of the above equation $\ge 0$, which means \dice loss is better than the regression loss.
        For objects such as cars and trailers which have length $\length > 4m$, this is trivially satisfied.

        Combining both cases, \dice loss outperforms the $\lOne$ and $\lTwo$ losses if the noise deviation $\normalSig$ exceeds the critical threshold $\normalSigCr$, \thatIs
        \begin{align}
            \normalSig > \normalSigCr = \max \left( \normalSigTh, \dfrac{\sqrt{2}}{\length}\normalErfInv(\length^2)\right).
            \label{eq:final_bound} %\tag*{\qed}%\rlap{$\qquad \Box$}
        \end{align}
    \end{proof}

    %============================================================================
    %============================================================================
    \subsection{Proof of \cref{th:1}}\label{sec:supp_proof_theorem_1}
    \begin{proof}
        Continuing from \cref{lemma:3}, the advantage of the trained weight obtained from \dice loss over the trained weight obtained from regression losses further results in
        \begin{align}
            \var(\layerWeightConvDice) &\le \var(\layerWeightConvReg) \nonumber \\
            \implies \expect(|\layerWeightConvDice\image-\depthGT|) &\le \expect(|\layerWeightConvReg\image-\depthGT|) \nonumber \\
            \implies \expect(|\depthPredDice-\depthGT|) &\le \expect(|\depthPredReg-\depthGT|) \nonumber \\
            \implies \expect(\!{}^d\iouThreeDMath) &\ge \expect(\!{}^r\iouThreeDMath),
        \end{align}
        assuming depth is the only source of error. 
        Because \apThreeD is an non-decreasing function of \iouThreeD, the inequality remains preserved. 
        Hence, we have ${}^d$\apThreeD $\ge {}^r$\apThreeD. 
    \end{proof}
    Thus, the average precision from the \dice model is better than the regression model, which means a better detector.

    %============================================================================
    %============================================================================
    \subsection{Properties of \Dice Loss.}\label{sec:supp_dice_properties}  

        We next explore the properties of model in \cref{lemma:3} trained with \dice loss.
        From \cref{lemma:1}, we write
        \begin{align}
            \expect\left(\norm{\layerWeightConvDice-\layerWeightOptimal}^2_2\right) 
            &= \stepConstant \var(\funcNoise) + \uselessConstant \nonumber
        \end{align}
        Substituting the result of \cref{lemma:2}, we have
        \begin{align}
            \expect\left(\norm{\layerWeightConvDice-\layerWeightOptimal}^2_2\right) &= \frac{\stepConstant}{\length^2}\normalErf\left(\frac{\length}{\sqrt{2}\normalSig} \right)+\uselessConstant
            \label{eq:dice_convergence}
        \end{align}

        Paper \cite{birnbaum1942inequality} says that for a normal random variable $X$ with mean $0$ and variance $1$ and for any $x > 0$, we have
        \begin{align}
            \frac{\sqrt{4+x^2}-x}{2} \sqrt{\frac{1}{2\pi}} e^{-\frac{x^2}{2}} &\le P\left(X > x\right) \nonumber \\
            \implies \frac{1}{x+\sqrt{4+x^2}} \sqrt{\frac{2}{\pi}} e^{-\frac{x^2}{2}} &\le P\left(X > x\right) \nonumber \\
            \implies \frac{1}{x+\sqrt{4+x^2}} \sqrt{\frac{2}{\pi}} e^{-\frac{x^2}{2}}  &\le 1 - P\left(X \le x\right) \nonumber \\
            % \implies \frac{1}{x+\sqrt{4+x^2}} \sqrt{\frac{2}{\pi}} e^{-\frac{x^2}{2}}  &\le \frac{1}{2} - \int_0^x \frac{1}{\sqrt{2\pi\normalVar}} e^{-\frac{X^2}{2\normalVar}} dX \nonumber \\
            \implies \frac{1}{x+\sqrt{4+x^2}} \sqrt{\frac{2}{\pi}} e^{-\frac{x^2}{2}}  &\le 1\!-\!\frac{1}{2}\!-\!\int_0^{x} \frac{1}{\sqrt{2\pi}} e^{-\frac{X^2}{2}} dX \nonumber \\
            \implies \frac{1}{x+\sqrt{4+x^2}} \sqrt{\frac{2}{\pi}} e^{-\frac{x^2}{2}}  &\le \frac{1}{2}\!-\!\int_0^{x} \frac{1}{\sqrt{2\pi}} e^{-\frac{X^2}{2}} dX \nonumber \\
            \implies \frac{1}{x+\sqrt{4+x^2}} \sqrt{\frac{2}{\pi}} e^{-\frac{x^2}{2}}  &\le \frac{1}{2} - \int_0^{\frac{x}{\sqrt{2}}} \frac{1}{\sqrt{\pi}} e^{-X^2} dX \nonumber \\
            \implies \frac{1}{x+\sqrt{4+x^2}} \sqrt{\frac{2}{\pi}} e^{-\frac{x^2}{2}}  &\le \frac{1}{2} - \frac{1}{2}\normalErf\left(\frac{x}{\sqrt{2}}\right) \nonumber \\
            \implies \normalErf\left(\frac{x}{\sqrt{2}}\right) &\le 1 - \frac{2}{x+\sqrt{4+x^2}} \sqrt{\frac{2}{\pi}} e^{-\frac{x^2}{2}} \nonumber
        \end{align}
        Substituting $x=\dfrac{\length}{\normalSig}$ above, we have,
        \begin{align}
            \normalErf\left(\frac{\length}{\sqrt{2}\normalSig} \right) &\le 1 - \frac{2\normalSig}{\length+\sqrt{4\normalVar+\length^2}} \sqrt{\frac{2}{\pi}} e^{-\frac{\length^2}{2\normalVar}}
            \label{eq:big}
        \end{align}

        %============================================================================
        \noIndentHeading{Case 1: Upper bound.}  
            The RHS of \cref{eq:big} is clearly less than $1$ since the term in the RHS after subtraction is positive.
            Hence, 
            \begin{align}
                \normalErf\left(\frac{\length}{\sqrt{2}\normalSig} \right) &\le 1 \nonumber
            \end{align}
            Substituting above in \cref{eq:dice_convergence}, we have 
            \begin{align}
                \expect\left(\norm{\layerWeightConvDice-\layerWeightOptimal}^2_2\right) &\le  \frac{\stepConstant }{\length^2} + \uselessConstant
            \end{align}
            Clearly, the deviation of the trained model with the \dice loss is inversely proportional to the object length $\length$. 
            The deviation from the optimal is less for large objects.

        \begin{table}[!t]
            \caption{\textbf{Assumption comparison} of \cref{th:1} vs \monoThreeD models.
            }
            \label{tab:assumption_comp}
            \centering
            \scalebox{\scaleFraction}{
            \setlength\tabcolsep{0.15cm}
            \begin{tabular}{l m c m c }
                \addlinespace[0.01cm]
                & \cref{th:1} & \monoThreeD Models\\ 
                \myTopRule
                Regression & Linear & Non-linear \\
                Noise $\noise$ PDF & Normal & Arbitrary\\
                Noise \& Image & Independent & Dependent\\
                Object Categories & $1$ & Multiple \\
                Object Size $\length$ & Ideal & Non-ideal\\
                Error & Depth & All $7$ parameters\\
                Loss $\loss$ & $\lOne, \lTwo$, \dice & $\smoothLOne, \lTwo$, \dice, CE \\ 
                Optimizers & SGD & SGD, Adam, AdamW\\
                Global Optima & Unique & Multiple \\
            \end{tabular}
            }
        \end{table}

        %============================================================================
        \noIndentHeading{Case 2: Infinite Noise variance} $\normalVar \rightarrow \infty$. 
            Then, one of the terms in the RHS of \cref{eq:big} $\dfrac{2\normalSig}{\length+\sqrt{4\normalVar+\length^2}} \rightarrow 1$.
            Moreover, $\dfrac{\length}{\normalSig} \rightarrow 0 \implies e^{-\frac{\length^2}{2\normalVar}} \approx \left(1 - \dfrac{\length^2}{2\normalVar}\right)$. 
            So, RHS of \cref{eq:big} becomes
            \begin{align}
                \normalErf\left(\frac{\length}{\sqrt{2}\normalSig} \right) &\approx 1 - \sqrt{\frac{2}{\pi}} \left(1 - \frac{\length^2}{2\normalVar}\right) \nonumber \\
                \implies \normalErf\left(\frac{\length}{\sqrt{2}\normalSig} \right) &\approx \left( 1 + \sqrt{\frac{2}{\pi}} + \sqrt{\frac{2}{\pi}}\frac{\length^2}{2\normalVar}\right)
            \end{align}
            Substituting above in \cref{eq:dice_convergence}, we have 
            \begin{align}
            \expect\left(\norm{\layerWeightConvDice-\layerWeightOptimal}^2_2\right) &\approx \frac{\stepConstant }{\length^2} \left( 1 + \sqrt{\frac{2}{\pi}} + \sqrt{\frac{2}{\pi}}\frac{\length^2}{2\normalVar}\right) \nonumber \\ 
                &~~~~ + \uselessConstant 
            \end{align} 
            Thus, the deviation from the optimal weight is inversely proportional to the noise deviation $\normalVar$.
            Hence, the deviation from the optimal weight decreases as $\normalVar$ increases for the \dice loss.
            This property provides noise-robustness to the model trained with the \dice loss.

    %============================================================================
    %============================================================================
    \subsection{Notes on Theoretical Result}\label{sec:supp_theory_notes}

        %============================================================================
        \noIndentHeading{Assumption Comparisons.}
            The theoretical result of \cref{th:1} relies upon several assumptions. 
            We present a comparison between the assumptions made by \cref{th:1} and those underlying \monoThreeD models, in \cref{tab:assumption_comp}. 
            While our analysis depends on these assumptions, it is noteworthy that the results are apparent even in scenarios where the assumptions do not hold true.
            Another advantage of having a linear regression setup is that this setup has a unique global minima (because of its convexity).

\begin{figure}[!t]
            \centering
            \includegraphics[width=0.8\linewidth]{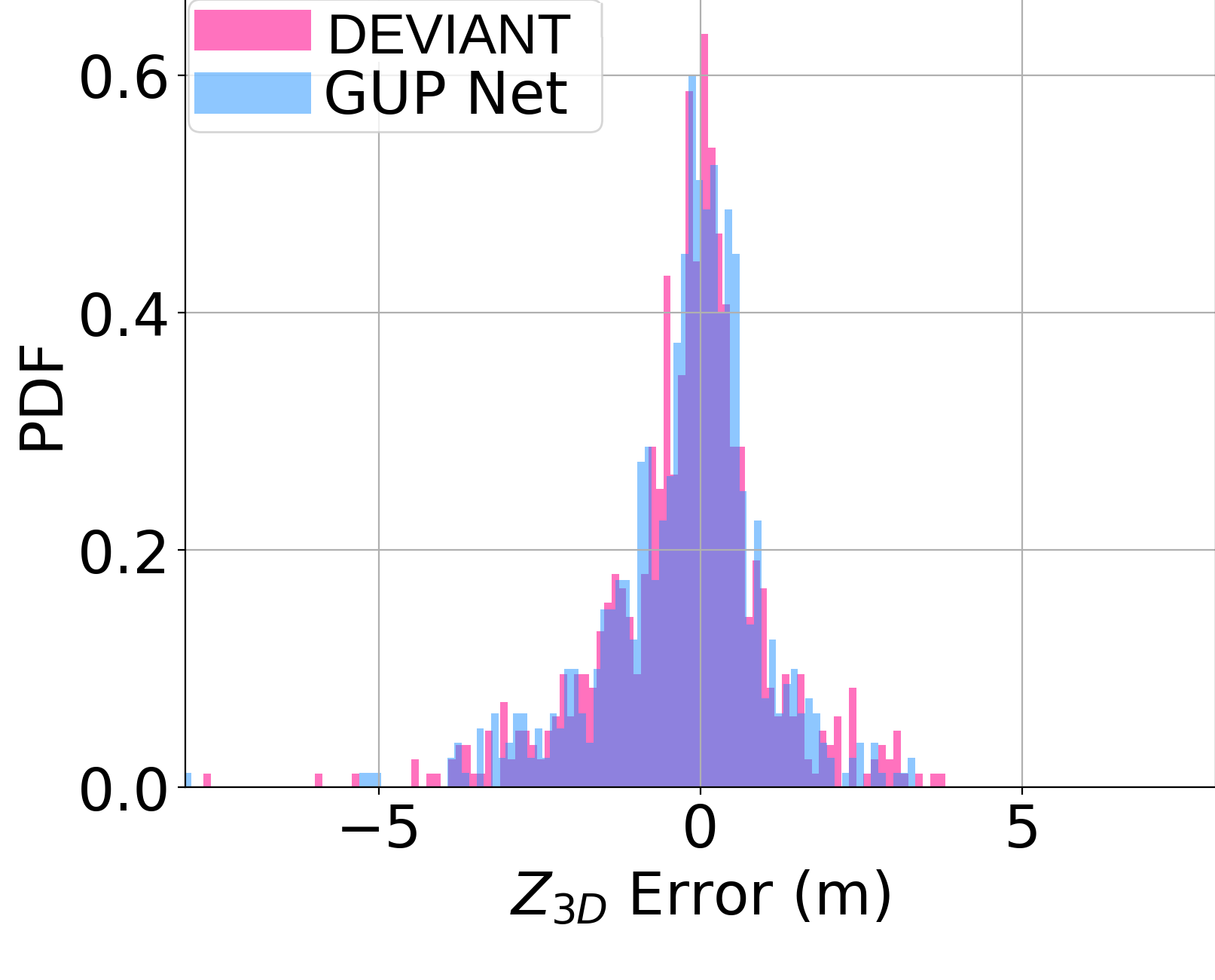}
            \caption{\textbf{Depth error histogram} of released \gupNet and \deviant \cite{kumar2022deviant} on the \kitti \val cars.
            The histogram shows that depth error is close to the Gaussian random variable.
            }
            \label{fig:depth_error_histogram}
        \end{figure}

        %============================================================================
        \noIndentHeading{Nature of Noise $\noise$.} 
            \cref{th:1} assumes that the noise $\noise$ is a normal random variable $\normal(0,\normalVar)$. 
            To verify this assumption, we take the two \sota released models \gupNet \cite{lu2021geometry} and \deviant \cite{kumar2022deviant} on the \kitti \cite{geiger2012we} \val cars. 
            We next plot the depth error histogram of both these models in \cref{fig:depth_error_histogram}.
            This figure confirms that the depth error is close to the Gaussian random variable.
            Thus, this assumption is quite realistic.

        %============================================================================
        \noIndentHeading{\cref{th:1} Requires Assumptions?}
            We agree that \cref{th:1} requires assumptions for the proof.
            However, our theory does have empirical support; most \monoThreeD works have no theory. 
            So, our theoretical attempt for \monoThreeD is a step forward! 
            We leave the analysis after relaxing some or all of these assumptions for future avenues.

        %============================================================================
        \noIndentHeading{Does \cref{th:1} Hold in Inference?}
            Yes, \cref{th:1} holds even in inference.
            \cref{th:1} relies on the converged weight $\layerWeightConv$, which in turn depends on the training data distribution. 
            Now, as long as the training and testing data distribution remains the same (a fundamental assumption in ML), \cref{th:1} holds also during inference.

    %============================================================================
    %============================================================================
    \subsection{More Discussions}

        %============================================================================
        \noIndentHeading{\methodName improves because it removes depth estimation and integrates \bev segmentation}.
            We clarify to remove this confusion. 
            First, \methodName also estimates depth. 
            \methodName depth estimates are better because of good segmentation, a \textit{form} of depth (thanks to \dice loss). 
            Second, predicted \bev segmentation needs processing with the \threeD head to output depth; so it can not replace depth estimation. 
            Third, integrating segmentation over all categories degrades \monoThreeD performance (\cite{li2022bevformer} and our \cref{tab:ablation} Sem. Category).

        %============================================================================
        \noIndentHeading{Why evaluation on outdoor datasets?}
            We experiment with outdoor datasets in this paper because indoor datasets rarely have large objects (mean length $>6m$).

%============================================================================
%============================================================================
%============================================================================
\section{Implementation Details}\label{sec:supp_implement_details}

        %============================================================================
        \noIndentHeading{Datasets.}
            Our experiments use the publicly available \kittiThreeSixty, \kittiThreeSixtyPanoptic and \nuscenes datasets.
            \kittiThreeSixty is available at \url{https://www.cvlibs.net/datasets/kitti-360/download.php} under CCA-NonCommercial-ShareAlike 
            (CC BY-NC-SA) 3.0 License.
            \kittiThreeSixtyPanoptic is available at \url{http://panoptic-bev.cs.uni-freiburg.de/} under Robot Learning License Agreement.
            \nuscenes is available at \url{https://www.nuscenes.org/nuscenes} under CC BY-NC-SA 4.0 International Public License.

        %============================================================================
        \noIndentHeading{Data Splits.}
            We detail out the detection data split construction of the \kittiThreeSixty dataset.
            \begin{itemize}
                \item \textit{\kittiThreeSixty Test split}: This detection benchmark \cite{liao2022kitti360} contains $300$ training and $42$ testing windows. 
                These windows contain $61{,}056$ training and $9{,}935$ testing images.
                The calibration exists for each frame in training, while it exists for every $10^\text{th}$ frame in testing.
                Therefore, our split consists of $61{,}056$ training images, while we run monocular detectors on $910$ test images (ignoring uncalibrated images).
    
                \item \textit{\kittiThreeSixty \val split}: The \kittiThreeSixty detection \val split partitions the official train into $239$ train and $61$ validation windows \cite{liao2022kitti360}. 
                The original \val split \cite{liao2022kitti360} contains $49{,}003$ training and $14{,}600$ validation images.
                However, this original \val split has the following three issues:
                \begin{itemize}
                    \item Data leakage (common images) exists in the training and validation windows.
                    \item Every \kittiThreeSixty image does not have the corresponding \bev semantic segmentation GT in the \kittiThreeSixtyPanoptic \cite{gosala2022bev} dataset, making it harder to compare \monoThreeD and \bev segmentation performance.
                    \item The \kittiThreeSixty validation set has higher sampling rate compared to the testing set.
                \end{itemize}
                To fix the data leakage issue, we remove the common images from training set and keep them only in the validation set. 
                Then, we take the intersection of \kittiThreeSixty and \kittiThreeSixtyPanoptic datasets to ensure that every image has corresponding \bev segmentation segmentation GT.
                After these two steps, the training and validation set contain $48{,}648$ and $12{,}408$ images with calibration and semantic maps. 
                Next, we subsample the validation images by a factor of $10$ as in the testing set. 
                Hence, our \kittiThreeSixty \val split contains $48{,}648$ training images and $1{,}294$ validation images.
                
            \end{itemize}

        %============================================================================
        \noIndentHeading{Augmentation.}
            We keep the same augmentation strategy as our baselines for the respective models.

        %============================================================================
        \noIndentHeading{Pre-processing.} We resize images to preserve their aspect ratio.
            \begin{itemize}
                \item \textit{\kittiThreeSixty.}
                    We resize the $[376, 1408]$ sized \kittiThreeSixty images, and bring them to the $[384, 1438]$ resolution.
            
                \item \textit{\nuscenes.}
                    We resize the $[900, 1600]$ sized \nuscenes images, and bring them to the $[256, 704]$, $[512, 1408]$ and $[640, 1600]$ resolutions as our baselines \cite{zhang2022beverse,zong2023hop}. 
            \end{itemize}

        %============================================================================
        \noIndentHeading{Libraries.}
            \imageToMaps and \panopticBEV experiments use PyTorch \cite{paszke2019pytorch}, while \beVerse and \hop use MMDetection3D \cite{mmdet3d2020}.

        %============================================================================
        \noIndentHeading{Architecture.}
            \begin{itemize}
                \item \textit{\imageToMapsWithMethod.}\imageToMaps \cite{saha2022translating} uses \resNetEighteen as the backbone with the standard Feature Pyramid Network (FPN) \cite{lin2017feature} and a transformer to predict depth distribution. 
                FPN is a bottom-up feed-forward CNN that computes feature maps with a downscaling factor of $2$, and a top-down network that brings them back to the high-resolution ones.
                There are total four feature maps levels in this FPN.
                We use the \orBoxNet with \resNetEighteen \cite{he2016deep} as the detection head.
                \item \textit{\panopticBEVWithMethod.}\panopticBEV \cite{gosala2022bev} uses \efficientDet \cite{tan2020efficientdet} as the backbone.
                We use \orBoxNet with \resNetEighteen \cite{he2016deep} as the detection head.
                \item \textit{\beVerseWithMethod.} \beVerse \cite{zhang2022beverse} uses Swin transformers \cite{liu2021swin} as the backbones.
                We use the original heads without any configuration change.
                \item \textit{\hopWithMethod.} \hop \cite{zong2023hop} uses \resNetFifty,  \resNetOneHundredOne \cite{he2016deep} and \vovNet \cite{park2021pseudo} as the backbones.
                Since \hop does not have the segmentation head, we use the one in \beVerse as the segmentation head.
            \end{itemize}
            We initialize the CNNs and transformers from \imageNet weights except for \vovNet, which is pre-trained on $15$ million \lidar data..
            We output two and ten foreground categories for \kittiThreeSixty and \nuscenes datasets respectively.

        %============================================================================
        \noIndentHeading{Training.}
            We use the training protocol as our baselines for all our experiments.
            We choose the model saved in the last epoch as our final model for all our experiments.
            \begin{itemize}
                \item \textit{\imageToMapsWithMethod.} 
                Training uses the Adam optimizer \cite{kingma2014adam}, a batch size of $30$, an exponential decay of $0.98$ \cite{saha2022translating} and gradient clipping of $10$ on single Nvidia A100 ($80$GB) GPU.
                We train the \bev Net in the first stage with a learning rate $1.0\!\times\!10^{-4}$ for $50$ epochs \cite{saha2022translating} .
                We then add the detector in the second stage and finetune with the first stage weight with a learning rate $0.5\!\times\!10^{-4}$ for $40$ epochs. 
                Training on \kittiThreeSixty \val takes a total of $100$ hours.
                For Test models, we finetune \imageToMaps \val stage 1 model with train+val data for $40$ epochs.
                \item \textit{\panopticBEVWithMethod.} 
                Training uses the Adam optimizer \cite{kingma2014adam} with Nesterov, a batch size of $2$ per GPU on eight Nvidia RTX A6000 ($48$GB) GPU.
                We train the \panopticBEV with the \dice loss in the first stage with a learning rate $2.5\!\times\!10^{-3}$ for $20$ epochs.
                We then add the \orBoxNet in the second stage and finetune with the first stage weight with a learning rate $2.5\!\times\!10^{-3}$ for $20$ epochs. 
                \panopticBEV decays the learning rate by $0.5$ and $0.2$ at $10$ and $15$ epoch respectively. 
                Training on \kittiThreeSixty \val takes a total of $80$ hours.
                For Test models, we finetune \panopticBEV \val stage 1 model with train+val data for $10$ epochs on four GPUs.
                \item \textit{\beVerseWithMethod.}
                Training uses the AdamW optimizer \cite{loshchilov2019decoupled}, a sample size of $4$ per GPU, the one-cycle policy \cite{zhang2022beverse} and gradient clipping of $35$ on eight Nvidia RTX A6000 ($48$GB) GPU \cite{zhang2022beverse}.
                We train the segmentation head in the first stage with a learning rate $2.0\!\times\!10^{-3}$ for $4$ epochs. 
                We then add the detector in the second stage and finetune with the first stage weight with a learning rate $2.0\!\times\!10^{-3}$ for $20$ epochs \cite{zhang2022beverse}. 
                Training on \nuscenes takes a total of $400$ hours.
                \item \textit{\hopWithMethod.}
                Training uses the AdamW optimizer \cite{loshchilov2019decoupled}, a sample size of $2$ per GPU, and gradient clipping of $35$ on eight Nvidia A100 ($80$GB) GPUs \cite{zong2023hop}.
                We train the segmentation head in the first stage with a learning rate $1.0\!\times\!10^{-4}$ for $4$ epochs. 
                We then add the detector in the second stage and finetune with the first stage weight with a learning rate $1.0\!\times\!10^{-4}$ for $24$ epochs \cite{zhang2022beverse}. 
                \nuscenes training takes a total of $180$ hours.
                For Test models, we finetune val model with train+val data for $4$ more epochs.
            \end{itemize}

        %============================================================================
        \noIndentHeading{Losses.}
            We train the \bev Net of \methodName in Stage 1 with the dice loss.
            We train the final \methodName pipeline in Stage 2 with the following loss:
            \begin{align}
                \loss &= \loss_{det} + \weightSeg \loss_{seg},
            \end{align}
            with $\loss_{seg}$ being the dice loss and $\weightSeg$ being the weight of the dice loss in the baseline. 
            We keep the $\weightSeg = 5$.
            If the segmentation loss is itself scaled such as \panopticBEV uses the $\loss_{seg}$ as $7$, we use $\weightSeg = 35$ with detection.

        %============================================================================
        \noIndentHeading{Inference.} 
            We report the performance of all \kittiThreeSixty and \nuscenes models by inferring on single GPU card.
            Our testing resolution is same as the training resolution. 
            We do not use any augmentation for test/validation. 
            
            We keep the maximum number of objects is $50$ per image for \kittiThreeSixty models. 
            We use score threshold of $0.1$ for \kittiThreeSixty models and class dependent threshold for \nuscenes models as in \cite{zhang2022beverse}.
            \kittiThreeSixty evaluates on windows and not on images. 
            So, we use a \threeD center-based NMS \cite{kumar2021groomed} to convert image-based predictions to window-based predictions for \methodName and all our \kittiThreeSixty baselines.
            This NMS uses a threshold of $4$m for all categories, and keeps the highest score \threeD box if multiple \threeD boxes exist inside a window.

%============================================================================
%============================================================================
%============================================================================
\section{Additional Experiments and Results}\label{sec:additional_exp}

        \begin{table}[!t]
            \caption{\textbf{Error analysis} on \kittiThreeSixty \val.
            }
            \label{tab:error_analysis}
            \centering
            \scalebox{\scaleFraction}{
            \setlength\tabcolsep{0.05cm}
            \begin{tabular}{ccccccc m bcc m bcc}
                \addlinespace[0.01cm]
                \multicolumn{7}{cm}{Oracle} & \multicolumn{3}{cm}{\apThreeDFifty~\bracketPercentage (\uparrowRHDSmall)} & \multicolumn{3}{c}{\apThreeDTwentyFive~\bracketPercentage (\uparrowRHDSmall)}\\ 
                $x$ & $y$ & $z$ & $l$ & $w$ & $h$ & $\theta$ & \MAPLarge & \MAPCar & \MAP & \MAPLarge & \MAPCar & \MAP\\
                \myTopRule
                & & & & & & & $8.71$	& $43.19$ & $25.95$	& $35.76$	& $52.22$ & $43.99$\\
                \cmark & & & & & & & $9.78$	& $41.63$ & $25.70$	& $36.07$	& $50.63$ & $43.35$\\
                & \cmark  && & & & & $9.57$	& $46.08$ & $27.82$	& $34.65$	& $53.03$ & $43.84$\\
                &&  \cmark & & & & & $9.90$	& $42.32$ & $27.11$	& $39.66$	& $53.08$ & $46.37$\\
                \cmark & \cmark & \cmark & & & & & $19.90$	& $47.37$ & $33.63$	& $41.84$	& $52.53$ & $47.19$\\
                & & & \cmark & \cmark & \cmark & & $9.49$	& $45.67$ & $27.58$	& $33.43$	& $51.53$ & $42.48$\\
                \cmark & \cmark & \cmark & \cmark & \cmark & \cmark & & $37.09$	& $46.27$ & $41.68$ & $44.58$	& $51.15$	& $47.87$\\
                %&  & &  &  & & \cmark & $8.19$	& $44.27$ & $26.23$	& $35.27$	& $52.34$ & $43.81$\\
                \cmark & \cmark & \cmark & \cmark & \cmark & \cmark & \cmark & $37.02$	& $47.03$ & $42.02$ & $44.46$	& $51.50$	& $47.98$\\
                % Oracle (GT \bev) & & & & & & & & $26.77$ & $51.79$ & $39.28$ & $49.74$ & $56.62$ & $53.18$\\
                \hline
            \end{tabular}
            }
        \end{table}

    We now provide additional details and results of the experiments evaluating \methodName’s performance.           

    %============================================================================
    %============================================================================
    \subsection{\kittiThreeSixty \val Results}

        %============================================================================
        \noIndentHeading{Error Analysis.}
            We next report the error analysis of the \methodName in \cref{tab:error_analysis} by replacing the predicted box data with the oracle box data as in \cite{ma2021delving}.
            We consider the GT box to be an oracle box for predicted box if the euclidean distance is less than $4m$. 
            In case of multiple GT being matched to one box, we consider the oracle with the minimum distance.
            \cref{tab:error_analysis} shows that depth is the biggest source of error for \monoThreeD task as also observed in \cite{ma2021delving}. 
            Moreover, the oracle does not lead to perfect results since the \kittiThreeSixtyPanoptic GT \bev semantic is only upto $50m$, while the \kittiThreeSixty evaluates all objects (including objects beyond $50m$).

        \begin{table}[!t]
            \caption{\textbf{Complexity analysis} on \kittiThreeSixty \val.
            }
            \label{tab:complexity_analysis}
            \centering
            \scalebox{\scaleFraction}{
            \setlength\tabcolsep{0.08cm}
            \begin{tabular}{cm c m c c c}
                Method & \monoThreeD & Inf. Time (s) & Param (M) & Flops (G)\\
                \myTopRule
                \gupNet \cite{lu2021geometry} & \cmark & $0.02$ & $16$ & $30$\\
                \deviant \cite{kumar2022deviant} & \cmark & $0.04$ & $16$ & $235$\\
                \imageToMaps \cite{saha2022translating} & \xmark & $0.01$ & $40$ & $80$\\
                \imageToMapsWithMethod & \cmark & $0.02$ & $53$ & $130$\\
                \panopticBEV \cite{gosala2022bev} & \xmark & $0.14$ & $24$ & $229$\\
                \panopticBEVWithMethod & \cmark & $0.15$ & $37$ & $279$\\ 
            \end{tabular}
            }
        \end{table} 

        \begin{table*}[!t]
            \caption{\textbf{\kittiThreeSixty \val results with naive baseline finetuned for large objects.}
            \methodName pipelines comfortably outperform this naive baseline on large objects.
            [Key: \firstKey{Best}, \secondKey{Second Best}, \retrained= Retrained]
            }
            \label{tab:det_seg_results_kitti_360_val_naive}
            \centering
            \scalebox{\scaleFraction}{
            \setlength\tabcolsep{0.15cm}
            \begin{tabular}{l | c m bcc | bcc m ccc}
                \addlinespace[0.01cm]
                \multirow{2}{*}{Method} & \multirow{2}{*}{Venue} &\multicolumn{3}{c|}{\apThreeDFifty~\bracketPercentage (\uparrowRHDSmall)} & \multicolumn{3}{cm}{\apThreeDTwentyFive~\bracketPercentage (\uparrowRHDSmall)} & \multicolumn{3}{c}{\bev Seg \iou~\bracketPercentage (\uparrowRHDSmall)}\\ 
                & & \MAPLarge & \MAPCar & \MAP & \MAPLarge & \MAPCar & \MAP & Large & Car & \meanFor \\
                \myTopRule
                \gupNet\retrained \cite{lu2021geometry} & ICCV21 &
                ${0.54}$ & \first{45.11} & ${22.83}$ & $0.98$ & ${50.52}$ & ${25.75}$ & \mathDash	& \mathDash	& \mathDash	\\
                \gupNet (Large FT)~\retrained \cite{lu2021geometry} & ICCV21 &
                ${0.56}$ & \mathDash & ${0.28}$ & $2.56$ & \mathDash & ${1.28}$ & \mathDash	& \mathDash	& \mathDash	\\
                \rowcolor{my_gray}\textbf{\imageToMapsWithMethod} & CVPR24 &
                \second{8.71}	& \second{43.19} & \second{25.95}	& \second{35.76}	& \second{52.22} & \second{43.99} & \second{23.23}	& \second{39.61}	& \second{31.42}	\\
                \rowcolor{my_gray}\textbf{\panopticBEVWithMethod} & CVPR24 &
                \first{13.22}	& $42.46$ & \first{27.84}	& \first{37.15}	& \first{52.53} & \first{44.84} & \first{24.30}	& \first{48.04}	& \first{36.17}	\\
            \end{tabular}
            }
        \end{table*}

        \begin{table*}[!t]
            \caption{\textbf{Impact of denoising} \bev segmentation maps with \mirNet \cite{zamir2022learning} on \kittiThreeSixty \val with \imageToMapsWithMethod. Denoising does not help.
            [Key: \bestKey{Best}]
            }
            \label{tab:ablation_more}
            \centering
            \scalebox{\scaleFraction}{
            \setlength\tabcolsep{0.15cm}
            \begin{tabular}{c m bcc | bcc m cccc}
                \addlinespace[0.01cm]
                \multirow{2}{*}{Denoiser} & \multicolumn{3}{c|}{\apThreeDFifty~\bracketPercentage (\uparrowRHDSmall)} & \multicolumn{3}{cm}{\apThreeDTwentyFive~\bracketPercentage (\uparrowRHDSmall)} & \multicolumn{3}{c}{\bev Seg \iou~\bracketPercentage (\uparrowRHDSmall)}\\ 
                & \MAPLarge & \MAPCar & \MAP & \MAPLarge & \MAPCar & \MAP & Large & Car & \meanFor\\
                \myTopRule
                \cmark & $2.73$ & \best{43.77} & $23.25$ & $14.34	$ & $51.23$ & $32.79$ & $21.42$ & \best{39.72} & $30.57$ \\
                \xmark & 
                \best{8.71}	& $43.19$ & \best{25.95}	& \best{35.76}	& \best{52.22} & \best{43.99} & \best{23.23}	& $39.61$ & \best{31.42}	\\
            \end{tabular}
            }
        \end{table*}

        \begin{table*}[!t]
            \caption{\textbf{Segmentation loss weight $\weightSeg$ sensitivity} on \kittiThreeSixty \val with \imageToMapsWithMethod.
            $\weightSeg=5$ works the best.
            [Key: \bestKey{Best}]
            }
            \label{tab:det_seg_results_sensitivity}
            \centering
            \scalebox{\scaleFraction}{
            \setlength\tabcolsep{0.15cm}
            \begin{tabular}{c m bcc m bcc m ccc}
                \addlinespace[0.01cm]
                \multirow{2}{*}{$\weightSeg$} & \multicolumn{3}{cm}{\apThreeDFifty~\bracketPercentage (\uparrowRHDSmall)} & \multicolumn{3}{cm}{\apThreeDTwentyFive~\bracketPercentage (\uparrowRHDSmall)} & \multicolumn{3}{c}{\bev Seg \iou~\bracketPercentage (\uparrowRHDSmall)}\\ 
                & \MAPLarge & \MAPCar & \MAP & \MAPLarge & \MAPCar & \MAP & Large & Car & \meanFor\\
                \myTopRule
                $0$ &$4.86$ & \best{45.09} & $24.98$ & $26.33$ & $52.31$ & $39.32$ & $0$ & $7.07$ & $3.54$ \\
                $1$ & $7.07$ & $41.71$ & $24.39$ & $32.92$ & $52.9$ & $42.91$ & $23.78$ & $40.58$ & $32.18$\\
                $3$ & $7.26$ & $43.45$ & $25.36$ & $34.47$ & \best{52.54} & $43.51$ & \best{23.40} & \best{40.15} & \best{31.78}\\
                $5$ & \best{8.71}	& $43.19$ & \best{25.95}	& \best{35.76} & $52.22$ & \best{43.99} & $23.23$	& $39.61$	& $31.42$ \\
                $10$ & $7.69$ & $43.41$ & $25.55$ & $34.22$ & $50.97$ & $42.60$ & $22.15$ & $39.83$ & $30.99$\\
            \end{tabular}
            }
        \end{table*}
        
        \begin{table*}[!t]
            \caption{\textbf{Reproducibility results} on \kittiThreeSixty \val with \imageToMapsWithMethod.
            \methodName outperforms \methodName without dice loss in the median and average cases.
            [Key: \firstKey{Best}, \secondKey{Second Best}]
            }
            \label{tab:det_seg_results_kitti_val_repeat}
            \centering
            \scalebox{\scaleFraction}{
            \setlength\tabcolsep{0.15cm}
            \begin{tabular}{c m c m bcc m bcc m cccc}
                \addlinespace[0.01cm]
                \multirow{2}{*}{Dice} &\multirow{2}{*}{Seed} & \multicolumn{3}{cm}{\apThreeDFifty~\bracketPercentage (\uparrowRHDSmall)} & \multicolumn{3}{cm}{\apThreeDTwentyFive~\bracketPercentage (\uparrowRHDSmall)} & \multicolumn{3}{c}{\bev Seg \iou~\bracketPercentage (\uparrowRHDSmall)}\\ 
                & & \MAPLarge & \MAPCar & \MAP & \MAPLarge & \MAPCar & \MAP & Large & Car & \meanFor \\
                \myTopRule
                \multirow{4}{*}{\xmark}
                & $111$ &$3.81$ & $44.63$ & $24.22$ & $24.96$ & $53.15$ & $39.06$ & $0$ & $5.99$ & $3.00$ \\
                & $444$ &$4.86$ & $45.09$ & $24.98$ & $26.33$ & $52.31$ & $39.32$ & $0$ & $7.07$ & $3.54$ \\
                & $222$ &$5.79$ & $46.71$ & $26.25$ & $24.32$ & $54.06$ & $39.19$ & $0$ & $5.32$ & $2.66$ \\
                \rowcolor{my_gray}& Avg & $4.82$ & $45.58$ & \second{25.15} & $25.20$ & $53.17$ & \second{39.19} & $0$ & $6.13$ & $3.06$\\
                \hline
                \multirow{4}{*}{\cmark}
                &  $111$ & $7.87$ & $44.03$ & $25.95$ & $33.55$ & $53.93$ & $43.74$ & $22.64$ & $40.64$ & $31.64$\\
                &  $444$ & $8.71$	& $43.19$ & $25.95$	& $35.76$ & $52.22$ & $43.99$ & $23.23$	& $39.61$	& $31.42$\\
                &  $222$ & $8.71$ & $42.87$ & $25.79$ & $34.71$ & $51.72$ & $43.22$ & $22.74$ & $40.01$ & $31.38$ \\
                \rowcolor{my_gray}& Avg & $8.43$ & $43.36$ & \first{25.90} & $34.67$ & $52.62$ & \first{43.65} & $22.87$ & $40.09$ & $31.48$ \\
            \end{tabular}
            }
        \end{table*}

        \begin{table*}[!t]
            \caption{\textbf{\Dice vs regression on methods with depth estimation}. 
            \Dice model again outperforms regression loss models, particularly for large objects.
            [Key: 
            \firstKey{Best}, \secondKey{Second Best}]
            }
            \label{tab:dice_vs_regression_on_depth}
            \centering
            \small
            \scalebox{\scaleFraction}{
            \setlength\tabcolsep{0.08cm}
            \begin{tabular}{l  l | c c c m b c c c c}
            Resolution & Method & BBone & Venue & Loss &\MAPLarge (\uparrowRHDSmall)& \MAPCar (\uparrowRHDSmall) & \MAPSmall (\uparrowRHDSmall) & \MAP (\uparrowRHDSmall) & \NDS (\uparrowRHDSmall) \\
            \myTopRule
            \multirow{4}{*}{$256\!\times\!704$} & \multirow{4}{*}{\hop\!+\methodName} & \multirow{4}{*}{R50} & ICCV23 & \mathDash & \second{27.4}    & \second{57.2}    & \second{46.4} & \second{39.9} & \second{50.9}\\
            & & & \mathDash & $\lOne$ & $27.0$    & $57.1$    & $46.5$ & $39.7$ & $50.7$\\
            & & & \mathDash & $\lTwo$ & \CYMyFix & \multicolumn{4}{c}{Did Not Converge} \\
            & & & CVPR24 & Dice & \first{28.2}    & \first{58.6}    & \first{47.8}    & \first{41.1}    & \first{51.5} \\
            \end{tabular}
        }
        \end{table*}
    
        %============================================================================
        \noIndentHeading{Computational Complexity Analysis.}
            We next compare the complexity analysis of \methodName pipeline in \cref{tab:complexity_analysis}. 
            For the flops analysis, we use the fvcore library as in \cite{kumar2022deviant}.

        %============================================================================
        \noIndentHeading{Naive baseline for Large Objects.}
            We next compare \methodName against a naive baseline for large objects detection, such as by fine-tuning \gupNet only on larger objects.
            \cref{tab:det_seg_results_kitti_360_val_naive} shows that \methodName pipelines comfortably outperform this baseline as well.
    
        %============================================================================
        \noIndentHeading{Does denoising \bev images help?}
            Another potential addition to the \methodName framework is using a denoiser between segmentation and detection heads. 
            We use the \mirNet \cite{zamir2022learning} as our denoiser and train the \bev segmentation head, denoiser and detection head in an end-to-end manner.
            \cref{tab:ablation_more} shows that denoising does not increase performance but the inference time.
            Hence, we do not use any denoiser for \methodName.
    
        %============================================================================
        \noIndentHeading{Sensitivity to Segmentation Weight.}
            We next study the impact of segmentation weight on \imageToMapsWithMethod in \cref{tab:det_seg_results_sensitivity} as in \cref{sec:ablation}.
            \cref{tab:det_seg_results_sensitivity} shows that $\weightSeg=5$ works the best for the \monoThreeD of large objects.
    
        %============================================================================
        \noIndentHeading{Reproducibility.}
            We ensure reproducibility of our results by repeating our experiments for $3$ random seeds.
            We choose the final epoch as our checkpoint in all our experiments as \cite{kumar2022deviant}.
            \cref{tab:det_seg_results_kitti_val_repeat} shows the results with these seeds. 
            \methodName outperforms \methodName without \dice loss in the median and average cases. 
            The biggest improvement shows up on larger objects.

    %============================================================================
    %============================================================================
    \subsection{\nuscenes Results}
        
        %============================================================================
        \noIndentHeading{Extended \val Results.}
            Besides showing improvements upon existing detectors in \cref{tab:nuscenes_val} on the \nuscenes \val split, we compare with more recent \sota detectors with large backbones in \cref{tab:nuscenes_val_more}. 

        %============================================================================
        \noIndentHeading{Dice vs regression on depth estimation methods}.
            We report \hop+R50 config, which uses depth estimation and compare losses in \cref{tab:dice_vs_regression_on_depth}.
            \cref{tab:dice_vs_regression_on_depth} shows that \Dice model again outperforms regression loss models.

        %============================================================================
        \noIndentHeading{\methodName Compatible Approaches.}
             \methodName conditions the detection outputs on segmented \bev features and so, requires foreground \bev segmentation.
             So, all approaches which produce latent \bev map in \cref{tab:nuscenes_test,tab:nuscenes_val} are compatible with \methodName.
             However, approaches which do not produce \bev features such as \sparseBEV \cite{liu2023sparsebev} are incompatible with \methodName.

        \begin{table*}[!t]
            \caption{\textbf{\nuscenes \val Detection results}. 
            \methodName pipelines outperform the baselines, particularly for large objects.
            [Key: 
            \firstKey{Best}, \secondKey{Second Best}, 
            B= Base, S= Small, T= Tiny, \released= Released, \reimplemented= Reimplementation, \cbgs= CBGS]
            }
            \label{tab:nuscenes_val_more}
            \centering
            \scalebox{\scaleFraction}{
            \setlength\tabcolsep{0.15cm}
            \begin{tabular}{l  l | c c m b c c c c}
                % \multirow{2}{*}{Resolution} & \multirow{2}{*}{Method} & \multirow{2}{*}{BBone} & \multicolumn{10}{c|}{3D Detection \bracketPercentage} & \multicolumn{2}{c}{Seg \iou \bracketPercentage \uparrowRHDSmall}\\
                Resolution & Method & BBone & Venue &\MAPLarge (\uparrowRHDSmall)& \MAPCar (\uparrowRHDSmall) & \MAPSmall (\uparrowRHDSmall) & \MAP (\uparrowRHDSmall) & \NDS (\uparrowRHDSmall) \\%& \mate\!\downarrowRHDSmall & \mase\!\downarrowRHDSmall & \maoe\!\downarrowRHDSmall & \mave\!\downarrowRHDSmall & \maae\!\downarrowRHDSmall & Car & Vehicle\\
                \myTopRule
                \multirow{7}{*}{$256\!\times\!704$}
                %& FIERY\cite{hu2021fiery}                            & $1$       & \mathDash & \mathDash & \mathDash & \mathDash & $35.8$ & \mathDash\\
                & CAPE\released\cite{xiong2023cape} & R50 & CVPR23 & $18.5$    & $53.2$    & $38.1$    & $31.8$    & $44.2$\\
                & \petrVTwo\cite{liu2023petrv2} & R50 & ICCV23 & \mathDash    & \mathDash    & \mathDash    & $34.9$    & $45.6$    \\
                & SOLOFusion\released\cbgs\cite{park2022time} & R50 & ICLR23 & $26.5$	& \second{57.3} & \first{48.5} & \second{40.6} & $49.7$\\
                & \beVerseTiny\released\cite{zhang2022beverse}& Swin-T & \arxiv & $18.5$        & $53.4$ & $38.8$           & $32.1$        & $46.6$           \\%& $43.6$        & $35.2$ \\
                & \cellcolor{my_gray}\textbf{\beVerseTinyWithMethod}         & \cellcolor{my_gray}Swin-T & \cellcolor{my_gray}CVPR24 & \cellcolor{my_gray}$19.5$ & \cellcolor{my_gray}$54.2$ & \cellcolor{my_gray}$41.1$    & \cellcolor{my_gray}$33.8$ & \cellcolor{my_gray}$48.1$\\
                %& \beVerseTiny\cite{zhang2022beverse}                & $5$       & $46.4$    & $31.5$    & $38.6$ & $17.2$ & $43.6$ & $35.2$ \\
                &\hop\released\cite{zong2023hop}  & R50 & ICCV23 & \second{27.4}    & $57.2$    & $46.4$ & $39.9$ & \second{50.9} \\%&\mathDash & \mathDash\\ 
                & \cellcolor{my_gray}\textbf{\hopWithMethod}      & \cellcolor{my_gray}R50 & \cellcolor{my_gray}CVPR24 & \cellcolor{my_gray}\first{28.2}    & \cellcolor{my_gray}\first{58.6}    & \cellcolor{my_gray}\second{47.8} & \cellcolor{my_gray}\first{41.1} & \cellcolor{my_gray}\first{51.5} \\%& $xxx$ & $xxx$\\
                \myTopRule
                \multirow{8}{*}{$512\!\times\!1408$}
                %&\beVerseSmall\cite{zhang2022beverse}                & $5$       & $51.4$    & $37.4$    & $44.3$ & $23.5$ & $47.9$ & $41.0$\\ 
                & \threeDPPE\cite{shu2023dppe} & R101 & ICCV23 & \mathDash & \mathDash & \mathDash & $39.1$ & $45.8$ \\
                & STS \cite{wang2022sts} & R101 & AAAI23 & \mathDash & \mathDash & \mathDash & $43.1$ & $52.5$ \\%& $0.525$ & $0.262$ & $0.380$ & $0.204$ & $0.542$ & \mathDash & \mathDash\\
                & P2D \cite{kim2023predict} & R101 & ICCV23 & \mathDash & \mathDash & \mathDash & $43.3$ & $52.8$ \\%& $0.550$ & \mathDash & $0.517$ & \mathDash & \mathDash\\
                &\bevDepth\cite{li2023bevdepth}            & R101 & AAAI23 & \mathDash & \mathDash & \mathDash   & $41.8$  & $53.8$   \\%& \mathDash & \mathDash & \mathDash & \mathDash \\
                &\bevDetFourD\cite{huang2022bevdet4d}                & R101 & \arxiv & \mathDash & \mathDash & \mathDash & $42.1$   & $54.5$     \\%& \mathDash \\
                % &\bevStereo\cite{li2023bevstereo}    & CNext & \mathDash & $57.5$    & $47.8$    \\%& \mathDash & \mathDash & \mathDash & \mathDash\\
                &\beVerseSmall\released\cite{zhang2022beverse}& Swin-S & \arxiv & $20.9$           & $56.2$           & $42.2$        & $35.2$        & $49.5$ \\ 
                & \cellcolor{my_gray}\textbf{\beVerseSmallWithMethod}                          & \cellcolor{my_gray}Swin-S & \cellcolor{my_gray}CVPR24 & \cellcolor{my_gray}$24.6$    & \cellcolor{my_gray}$58.7$    & \cellcolor{my_gray}$45.0$  & \cellcolor{my_gray}$38.2$ & \cellcolor{my_gray}$51.3$ \\
                & \hop\reimplemented\cite{zong2023hop}             & R101 & ICCV23 & 
                \second{31.4}    & \second{63.7}    & \second{52.5}    & \second{45.2}    & \first{55.0} \\%& \mathDash & \mathDash\\ 
                & \cellcolor{my_gray}\textbf{\hopWithMethod}                          & \cellcolor{my_gray}R101 & \cellcolor{my_gray}CVPR24 & \cellcolor{my_gray}\first{32.9}    & \cellcolor{my_gray}\first{65.0}    & \cellcolor{my_gray}\first{53.1} & \cellcolor{my_gray}\first{46.2} & \cellcolor{my_gray}\second{54.7} \\%& xxx & xxx\\
                \myTopRule
                \multirow{6}{*}{$640\!\times\!1600$}
                & BEVDet \cite{huang2021bevdet} & \vovNet & \arxiv & $29.6$ & $61.7$ & $48.2$ & $42.1$ & $48.2$ \\%& $0.529$ & $0.236$ & $0.396$ & $0.979$ & $0.152$\\
                %& \bevFormer\cite{li2022bevformer} & \vovNet & & & & $43.5$ & $49.5$ & $0.589$ & $0.254$ & $0.402$ & $0.842$ & $0.131$\\
                %& SpatialDETR\cite{doll2022spatialdetr,shu2023dppe} & \vovNet & \mathDash & \mathDash & \mathDash & $42.4$ & $48.6$ \\%& $0.613$ & $0.253$ & $0.402$ & $0.857$ & $0.131$\\%& $0.569$ & $0.255$ & $0.394$ & $0.796$ & $0.138$\\
                & \petrVTwo\cite{liu2023petrv2} & R101 & ICCV23 & \mathDash    & \mathDash    & \mathDash    & $42.1$    & $52.4$    \\%& $0.561$    & $0.243$    & $0.361$    & $0.343$    & $0.12$\\
                & CAPE\released\cite{xiong2023cape} & \vovNet & CVPR23 & $31.2$    & $63.2$    & $51.9$    & $44.7$    & $54.4$\\
                & \bevDetFourD\cbgs\cite{huang2022bevdet4d} & Swin-B & \arxiv & \mathDash    & \mathDash    & \mathDash & $42.6$ & $55.2$\\ 
                % & \streamPETR\cite{wang2023stream} & \vovNet & \\
                & \hop\reimplemented\cite{zong2023hop}  & \vovNet & ICCV23 & \second{36.5}    & \second{69.1}    & \second{56.1}    & \second{49.6}    & \second{58.3} \\%& \mathDash & \mathDash\\ 
                & \cellcolor{my_gray}\textbf{\hopWithMethod}               & \cellcolor{my_gray}\vovNet & \cellcolor{my_gray}CVPR24 & \cellcolor{my_gray}\first{40.3}    & \cellcolor{my_gray}\first{71.7}    & \cellcolor{my_gray}\first{58.8} & \cellcolor{my_gray}\first{52.7} & \cellcolor{my_gray}\first{60.2} \\
                \myTopRule
                \multirow{6}{*}{$900\!\times\!1600$} 
                % & LSS\cite{philion2020lift}                         & R50 &&&&&&&&&& & $32.1$ & $32.1$\\
                % & FIERY\cite{hu2021fiery}                           & R50 &&&&&&&&&& & $38.2$ & \mathDash\\
                %&VPN \cite{pan2020cross} in \cite{li2022bevformer}   & $2$       & $33.4$    & $25.7$    & \mathDash & \mathDash & $36.6$ & $37.3$ \\
                % & \monoDISMulti\cite{simonelli2020disentangling} & R34 & $16.5$ & $47.8$ & $38.1$	& $30.4$ & $38.4$	\\%& $0.738$ & $0.263$ & $0.546$ & $1.553$ & $0.134$\\
                & \fcosThreeD\!\cite{wang2021fcos3d} & R101 & ICCVW21 & \mathDash & \mathDash & \mathDash & $34.4$ & $41.5$ \\%& $0.690$ & $0.249$ & $0.452$ & $1.434$ & $0.124$ & \mathDash & \mathDash\\
                %&LSS\cite{philion2020lift} in \cite{li2022bevformer} & $2$       & $41.0$    & $34.4$    & \mathDash & \mathDash & $43.0$ & $42.8$\\
                & \pgd\cite{wang2021probabilistic} & R101 & CoRL21 & \mathDash & \mathDash & \mathDash & $36.9$ & $42.8$ \\%& $0.626$ & $0.245$ & $0.451$ & $1.509$ & $0.127$ & \mathDash & \mathDash\\
                % & \pgd\cite{wang2021probabilistic} & R101 & \\
                & \detrThreeD\cite{wang2021detr3d}& R101 & CoRL21 & $22.4$ & $60.3$ & $41.1$ & $34.9$ & $43.4$ \\%& $0.641$ & $0.255$ & $0.394$ & $0.845$ & $0.133$& \mathDash & \mathDash\\
                & \petr \cite{liu2022petr} & R101 & ECCV22 & \mathDash & \mathDash & \mathDash & $37.0$ & $44.2$ \\%& $0.593$ & $0.249$ & $0.383$ & $0.808$ & $0.132$& \mathDash & \mathDash\\
                & \bevFormer\cite{li2022bevformer} & R101 & ECCV22 & $27.7$ & $48.5$ & $34.5$ & $41.5$ & $51.7$ \\%& $0.582$ & $0.256$ & $0.375$ & $0.378$ & $0.126$& \mathDash & \mathDash\\
                & \polarFormer\cite{jiang2023polarformer} & \vovNet & AAAI23 & \mathDash & \mathDash & \mathDash & $50.0$ & $56.2$ \\%& $0.556$ & $0.256$ & $0.364$ & $0.440$ & $0.127$& \mathDash & \mathDash\\
            \end{tabular}
            }
        \end{table*}

    %============================================================================
    %============================================================================
    \subsection{Qualitative Results}

\begin{figure*}[!t]
            \centering
            \begin{subfigure}{\figureScaleFraction\linewidth}
                \includegraphics[width=\linewidth]{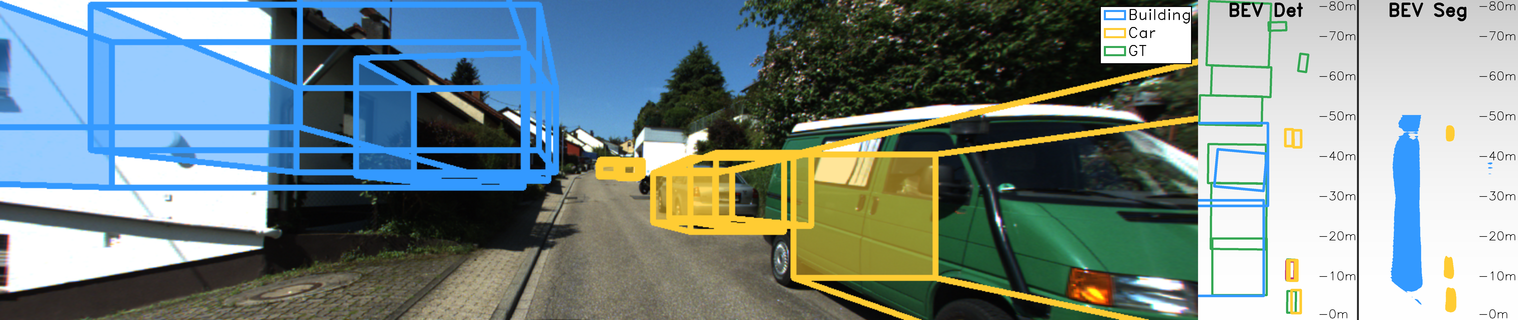}
            \end{subfigure}
            \begin{subfigure}{\figureScaleFraction\linewidth}
                \includegraphics[width=\linewidth]{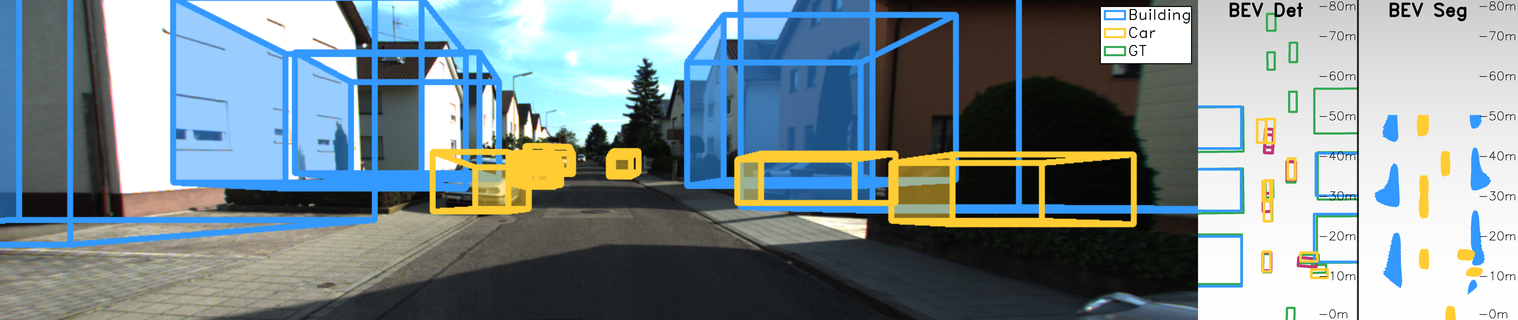}
            \end{subfigure}
            \begin{subfigure}{\figureScaleFraction\linewidth}
                \includegraphics[width=\linewidth]{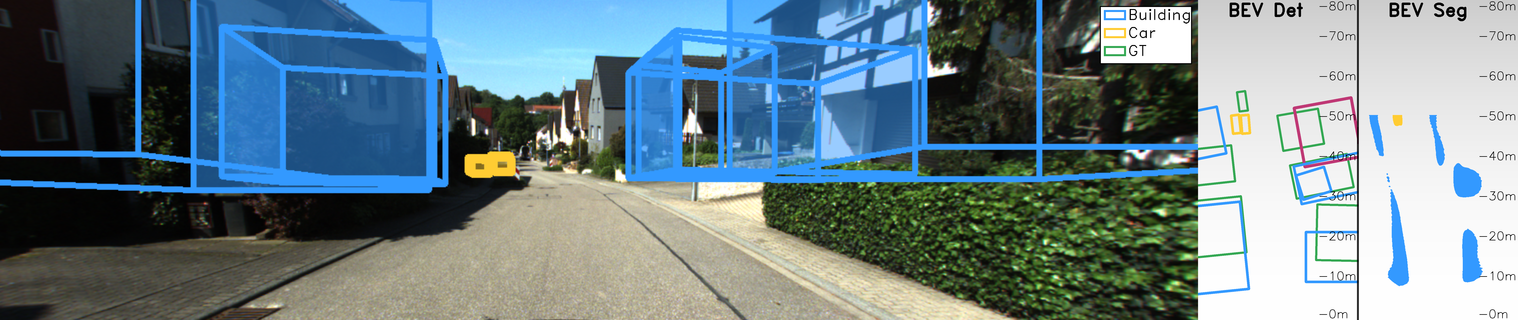}
            \end{subfigure}
            \begin{subfigure}{\figureScaleFraction\linewidth}
                \includegraphics[width=\linewidth]{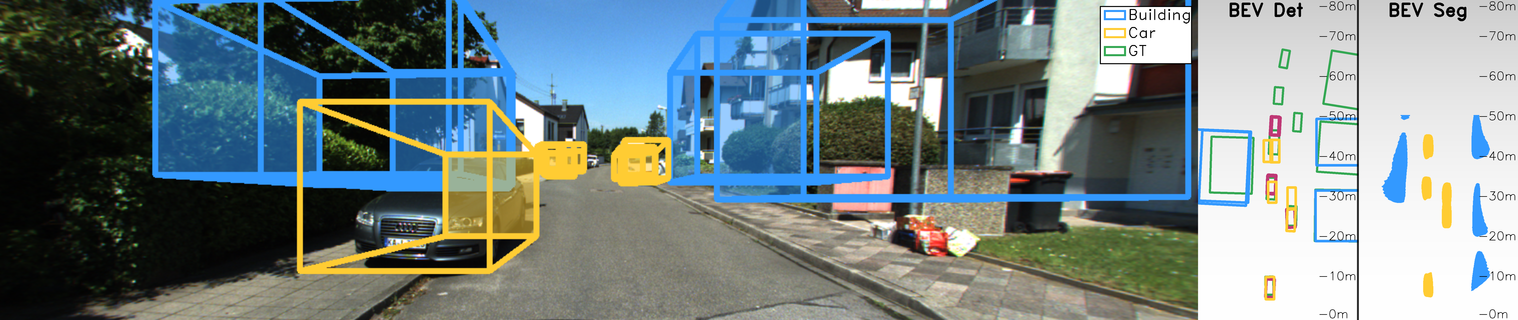}
            \end{subfigure}
            \begin{subfigure}{\figureScaleFraction\linewidth}
                \includegraphics[width=\linewidth]{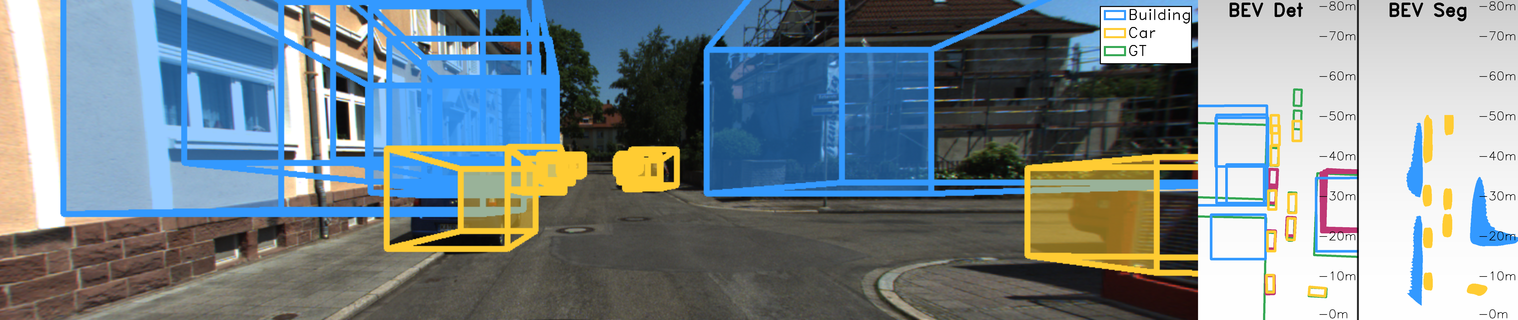}
            \end{subfigure}
            \begin{subfigure}{\figureScaleFraction\linewidth}
                \includegraphics[width=\linewidth]{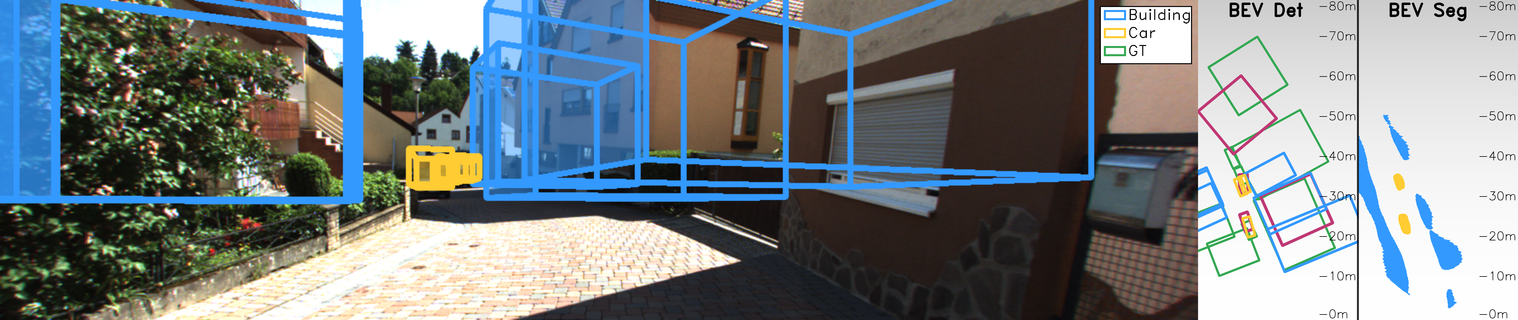}
            \end{subfigure}
            \begin{subfigure}{\figureScaleFraction\linewidth}
                \includegraphics[width=\linewidth]{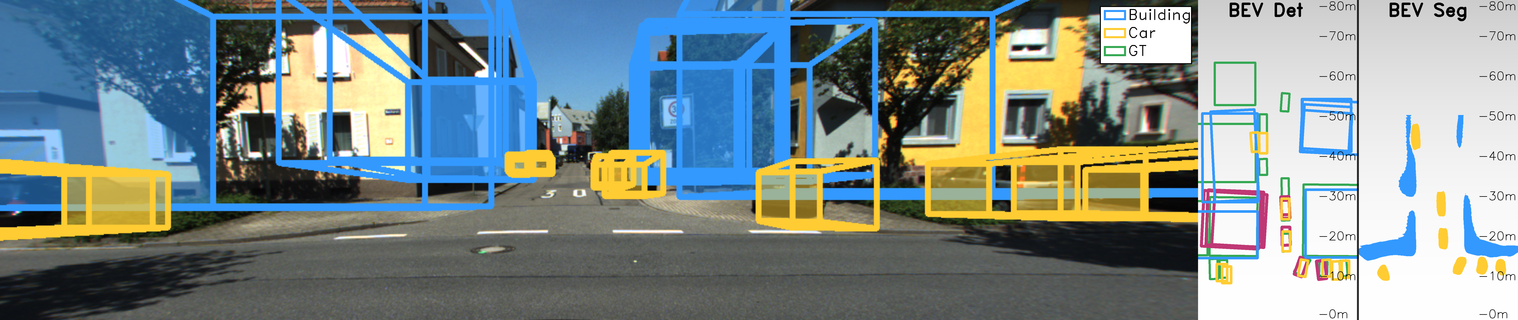}
            \end{subfigure}
            \caption{\textbf{\kittiThreeSixty Qualitative Results}. 
            \panopticBEVWithMethod detects more \textcolor{my_blue}{large objects} (buildings) than \textcolor{sns_orange}{\monodetr} \cite{zhang2023monodetr}.
            We depict the predictions of \panopticBEVWithMethod in the image view on the left, the predictions of \panopticBEVWithMethod, the baseline \monodetr \cite{zhang2023monodetr}, and ground truth in \bev in the middle, and \bev semantic segmentation predictions from \panopticBEVWithMethod on the right. 
            [Key: \textcolor{my_blue}{Buildings} and \textcolor{rayShade}{Cars} of \panopticBEVWithMethod; \textcolor{sns_orange}{all classes} of \monodetr \cite{zhang2023monodetr}, and \textcolor{darkGreen3}{Ground Truth} in BEV]. 
            }
            \label{fig:qualitative_kitti_360}
        \end{figure*}

            %============================================================================
            \noIndentHeading{\kittiThreeSixty.}
                We now show some qualitative results of models trained on \kittiThreeSixty \val split in \cref{fig:qualitative_kitti_360}. 
                We depict the predictions of \panopticBEVWithMethod in image view on the left, the predictions of \panopticBEVWithMethod, the baseline \monodetr \cite{zhang2023monodetr}, predicted and GT boxes in \bev in the middle and \bev semantic segmentation predictions from \panopticBEVWithMethod on the right. 
                In general, \panopticBEVWithMethod detects more larger objects (buildings) than \gupNet\cite{lu2021geometry}.
    
\begin{figure*}[!t]
            \centering
            \begin{subfigure}{\figureScaleFraction\linewidth}
                \includegraphics[width=\linewidth]{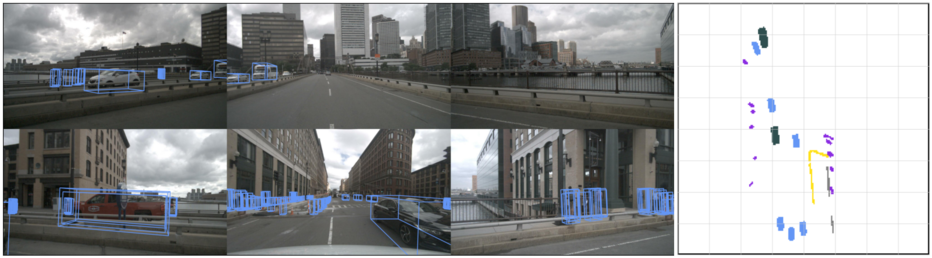}
            \end{subfigure}
            \begin{subfigure}{\figureScaleFraction\linewidth}
                \includegraphics[width=\linewidth]{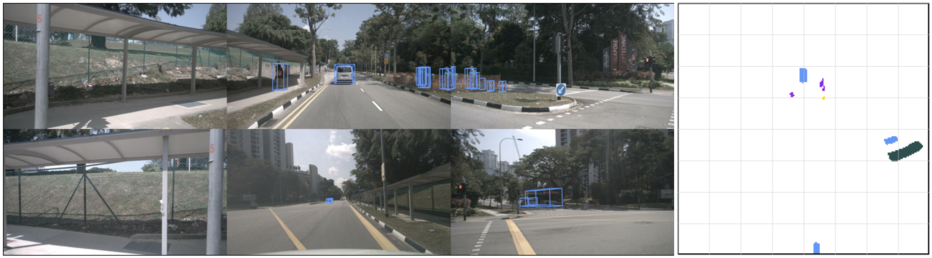}
            \end{subfigure}
            \begin{subfigure}{\figureScaleFraction\linewidth}
                \includegraphics[width=\linewidth]{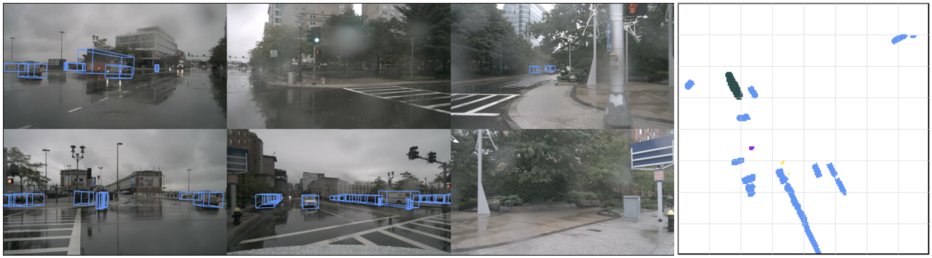}
            \end{subfigure}
            \begin{subfigure}{\figureScaleFraction\linewidth}
                \includegraphics[width=\linewidth]{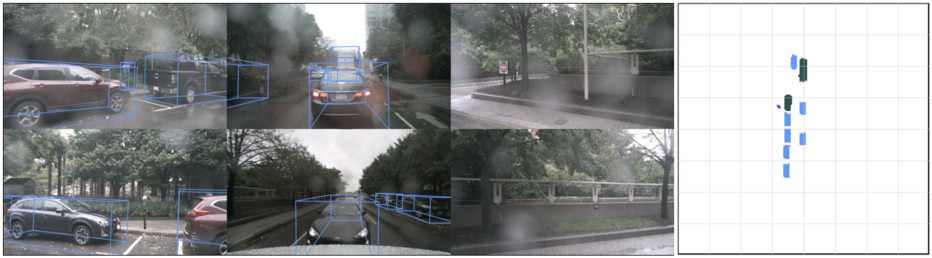}
            \end{subfigure}
            \begin{subfigure}{\figureScaleFraction\linewidth}
                \includegraphics[width=\linewidth]{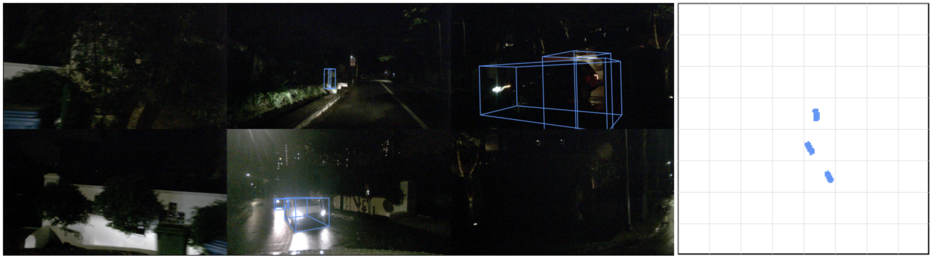}
            \end{subfigure}
            \caption{\textbf{\nuscenes Qualitative Results}. 
            The first row shows the front\_left, front, and front\_right cameras, while the second row shows the back\_left, back, and back\_right cameras. 
            [Key: \textcolor{my_blue}{Cars}, \textcolor{building}{Vehicles}, \textcolor{violet}{Pedestrian}, \textcolor{my_yellow}{Cones} and \textcolor{gray}{Barrier} of \beVerseSmallWithMethod at $200\!\times\!200$ resolution in \bev].}
            \label{fig:qualitative_nuscenes}
        \end{figure*}

            %============================================================================
            \noIndentHeading{\nuscenes.}
                We now show some qualitative results of models trained on \nuscenes \val split in \cref{fig:qualitative_nuscenes}.
                As before, we depict the predictions of \beVerseSmallWithMethod in image view from six cameras on the left and \bev semantic segmentation predictions from \methodName on the right.

            %============================================================================
            \noIndentHeading{\kittiThreeSixty Demo Video.} 
                We next put a short demo video of \methodName model
                trained on \kittiThreeSixty \val split compared with \monodetr at \url{https://www.youtube.com/watch?v=SmuRbMbsnZA}.
                We run our trained model independently on each frame of \kittiThreeSixty. 
                None of the frames from the raw video appear in the training set of \kittiThreeSixty \val split. 
                We use the camera matrices available with the video but do not use
                any temporal information. 
                Overlaid on each frame of the raw input videos, we
                plot the projected \threeD boxes of the predictions, predicted and GT boxes in \bev in the middle and \bev semantic segmentation predictions from \panopticBEVWithMethod. 
                We set the frame rate of this demo at $5$ fps similar to \cite{kumar2022deviant}. 
                The attached demo video demonstrates impressive results on larger objects.

%============================================================================
%============================================================================
%============================================================================
\section{Acknowledgements}
    This research was partially sponsored by the Bosch Research North America, Bosch Center for AI and the Army Research Office (ARO) grant W911NF-18-1-0330. 
    This document’s views and conclusions are those of the authors and do not represent the official policies, either expressed or implied, of the ARO or the U.S. government.
    
    We thank several members of the Computer Vision community for making this project possible.
    We deeply appreciate Rakesh Menon, Vidit, Abhishek Sinha, Avrajit Ghosh, Andrew Hou, Shengjie Zhu, Rahul Dey, Saurabh Kumar and Ayushi Raj for several invaluable discussions during this project.
    Rakesh suggested the \monodle \cite{ma2021delving} baseline for \kittiThreeSixty models because \monodle normalizes loss with GT box dimensions.
    Shengjie, Avrajit, Rakesh, Vidit, and Andrew proofread our manuscript and suggested several changes.
    Shengjie helped us parse the \kittiThreeSixty dataset, while 
    Andrew helped in the \kittiThreeSixty leaderboard evaluation.
    We also thank Prof.~Yiyi Liao from Zhejiang University for discussions on the \kittiThreeSixty conventions and evaluation protocol.
    We finally thank anonymous NeurIPS and CVPR reviewers for their exceptional feedback and constructive criticism that shaped this final manuscript.

%============================================================================
%============================================================================
%============================================================================
% \cleardoublepage
% {
%     \small
%     \bibliographystyle{ieeenat_fullname}
%     \bibliography{references}
% }

\end{document}

%% file: notations.tex
\definecolor{my_blue}{rgb}{0.2, 0.6, 1}  % dodgerblue
\definecolor{my_magenta}{rgb}{1.0, 0.2, 0.6} % triadic to dodgerblue
\definecolor{my_yellow}{rgb}{1.0, 0.8, 0.2} % triadic to magenta
\definecolor{my_green}{rgb}{0.0, 0.9, 0.24}
\definecolor{my_green_2}{rgb}{0.0, 0.4, 0.0}

\definecolor{sns_blue}{rgb}{0.21, 0.06, 0.42}
\definecolor{sns_violet}{rgb}{0.45, 0.12, 0.51}
\definecolor{sns_orange}{rgb}{0.75, 0.22, 0.46}
\definecolor{sns_yellow}{rgb}{0.99, 0.91, 0.66}

\definecolor{white}{rgb}{1.0, 1.0, 1.0}
\definecolor{darkGreen}{rgb}{0.01, 0.8, 0.24}%{0.31, 0.94, 0.3}%{0.09,0.88,0.00}{0.29, 0.83, 0.38}
\definecolor{darkGreen2}{rgb}{0.22,0.42, 0.33}
\definecolor{darkGreen3}{rgb}{0.20,0.66, 0.33}
\definecolor{cvprblue}{rgb}{0.21,0.49,0.74}
\definecolor{LightCyan}{rgb}{0.88,1,1}
\definecolor{lightgreen}{HTML}{90EE90}
\definecolor{new_green}{rgb}{0.75,0.97,0.44}
\definecolor{Gray}{gray}{0.95}
\definecolor{lightgray}{rgb}{0.96, 0.96, 0.96}

\definecolor{set1_cyan}{rgb}{0.23, 0.87, 1.0}
\definecolor{building}{rgb}{0.2, 0.33, 0.33}

\definecolor{my_violet}{rgb}{0.79, 0.40, 1} %{0.73,0.62,0.91}
\definecolor{my_yellow_2}{rgb}{0.9, 0.8, 0.54}
\definecolor{my_red}{rgb}{1,0,0}
\definecolor{my_purple}{rgb}{0.27,0.8, 0.8}
\definecolor{my_orange}{rgb}{1.0,0.6,0.35}
\definecolor{my_golden}{rgb}{1.0, 0.75, 0.0}
\colorlet{my_gray}{gray!12}

\definecolor{projectionColor}{rgb}{0.2, 0.6, 1}
\definecolor{rayColor}{rgb}{0.0,0.0,0.0}
\definecolor{axisColor}{rgb}{0.0, 0.0, 0.0}
\colorlet{projectionBorderShade}{rayColor!100}
\colorlet{projectionFillShade}{projectionColor!20}
\colorlet{rayShade}{my_yellow}
\colorlet{axisShade}{axisColor!20}
\colorlet{axisShadeDark}{axisColor!100}

\definecolor{backward_color}{rgb}{1.0, 0.6, 0.2}
\definecolor{forward_color}{rgb}{0.2, 1.0, 0.6}

\definecolor{gain}{HTML}{34a853}
\definecolor{lost}{HTML}{ea4335}

\colorlet{proposedShade}{darkGreen}
\colorlet{vanillaShade}{red!90}

\colorlet{theme_color}{sns_orange}%my_yellow
\colorlet{theme_color_light}{sns_yellow!25}

\newcommand{\figureScaleFraction}{0.8}

\newcommand{\noIndentHeading}[1]{\noindent\textbf{#1}}

\definecolor{XLcolor}{rgb}{0.858, 0.188, 0.478}
\newcommand{\thatIs}{\textit{i.e.}\xspace}

\newcommand{\cmark}{\checkmark}%\ding{51}}%
\newcommand{\xmark}{\ding{53}}

\DeclareMathOperator{\sign}{sgn}

\newcommand{\scaleFraction}{0.9}

\newcommand{\myTopRule}{\Xhline{2\arrayrulewidth}}

\newcolumntype{t}{!{\vrule width 1.5\arrayrulewidth}}
\newcolumntype{m}{!{\vrule width 2.5\arrayrulewidth}}
\newcolumntype{a}{>{\columncolor{theme_color_light}}l}
\newcolumntype{b}{>{\columncolor{theme_color_light}}c}

\newcommand{\CYMyFix}{\cellcolor{theme_color_light}}

\colorlet{cyan_highlight}{my_blue!85}

\colorlet{darkGreen_highlight}{darkGreen!75}

\colorlet{my_magenta_highlight}{my_magenta!50}

\colorlet{my_yellow_highlight}{my_yellow!55}

\providecommand\rightarrowRHD{\relbar\joinrel\mathrel\RHD}

\newcommand{\uparrowRHD}  {\rotatebox[origin=c]{90}{$\rightarrowRHD$}}
\newcommand{\downarrowRHD}{\rotatebox[origin=c]{270}{$\rightarrowRHD$}}
\newcommand{\uparrowRHDSmall}  {\raisebox{0.05\normalbaselineskip}{\scalebox{0.7}{\uparrowRHD}}}   %$\uparrow$
\newcommand{\downarrowRHDSmall}{\raisebox{0.07\normalbaselineskip}{\scalebox{0.7}{\downarrowRHD}}} %$\downarrow$

\newcommand{\monoThreeD}{Mono3D\xspace}

\newcommand{\twoD}{$2$D\xspace}
\newcommand{\threeD}{$3$D\xspace}

\newcommand{\iou}{IoU\xspace}

\newcommand{\iouThreeD}{IoU$_{3\text{D}}$\xspace}
\newcommand{\iouThreeDMath}{\text{\iou}_{3\text{D}}\xspace}
\newcommand{\lidar}{LiDAR\xspace}

\newcommand{\radar}{radar\xspace}

\newcommand{\bev}{BEV\xspace}

\newcommand{\resNetEighteen}{ResNet-18\xspace}
\newcommand{\resNetFifty}{ResNet-50\xspace}
\newcommand{\resNetOneHundredOne}{ResNet-101\xspace}

\newcommand{\vovNet}{V2-99\xspace}
\newcommand{\efficientDet}{EfficientDet\xspace}

\newcommand{\kitti}{KITTI\xspace}
\newcommand{\nuscenes}{nuScenes\xspace}
\newcommand{\waymo}{Waymo\xspace}

\newcommand{\imageNet}{ImageNet\xspace}

\newcommand{\val}{Val\xspace}

\newcommand{\kittiThreeSixty}{KITTI-360\xspace}
\newcommand{\kittiThreeSixtyPanoptic}{\kittiThreeSixty PanopticBEV\xspace}
\newcommand{\semanticKITTI}{Semantic KITTI\xspace}

\newcommand{\loss}{\mathcal{L}}

\newcommand{\lOne}{\loss_1}
\newcommand{\lTwo}{\loss_2}
\newcommand{\smoothLOne}{\text{Smooth}~\lOne}

\newcommand{\lDice}{\loss_{dice}}
\newcommand{\dice}{dice\xspace}
\newcommand{\Dice}{Dice\xspace}

\newcommand{\weightSeg}{\lambda_{seg}}

\newcommand{\MAP}{mAP\xspace}
\newcommand{\MAPLarge}{AP$_{Lrg}$\xspace}
\newcommand{\MAPCar}{AP$_{Car}$\xspace}
\newcommand{\MAPSmall}{AP$_{Sml}$\xspace}
\newcommand{\ap}{AP\xspace}
\newcommand{\apMath}{\text{\ap}}
\newcommand{\apThreeD}{$\apMath_{3\text{D}}$\xspace}

\newcommand{\NDS}{NDS\xspace}

\newcommand{\bracketPercentage}{[\%]}

\newcommand{\apThreeDFifty}{\ap$_{\!3\text{D}\!}$ 50\xspace}
\newcommand{\apThreeDTwentyFive}{\ap$_{\!3\text{D}\!}$ 25\xspace}

\newcommand{\meanFor}{M$_\text{For}$}
\newcommand{\meanEleven}{M$_\text{All}$}

\newcommand{\first}[1]{$\textcolor{sns_blue}{\mathbf{#1}}$}
\newcommand{\second}[1]{$\textcolor{sns_orange}{\mathbf{#1}}$}
\newcommand{\firstKey}[1]{\textcolor{sns_blue}{\textbf{#1}}}
\newcommand{\secondKey}[1]{\textcolor{sns_orange}{\textbf{#1}}}
\newcommand{\sota}{SoTA\xspace}
\newcommand{\best}[1]{$\textcolor{sns_blue}{\mathbf{#1}}$}
\newcommand{\bestKey}[1]{\textcolor{sns_blue}{\textbf{#1}}}
\newcommand{\gain}[1]{\textcolor{gain}{#1}}
\newcommand{\lost}[1]{\textcolor{lost}{#1}}
\newcommand{\good}[2]{{$#1$} {({\gain{\textbf{+}$\mathbf{#2}$}})}}
\newcommand{\bad}[2]{{$#1$} {({\lost{--$#2$}})}}
\newcommand{\arxiv}{ArXiv\xspace}

\newcommand{\mathDash}{$-$}

\def\relicon{\resizebox{.009\textwidth}{!}{\includegraphics{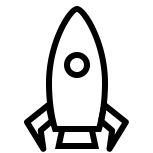}}}
\newcommand{\released}{$\!^{\relicon}$}
\newcommand{\retrained}{$\!^\dagger$}
\newcommand{\reimplemented}{$\!^*$}
\newcommand{\cbgs}{$\!^{\mathsection}$}
\newcommand{\future}{$\!^{\tiny{\fullmoon}\hspace{-0.03cm}\tiny{\newmoon}}$}%{$\!{\circ}$}

\newcommand{\groomedNMS}{GrooMeD-NMS\xspace}

\newcommand{\gupNet}{GUP Net\xspace}
\newcommand{\deviant}{DEVIANT\xspace}
\newcommand{\pseudoLidar}{Pseudo-{\lidar}\xspace}

\newcommand{\monodle}{MonoDLE\xspace}
\newcommand{\monodetr}{MonoDETR\xspace}

\newcommand{\detrThreeD}{DETR3D\xspace}
\newcommand{\fcosThreeD}{FCOS3D\xspace}
\newcommand{\pgd}{PGD\xspace}
\newcommand{\petr}{PETR\xspace}
\newcommand{\petrVTwo}{PETRv2\xspace}
\newcommand{\bevFormer}{BEVFormer\xspace}
\newcommand{\polarFormer}{PolarFormer\xspace}
\newcommand{\bevDepth}{BEVDepth\xspace}
\newcommand{\bevStereo}{BEVStereo\xspace}
\newcommand{\bevDetFourD}{BEVDet4D\xspace}
\newcommand{\beVerse}{BEVerse\xspace}
\newcommand{\beVerseTiny}{BEVerse-T\xspace}
\newcommand{\beVerseSmall}{BEVerse-S\xspace}
\newcommand{\orBoxNet}{Box Net\xspace}
\newcommand{\uvtr}{UVTR\xspace}
\newcommand{\veDet}{VEDet\xspace}
\newcommand{\cape}{CAPE\xspace}
\newcommand{\xThreeKDAll}{X3KD$_{\text{all}}$\xspace}
\newcommand{\frustumFormer}{FrustumFormer\xspace}
\newcommand{\threeDPPE}{3DPPE\xspace}
\newcommand{\hop}{HoP\xspace}
\newcommand{\parametricBEV}{ParametricBEV\xspace}
\newcommand{\sparseBEV}{SparseBEV\xspace}
\newcommand{\saBEV}{SA-BEV\xspace}
\newcommand{\fbBEV}{FB-BEV\xspace}

\newcommand{\beVerseWithMethod}{BEVerse+SeaBird\xspace}
\newcommand{\beVerseTinyWithMethod}{BEVerse-T+SeaBird\xspace}
\newcommand{\beVerseSmallWithMethod}{BEVerse-S+SeaBird\xspace}

\newcommand{\hopWithMethod}{HoP+SeaBird\xspace}
\newcommand{\stxd}{STXD\xspace}
\newcommand{\cubeRCNN}{Cube R-CNN\xspace}

\newcommand{\imageToMaps}{I2M\xspace}
\newcommand{\imageToMapsLong}{Image2Maps\xspace}
\newcommand{\imageToMapsWithMethod}{I2M+SeaBird\xspace}
\newcommand{\panopticBEV}{PBEV\xspace}
\newcommand{\panopticBEVLong}{PanopticBEV\xspace}
\newcommand{\panopticBEVWithMethod}{PBEV+SeaBird\xspace}

\newcommand{\lidarBoxNet}{L-BoxNet\xspace}
\newcommand{\voteNet}{L-VoteNet\xspace}

\newcommand{\mirNet}{MIRNet-v2\xspace}

\newcommand{\normal}{\mathcal{N}}
\newcommand{\depthPred}{\hat{\posZ}}
\newcommand{\depthPredReg}{{~}^r\depthPred}
\newcommand{\depthPredDice}{{~}^d\depthPred}

\newcommand{\depthGT}{\posZ}
\newcommand{\normalSig}{\sigma}
\newcommand{\normalSigTh}{\sigma_{m}}
\newcommand{\normalSigCr}{\sigma_{c}}
\newcommand{\normalVar}{\sigma^2}
\newcommand{\normalCDF}{\Phi}
\newcommand{\normalErf}{\text{Erf}}
\newcommand{\normalErfInv}{\text{Erf}^{-1}}
\newcommand{\expect}{\mathbb{E}}
\newcommand{\var}{\text{Var}}
\newcommand{\step}{s}

\newcommand{\stepConstant}{{c_1}}
\newcommand{\stepSumTrue}{s}%\mathcal{S}}
\newcommand{\image}{\mathbf{h}}
\newcommand{\noise}{\eta}
\newcommand{\funcNoise}{\epsilon}
\newcommand{\layerWeight}{\mathbf{w}}
\newcommand{\layerWeightOptimal}{\mathbf{w_*}}
\newcommand{\instant}{t}
\newcommand{\instantTwo}{j}
\newcommand{\layerWeightZero}{\layerWeight_{0}}
\newcommand{\layerWeightMean}{{}^\loss\bm{\mu}}
\newcommand{\layerWeightConvVanilla}{\layerWeight_\infty}
\newcommand{\layerWeightTimeVanilla}{\layerWeight_{\instant}}
\newcommand{\layerWeightConv}{{}^\loss\layerWeightConvVanilla}
\newcommand{\layerWeightTime}{{}^\loss\layerWeightTimeVanilla}
\newcommand{\gradient}{\mathbf{g}}
\newcommand{\gradTimeVanilla}{\gradient_\instant}
\newcommand{\gradTime}{{}^\loss\gradTimeVanilla}
\newcommand{\gradTimeTwo}{{}^\loss\gradient_\instantTwo}
\newcommand{\uselessConstant}{{c_2}}
\newcommand{\length}{\ell}

\newcommand{\layerWeightConvReg}{{}^r\layerWeightConvVanilla}
\newcommand{\layerWeightConvDice}{{}^d\layerWeightConvVanilla}

\newcommand{\weight}{\mathbf{w}}

\newcommand{\varZ}{Z}

\newcommand{\posZ}{z}

\newcommand{\norm}[1]{\left\lVert#1\right\rVert}%\left|\!\left|#1\right|\!\right|}

%% file: images/problem_setup.tex
\begin{tikzpicture}[scale=0.27, every node/.style={scale=0.45}, every edge/.style={scale=0.45}]
\tikzset{vertex/.style = {shape=circle, draw=black!70, line width=0.06em, minimum size=1.4em}}
\tikzset{edge/.style = {-{Triangle[angle=60:.06cm 1]},> = latex'}}

    % =======================================================
    % =================== Problem Setup======================
    % =======================================================
    \node [scale= 2, color=black] at (11.7, -7.5)  {(a)};

    % =============== Arrow 2 feature maps ====================
    \draw [draw=axisShadeDark, line width=0.08em, shorten <=0.5pt, shorten >=0.5pt, >=stealth]
           (6.05,1.3) node[]{}
        -- (6.05,4.1)  node[]{};
    \draw [-{Triangle[angle=60:.2cm 1]}, draw=axisShadeDark, line width=0.08em, shorten <=0.5pt, shorten >=0.5pt, >=stealth]
           (5.95,4.0) node[]{}
        -- (8.55,4.0)  node[]{};
        
    \draw [draw=axisShadeDark, line width=0.08em, shorten <=0.5pt, shorten >=0.5pt, >=stealth]
           (6.05,-0.3) node[]{}
        -- (6.05,-2.6)  node[]{};
    \draw [-{Triangle[angle=60:.2cm 1]}, draw=axisShadeDark, line width=0.08em, shorten <=0.5pt, shorten >=0.5pt, >=stealth]
           (5.95,-2.5) node[]{}
        -- (8.55,-2.5)  node[]{};
    
    % =============== 2 models maps ====================
    \node[trapezium, draw=black!100, line width=0.05em, rotate=270, fill=my_blue!30, opacity=1.0, trapezium stretches=true, minimum width=2.5cm, minimum height=1cm, trapezium left angle=75, trapezium right angle=75] (t) at (9.25,4) {};

    \node[trapezium, draw=black!100, line width=0.05em, rotate=270, fill=darkGreen3!80, opacity=1.0, trapezium stretches=true, minimum width=2.5cm, minimum height=1cm, trapezium left angle=75, trapezium right angle=75] (t) at (9.25,-2.5) {};

    \node [scale= 2] at (6.0, 0.5)  {Image $\image$};
    \node [scale= 2] at (9.25, 4){$\layerWeight$};
    \node [scale= 2, color=darkGreen3] at (14.8, -2.5)  {Length $\length$};

    \draw [-{Triangle[angle=60:.2cm 1]}, draw=darkGreen3, line width=0.08em, shorten <=0.5pt, shorten >=0.5pt, >=stealth]
           (10.05,-2.5) node[]{}
        -- (12.4,-2.5)  node[]{};

    % =============== Arrow from z to noise ====================
    \draw [draw=my_blue!30, line width=0.18em, shorten <=0.5pt, shorten >=0.5pt, >=stealth]
           (10.05,4) node[]{}
        -- (11.4,4)  node[]{};
    \node [scale= 2] at (11.9, 4)  {$\bigoplus$}; 

    % ================= Add noise =======================
    \node [scale= 2] at (11.9,9)  {Noise $\noise\!\sim\!\normal(0,\normalVar)$};

    \draw [-{Triangle[angle=60:.2cm 1]}, draw=blue, line width=0.08em, shorten <=0.5pt, shorten >=0.5pt, >=stealth]
           (11.9,7.8) node[]{}
        -- (11.9,4.5)  node[]{};
    % ================= Pred z =======================
    \draw [-{Triangle[angle=60:.2cm 1]}, draw=my_blue, line width=0.18em, shorten <=0.5pt, shorten >=0.5pt, >=stealth]
           (12.4,4) node[]{}
        -- (14.2,4)  node[scale= 2]{};

    \node [scale= 2, color=my_blue] at (13.0,5.35)  {$\depthPred$};

    % ================= Loss =======================
    \draw[draw=black!100, fill=sns_orange, thick](15.5,4.0) circle (1.3) node[scale= 2]{$\loss$};

    % ================= GT depth =======================
    \draw [-{Triangle[angle=60:.2cm 1]}, draw=darkGreen3, line width=0.08em, shorten <=0.5pt, shorten >=0.5pt, >=stealth]
           (15.5,0.7) node[]{}
        -- (15.5,2.7)  node[]{};

    \node [scale= 2, color=darkGreen3] at (15.5,-0.3)  {GT $\depthGT$};

    % =======================================================
    % =================== BEV Planes =======================
    % =======================================================
    \node [scale= 2, color=black] at (24, -7.5)  {(b)};
    \draw [draw=black!100, line width=0.06em](20, 8) rectangle (28, 2) node[]{};
    \draw [draw=black!100, line width=0.06em](20, 1) rectangle (28,-5) node[]{};

    % Camera
    \draw[black,fill=black!50] (23.7,-5.7) rectangle (24.3,-6.7);
    \coordinate (c10) at (24,-5.7);
    \coordinate (c11) at (23.7,-5.2);
    \coordinate (c12) at (24.3,-5.2);
    \filldraw[draw=black, fill=gray!20] (c10) -- (c11) -- (c12) -- cycle;

    % ================= Axis conventions =======================
    \draw [-{Triangle[angle=60:.15cm 1]}, draw=black, line width=0.04em, shorten <=0.5pt, shorten >=0.5pt, >=stealth]
           (19.0,-6.6) node[]{}
        -- (19.0,-4.6)  node[scale=2,text width=0.2cm]{$Z$\\~};

    \draw [-{Triangle[angle=60:.15cm 1]}, draw=black, line width=0.04em, shorten <=0.5pt, shorten >=0.5pt, >=stealth]
           (18.9,-6.5) node[]{}
        -- (20.9,-6.5)  node[scale=2]{~~$X$};

    % =================== Plane Names =======================
    \node [scale= 2] at (24,9)   {BEV};
    \node [scale= 2, color=darkGreen3] at (18.7,7) {GT};
    \node [scale= 2, color=my_blue  ] at (18.7,0) {Pred};
    
    % GT BEV Car and lines
    \draw [draw=rayShade, line width=0.1em]
           (24,2.08) node[]{}
        -- (24,7.95)  node[]{};
    \draw [draw=black!100, line width=0.06em, fill=darkGreen3](23.5, 4.2) rectangle (24.5, 2.2) node[]{};
    % GT Center dot and coordinates
    \draw[draw=black!100, fill=black!100, thick](24,3.2) circle (0.1) node[]{};
    \node [scale= 1.75, color=darkGreen3] at (22.25,3.2)  {$(0, \depthGT)$};

    % GT BEV Car length
    \draw [draw=gray!50, line width=0.08em]
           (25,4.2) node[]{}
        -- (25,2.2) node[pos=0.5,scale=2]{~~~$\length$};

    % Pred BEV Car and lines
    \draw [draw=rayShade, line width=0.1em]
           (24,0.98) node[]{}
        -- (24,-4.95)  node[]{};
    \draw [draw=black!100, line width=0.06em, fill=my_blue](23.5, 0.2) rectangle (24.5, -1.8) node[]{};
    % Pred Center dot and coordinates
    \draw[draw=black!100, fill=black!100, thick](24,-0.8) circle (0.1) node[]{};
    \node [scale= 1.75, color=my_blue] at (22.25,-0.8)  {$(0, \depthPred)$};

    % Pred BEV Car length
    \draw [draw=gray!50, line width=0.08em]
           (25,0.2) node[]{}
        -- (25,-1.8) node[pos=0.5,scale=2]{~~~$\length$};

    % =======================================================
    % =================== Cross Sectional View ==============
    % =======================================================
    
    % =================== GT Cross Sectional View ===========
    \node [scale= 2, color=black] at (32, -7.5)  {(c)};
    \node [scale= 2] at (32,9)   {CS View};
    \draw [-{Triangle[angle=60:.2cm 1]}, draw=rayShade, line width=0.1em, shorten <=0.5pt, shorten >=0.5pt, >=stealth]
           (33, 1.5) node[]{}
        -- (33, 8) node[scale=2,text width=0.5cm]{~\\~~~~$\varZ$};
    \draw [draw=black!100, line width=0.08em]
           (30.5, 2) node[]{}
        -- (33.5, 2) node[]{};

    % GT Car Object
    \draw [draw=darkGreen3, line width=0.12em]
           (31.42, 2.2) node[]{}
        -- (32.95, 2.2) node[]{};
    \draw [draw=darkGreen3, line width=0.12em]
           (31.5, 2.15) node[]{}
        -- (31.5, 4.25) node[]{};
    \draw [draw=darkGreen3, line width=0.12em]
           (31.42, 4.2) node[]{}
        -- (32.95, 4.2) node[]{};

    % GT center marker and depth label
    \draw [draw=black!100, line width=0.08em]
           (32.85, 3.2) node[]{}
        -- (33.15, 3.2) node[]{};
    \node [scale= 1.75, color=darkGreen3] at (33.4,3.2)  {$\depthGT$};

    % Length of car
    \draw [draw=gray!50, line width=0.1em]
           (33.8, 2.2) node[]{}
        -- (33.8, 4.2) node[pos=0.5,scale=2,text width=0.5cm]{~~~$\length$};

    % \node [scale=2, color=black,rotate=90] at (31.5,1.5)  {$1$};
    \node [scale=2, color=black,rotate=90] at (29.8,1.7)  {$P(\varZ)$};

    % =================== Pred Cross Sectional View =======================
    \draw [-{Triangle[angle=60:.2cm 1]}, draw=rayShade, line width=0.1em, shorten <=0.5pt, shorten >=0.5pt, >=stealth]
           (33, -5.5) node[]{}
        -- (33, 1.5) node[scale=2,text width=0.5cm]{~\\~~~~$\varZ$};
    \draw [draw=black!100, line width=0.08em]
           (30.5, -5) node[]{}
        -- (33.5, -5) node[]{};

    % Car Object
    \draw [draw=my_blue  , line width=0.12em]
           (31.42, 0.2) node[]{}
        -- (32.95, 0.2) node[]{};
    \draw [draw=my_blue  , line width=0.12em]
           (31.5, 0.15) node[]{}
        -- (31.5, -1.85) node[]{};
    \draw [draw=my_blue  , line width=0.12em]
           (31.42, -1.8) node[]{}
        -- (32.95, -1.8) node[]{};

    % GT center marker and depth label
    \draw [draw=black!100, line width=0.08em]
           (32.85, -0.8) node[]{}
        -- (33.15, -0.8) node[]{};
    \node [scale= 1.75, color=my_blue] at (33.4,-0.7)  {$\depthPred$};

    % Length of car
    \draw [draw=gray!50, line width=0.1em]
           (33.8, 0.2) node[]{}
        -- (33.8, -1.8) node[pos=0.5,scale=2,text width=0.5cm]{~~~$\length$};

    \draw [draw=black!50, dash pattern=on 2pt off 1.3pt, line width=0.05em]
           (31.5, -4.9) node[]{}
        -- (31.5, -1.9) node[]{};

    \node [scale=2, color=black,rotate=90] at (31.5,-5.5)  {$1$};
    \node [scale=2, color=black,rotate=90] at (29.8,-5.0)  {$P(\varZ)$};
    
\end{tikzpicture}